\documentclass[10pt,twocolumn,letterpaper]{article}

\usepackage[pagenumbers]{cvpr} % To force page numbers, e.g. for an arXiv version

\usepackage[dvipsnames]{xcolor}

\definecolor{cvprblue}{rgb}{0.21,0.49,0.74}
\usepackage[pagebackref,breaklinks,colorlinks,citecolor=cvprblue]{hyperref}

\title{RISE-SDF: a \underline{R}elightable \underline{I}nformation-\underline{S}har\underline{e}d Signed Distance Field for Glossy Object Inverse Rendering}

\author{Deheng Zhang$^{1,2}$\footnotemark[1] \hspace{5mm}  Jingyu Wang$^{1}$\footnotemark[1] \hspace{5mm}  Shaofei Wang$^1$ \hspace{5mm}  Marko Mihajlovic$^1$ \\
Sergey Prokudin$^1$ \hspace{5mm}  Hendrik P.A. Lensch$^2$ \hspace{5mm}  Siyu Tang$^1$\\
$^1$ETH Zürich \hspace{5mm}  $^2$University of Tübingen\\
}

\usepackage{pifont}% http://ctan.org/pkg/pifont
\usepackage{multirow}
\usepackage{colortbl}
\usepackage{xcolor}
\usepackage{graphicx}
\usepackage{tabularx}
\usepackage{subcaption}
\usepackage{multicol}
\usepackage{enumitem}
\usepackage{latexsym}
\usepackage{amssymb}
\usepackage{bbm}

\definecolor{yellow}{rgb}{1,0.97, 0.65}
\definecolor{lightyellow}{rgb}{1,1, 0.8}
\definecolor{orange}{rgb}{1, 0.85, 0.7}
\definecolor{tablered}{rgb}{1, 0.7, 0.7}

\newcommand{\cmark}{\ding{51}}%
\newcommand{\xmark}{\ding{55}}%
\newcommand{\rowname}[1]

\newcommand\blfootnote[1]{%
  \begingroup
  \renewcommand\thefootnote{}\footnote{#1}%
  \addtocounter{footnote}{-1}%
  \endgroup
}
\newlength{\tempdima}

\begin{document}
\maketitle
{
    \renewcommand{\thefootnote}%
        {\fnsymbol{footnote}}
        \footnotetext[1]{Equal contribution.}
}
\blfootnote{Project page: \url{https://dehezhang2.github.io/RISE-SDF/}}

\begin{figure*}[h]
    \centering
    \includegraphics[width=\textwidth]{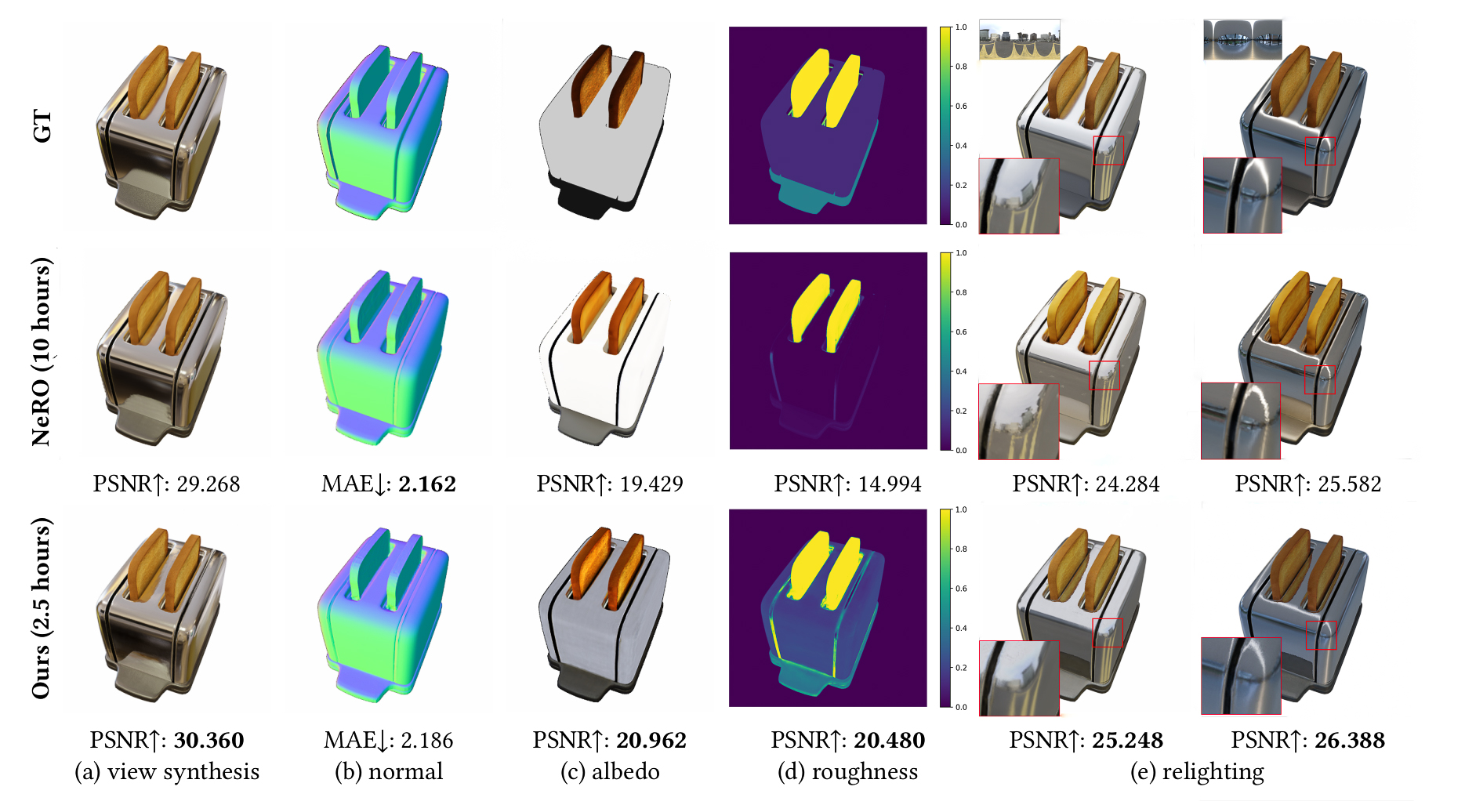}
    \caption{\textbf{RISE-SDF.} We present RISE-SDF, a method for reconstructing the geometry and material of glossy objects while achieving high-quality relighting. Our results, compared with the state-of-the-art method~\cite{liu2023nero}, show superior albedo and roughness estimation with significantly less training time. As an end-to-end relightable model, our algorithm generates high-quality relighting images without noise or aliasing.}
    \label{fig:teaser}
    \vspace{-2em}
\end{figure*}

\begin{abstract}
Inverse rendering aims to reconstruct the 3D geometry, material parameters, and lighting conditions in a 3D scene from multi-view input images. To address this problem, some recent methods utilize a neural field combined with a physically based rendering model to reconstruct the scene parameters. Although these methods achieve impressive geometry reconstruction for glossy objects, the performance of material estimation and relighting remains limited. In this paper, we propose a novel end-to-end relightable neural inverse rendering system that achieves high-quality reconstruction of geometry and material properties, thus enabling high-quality relighting. The cornerstone of our method is a two-stage approach for learning a better factorization of scene parameters. In the first stage, we develop a reflection-aware radiance field using a neural signed distance field (SDF) as the geometry representation and deploy an MLP (multilayer perceptron) to estimate indirect illumination. In the second stage, we introduce a novel information-sharing network structure to jointly learn the radiance field and the physically based factorization of the scene. For the physically based factorization, to reduce the noise caused by Monte Carlo sampling, we apply a split-sum approximation with a simplified Disney BRDF and cube mipmap as the environment light representation. In the relighting phase, to enhance the quality of indirect illumination, we propose a second split-sum algorithm to trace secondary rays under the split-sum rendering framework. Furthermore, there is no dataset or protocol available to quantitatively evaluate the inverse rendering performance for glossy objects. To assess the quality of material reconstruction and relighting, we have created a new dataset with ground truth BRDF parameters and relighting results. Our experiments demonstrate that our algorithm achieves state-of-the-art performance in inverse rendering and relighting, with particularly strong results in the reconstruction of highly reflective objects. 
\end{abstract}    
\section{Introduction}
\label{sec:intro}
Inverse rendering is an ill-posed problem in computer graphics, which aims to reconstruct the 3D geometry, material parameters (texture, metallic, and roughness), as well as light conditions in a 3D scene given multi-view input images and corresponding camera poses. With the reconstructed assets, we can further change the light condition to render the object in new environments. This will serve as an essential step in 3D assets acquisition for the gaming and movie industries. To solve this problem, traditional inverse rendering algorithms \cite{NimierDavidVicini2019Mitsuba2} make the forward rendering process differentiable and optimize the scene parameters to achieve the required factorization. However, reconstructing complex real-world geometry remains a challenge within these pipelines. Recently, the neural radiance field (NeRF) \cite{Mildenhall2020ECCV} has demonstrated a superior ability to achieve novel view synthesis. Follow-up works such as NeuS~\cite{wang2021neus}, and VolSDF~\cite{yariv2021volume} further improve the quality of geometry reconstruction using neural signed distance functions. To improve the performance for glossy objects, Ref-NeRF~\cite{verbin2022ref} further changes the directional encoding to consider the reflected direction and roughness into the reconstruction pipeline. However, these methods only recover the geometry of the 3D object and do not reconstruct the material and light information in the scene, thus they can only achieve novel view synthesis without the ability to relight the object. Utilizing the neural radiance field as the geometry representation, neural inverse rendering methods~\cite{boss2021nerd, jin2023tensoir, liang2023envidr, liu2023nero, zhang2021nerfactor, srinivasan2021nerv, mai2023neural, Sengupta2019-uf, Yang2022-de, Knodt2021-di} further factorize the radiance field with physically based rendering formulation. However, none of the current algorithms simultaneously achieve high-quality scene parameter reconstruction, novel view synthesis, and relighting for glossy objects. Additionally, the currently available glossy object datasets do not provide ground truth material parameters or relighting information, making it difficult to evaluate model performance.

\noindent In this paper, we propose an end-to-end relightable inverse rendering system for glossy objects. First, we explore how to construct high-quality geometry for glossy objects as an initial step in appearance parameter estimation. We find that an accelerated version of NeuS~\cite{wang2021neus} with a reflection-aware radiance field yields the best results for glossy geometry. More specifically, we use three networks to predict the diffuse, specular direct illumination, the blending weights, and an additional MLP to predict the indirect illumination solely for the surface intersection. In stage two, we propose an information-sharing MLP structure to learn the appearance parameters and the environment light. 
Due to the observation that Monte Carlo integration generates noise, we adopt the split-sum approximation~\cite{karis2013real}. Our method uses a pre-integrated Disney BRDF~\cite{burley2012physically}, stored in a 2D lookup texture (LUT), as the physically based appearance model. Additionally, a differentiable multi-level mipmap serves as the environment map.  We further train this representation jointly with the radiance field in stage one to maintain a high-quality geometry during the optimization. For the relighting, we propose a second, modified split-sum algorithm evaluated on secondary rays to achieve consistent relighting that accounts for indirect illumination in the scene. To evaluate our model's performance, we introduce a new synthetic dataset for glossy objects with ground truth relighting and material maps. Our results demonstrate that our algorithm not only excels in novel view synthesis and geometry reconstruction quality but also achieves state-of-the-art performance as an end-to-end relightable radiance field.

\section{Related Work}
\label{sec:related_work}
\subsection{Traditional Rendering}
% \textbf{Forward Rendering.} Physics-based rendering applies ray tracing and recursive Monte Carlo sampling to simulate light integration. Additionally, bidirectional reflection distribution functions (BRDFs) are constructed to represent the material property. The microfacet model~\cite{cook1982reflectance} is proposed to better formulate the glossy reflection by simulating local facets on the surface. As a part of the microfacet model, the normal distribution function (NDF) describes the spread of the light reflection. GGX~\cite{trowbridge1975average} is one commonly used NDF because of its simplicity during computing. Disney principled BRDF \cite{burley2012physically} is then proposed to give more controllability during 3D asset creation. The split-sum approximation~\cite{karis2013real} is proposed to accelerate the rendering process, which introduces a multi-level mipmap to represent the environment light. We also employ this representation in our inverse rendering pipeline. 
\textbf{Forward Rendering.} Physically based rendering (PBR) seeks to render
images in a way that simulates light transport of scenes in real-world.  It
usually employs physically plausible bidirectional reflection distribution
functions (BRDFs) to describe the way how light is scattered by object surfaces.
Microfacet model~\cite{cook1982reflectance} is the most widely used BRDF model
for PBR.  It simulates a macro surface as a collection of microfacets.  Contrary
to earlier BRDF models~\cite{Phong1975ACM}, the microfacet model is considered
more physically plausible as it satisfies reciprocity and energy conservation.
While there exists microfacet-based diffuse BRDF
model~\cite{OrenAndNayar1994SIGGRAPH}, in this work we follow modern rendering
pipelines~\cite{burley2012physically,karis2013real} and focus on specular
microfacet BRDFs in which microfacets are assumed to be perfect mirrors.  As an
important component of the microfacet model, the normal distribution function
(NDF) describes the distribution of the orientations (\ie\ normals) of
microfacets on the surface.  GGX~\cite{trowbridge1975average} is one commonly
used NDF because of its simplicity for evaluation. Popular PBR models such as
the Disney BRDF~\cite{burley2012physically} uses GGX model and variants as its
core relfection models.  Late, the split-sum approximation~\cite{karis2013real}
is proposed to achieve real-time rendering in game engines.  It introduces a
multi-level mipmap to store pre-convolved environment light, such that it can be
efficiently queried to render images without time-consuming Monte Carlo
sampling.  We also employ split-sum in our inverse rendering pipeline mainly due
to its efficiency.

\noindent{\textbf{Inverse Rendering.}} Inverse rendering aims to recover the geometry, material, and light condition given the 2D images. In our work, we assume the camera pose is known, the light source is far away from the object, and the BRDFs are isotropic. The traditional inverse rendering algorithm~\cite{ramamoorthi2001signal} represents both BRDF and environment light with spherical harmonics. Later methods apply differentiable rasterization~\cite{loper2014opendr,liu2019soft,chen2019learning} or ray tracing \cite{NimierDavidVicini2019Mitsuba2,li2018differentiable,azinovic2019inverse} to reconstruct the material. However, \citet{Zeltner2021MonteCarlo} show that it is challenging to make Monte Carlo sampling differentiable without increasing the noise and suggests using a PDF proportional to the differentiation of the GGX distribution \cite{trowbridge1975average}. Our work finds that using the split-sum approximation \cite{karis2013real} bypasses this problem and achieves noise-free reconstruction results.

\subsection{Neural Radiance Field}

\textbf{Novel View Synthesis.}
NeRF \cite{Mildenhall2020ECCV} proposes a radiance field-based representation of the 3D scene. The method is trained using 2D multi-view images for which the camera pose is provided and uses two MLPs to predict the density $\sigma$ and radiance $\textbf{c}$ of a given sample on the ray. Then the volumetric rendering equation is applied to render the image for different views. Since the information is stored in 3D, view consistency is ensured. Variants of NeRF modify the structure to either improve the rendering quality or speed up the training of NeRF. For example, Ref-NeRF \cite{verbin2022ref} uses the reflected direction of the camera ray as the input of the MLP to improve the rendering quality for specular materials. \cite{wuneural} further improve the glossy reconstruction by using a neural directional encoding framework to separate the direct and indirect illumination encoding. \citet{fridovich2022plenoxels, chen2022tensorf, muller2022instant} modify the data structure of the radiance field to improve the training speed of NeRF. In our paper, we use instant-NGP~\cite{muller2022instant} with the hash grid as our backbone model for radiance prediction, since this algorithm converges faster compared to the original NeRF. Furthermore, recently proposed Gaussian Splatting \cite{kerbl20233d} utilizes 3D Gaussians with center and covariance matrix as the 3D representation, and uses rasterization-based rendering to achieve a higher reconstruction speed. However, since the normal information is not well reconstructed in this method, which is essential during inverse rendering, we do not apply this rendering technique as our backbone. 

\noindent{\textbf{Neural Geometry Reconstruction.}}
Based on the density representation in NeRF \cite{Mildenhall2020ECCV}, some methods \cite{wang2021neus, yariv2021volume} construct neural SDF fields with functions to transform the signed distance to density. Follow-up methods \cite{10.1145/3550454.3555451,wang2023neus2,li2023neuralangelo} further improve the speed and quality of the geometry reconstruction through hash encoding and progressive training. \citet{yariv2023bakedsdf} extend the reconstruction to unbounded scenes. In our paper, we use the pipeline proposed in \cite{10.1145/3550454.3555451} as the fundamental geometry representation and use the progressive training proposed in \cite{li2023neuralangelo} to improve the quality.
\vspace{-0.5em}
\subsection{Neural Inverse Rendering}
\vspace{-0.5em}
Following the idea of implicit representation in NeRF \cite{Mildenhall2020ECCV}, neural inverse rendering algorithms use MLPs to store the material information. NDR and NDRMC~\cite{munkberg2022extracting,hasselgren2022shape} use an MLP to predict the BRDF material parameters, and DMTet \cite{shen2021deep} as the geometry representation. Compared with neural SDF, these methods result in discontinuous artifacts during the reconstruction. NeRFactor~\cite{zhang2021nerfactor}, NeRD~\cite{boss2021nerd}, Neural-PIL \cite{boss2021neuralpil}, NeRV~\cite{srinivasan2021nerv}, Neural Transfer Field~\cite{lyu2022nrtf}, InvRender~\cite{zhang2022invrender} and TensoIR~\cite{jin2023tensoir} use a density field as the geometry representation and an environment tensor with Monte Carlo sampling for the light reconstruction. To solve the ambiguity of the base color and environment light, \cite{kouros2024unveiling, chen2024intrinsicanything, 10.1111:cgf.15147} show the importance of adding a material prior to inverse rendering. However, these methods fail to reconstruct the material properties of glossy objects. To improve the performance for glossy objects, Neural microfacet fields (NMF) \cite{mai2023neural} use the microfacet model \cite{cook1982reflectance} with the GGX distribution \cite{trowbridge1975average}  to improve the reconstruction quality, but the geometry and relighting quality are still sub-optimal as they use radiance representation and Monte Carlo sampling for rendering. To improve the geometry reconstruction, NeRO \cite{liu2023nero} and ENVIDR \cite{liang2023envidr} use neural SDF as the geometry representation, but they are still limited in terms of relighting availability and quality. Their neural representation of the environment map means they can only achieve relighting via pre-distillation or export to the Blender \cite{blender} renderer, which results in aliasing or blurriness. Gaussian splatting-based inverse rendering \cite{shi2023gir,liang2024gsir,wu2024deferredgs,R3DG2023,jiang2023gaussianshader} uses the 3D Gaussian as the geometry representation to improve the inverse rendering speed, but these methods either have bad reconstruction on glossy objects or have bad geometry due to the Gaussian representation. Our algorithm uses SDFs as the geometry representation and the split-sum approximation as the light representation. More importantly, to achieve better relighting quality, we apply a second split algorithm to trace indirect illumination. Compared with NeRO and ENVIDR, we can directly switch the environment map with an HDR file without any initialization or data export and achieve high-quality relighting. 

\section{Method}
We first introduce the necessary concepts for volume rendering and physically based rendering (PBR) in \cref{sec:background}. In \cref{sec:stage1}, we describe the first stage of our framework that reconstructs the radiance field and the geometry. In \cref{sec:stage2}, we combine volume rendering with traditional physically based split-sum approximation under the information-sharing MLP framework for material and lighting estimation. Furthermore, we introduce our proposed indirect illumination MLP and a novel second split-sum rendering algorithm in \cref{sec:stage3} to handle indirect illumination. This enables high-quality reconstruction and relighting with indirect illumination. 

\subsection{Preliminaries}
\label{sec:background}
\noindent{\textbf{Volume Rendering Model.}} The general volume rendering equation of NeRF is as follows:
\begin{align}
\label{eqn:vol_render_neus}
   \mathbf{C}(\mathbf{d}) = \sum\nolimits^N_{i=1} w_i \mathbf{c}_i,\, \text{where}\ w_i=\Pi^{i-1}_{j=1} (1 - \alpha_j)\alpha_i
\end{align}
here $\mathbf{d}$ represents the view direction, $\mathbf{c}_i$ represents the per-sample radiance.  $\alpha_i = 1 - \exp (-\sigma_t^i) \delta_i$ where $\delta_i$ and $\sigma_t^i$ represent interval length and density of sample $i$, respectively.  Instead of using this density representation, we utilize NeuS~\cite{wang2021neus} to better represent the underlying geometry as a neural signed distance field (SDF). The neural SDF defines a signed distance function $s : \mathbb{R}^3 \mapsto \mathbb{R}$, which is predicted by neural networks. The surface $\mathcal{S}$ is represented by the zero-level set of the neural SDF: 
\begin{align}
 \label{eqn:sdf_isosurface}
  \mathcal{S} = \{\mathbf{x} \in \mathbb{R}^3 |  s(\mathbf{x}) = 0\}
\end{align}
%
%In order to make the rendering equation consistent with NeRF, 
NeuS defines a S-density function $\phi_{\sigma}(\mathbf{x}) = {\sigma}e^{-{\sigma}\mathbf{x}}/(1 + e^{-{\sigma}\mathbf{x}})^2$ to convert the SDF value into density.
% and it is the derivative of the Sigmoid function $\Phi_{\sigma}(\mathbf{x}) = (1 + e^{-{\sigma}\mathbf{x}})^{-1}$.
The standard deviation of the density function is given by $1/{\sigma}$, which is learnable during the training process. To further make sure that the rendering equation is unbiased, NeuS introduces a normalized S-density as the rendering weight. As a result, the opacity $\alpha$ of the sample $i$ becomes:
\begin{align}
\label{eqn:alpha_occ}
   \alpha_i = \max\left({{\Phi_{\sigma}(s(\mathbf{x}_i)) - \Phi_{\sigma}(s(\mathbf{x}_{i+1}))\over \Phi_{\sigma}(s(\mathbf{x}_i))}}, 0\right)
\end{align}
%
% \begin{equation}
% \begin{aligned}
%  \mathcal{L}_{eik} &= {1 \over N}\sum^N_{i=1}(||\nabla s(\mathbf{x}_i)||_2 - 1)^2 \\
%  \mathcal{L}_{curv} &= \sum_x({\mathbf{n} } \cdot {\mathbf{n}_{\epsilon}} - 1)^2
% \end{aligned}
%  \label{eqn:ggx}
% \end{equation}

\noindent{\textbf{Physically Based Rendering (PBR).}} In our PBR pipeline , we follow~\cite{mai2023neural} and treat each sample during the volume rendering as a small local surface. Shading of each local surface follows the standard rendering equation:
\begin{align}
\label{eqn:rendering_equation}
 \mathbf{c}^{pbr}_i = \int_{H^2}f_r(\mathbf{x}_i, \mathbf{d}, \mathbf{\omega}_j)L_i(\mathbf{x}_i, \mathbf{\omega}_j) (\mathbf{n}_i \cdot \mathbf{\omega}_j) d\mathbf{\omega}_j 
\end{align}
where $-\mathbf{d}$ and $\mathbf{\omega}_j$ are the outgoing and incoming
directions (surface-to-camera direction and surface-to-light direction, respectively). This equation shows that the outgoing radiance $\mathbf{c}^{pbr}_i$ is the integration of the BRDF $f_r$ multiplied by the incoming radiance and cosine term over the hemisphere defined by the surface normal $\mathbf{n}_i$. In our paper, the BRDF is defined as a simplified version of the Disney BRDF~\cite{burley2012physically}:
\begin{align}
\label{eqn:our_bsdf}
 & f_r(\mathbf{x}_i, \mathbf{d}, \mathbf{\omega}_j) = (1 -m_i) {\mathbf{a}_i \over \pi} + {D(\rho_i)F(m_i, \mathbf{a}_i)G \over 4 |\mathbf{d} \cdot \mathbf{n}_i| |\mathbf{\omega}_j \cdot \mathbf{n}_i|}
\end{align}
where albedo $\mathbf{a}_i$, metallic $m_i$, and roughness $\rho_i$ are spatially varying material parameters, $\mathbf{n}_i$ is the surface normal. The first term of this equation represents the diffuse shading, while the second term
corresponds to specular shading that is modeled by the microfacet model~\cite{cook1982reflectance}. $F$ stands for the Fresnel term. $D$ is the normal distribution function (NDF) depending on roughness $\rho$.  We use GGX~\cite{trowbridge1975average} for $D$.
%thanks to its efficient computation.
$G$ is the shadowing term, which models the shadowing effect between microfacets. In this work, we use the split-sum approximation~\cite{karis2013real} due to its efficiency in estimating the integral in \cref{eqn:rendering_equation}. It approximates the radiance by pre-integration such that only a lookup per pixel is required to evaluate the reflected radiance finally: 
\begin{equation}
    \begin{aligned}
        \label{eqn:split-sum}
         \mathbf{c}^{pbr}_i(\mathbf{x}_i, \mathbf{d}) 
         \approx &({1 \over N_{mc}} \sum^{N_{mc}}_{j=1} {f_r(\mathbf{x}_i, \mathbf{d}, \mathbf{\omega}_j) (\mathbf{n}_i \cdot \mathbf{\omega}_j) \over p(\mathbf{\omega}_j; \mathbf{\hat{d}}_i, \rho_i)})  \\ 
         &\cdot({1 \over N_{mc}} \sum^{N_{mc}}_{j=1} L_i(\mathbf{x}_i, \mathbf{\omega}_j) )
    \end{aligned}
\end{equation}

\noindent given the sample location $\mathbf{x}_i$, ray direction $\mathbf{d}$, normal $\mathbf{n}_i$, and reflected ray direction $\mathbf{\hat{d}}_i$ w.r.t the normal. 
$N_{mc}$ is the number of samples during pre-integration and $p$ represents the probability of sampling the pair of directions. 
%$pdf$ is a distribution on the direction $\mathbf{\omega}_j$ (in practice we use NDF as the density function). 

% For a given normal defined in the local frame $\mathbf{\omega_h} = (x_n, y_n, z_n)$
% The GGX is defined as:
% \begin{equation}
% \begin{aligned}
%  D(\mathbf{\omega_h}) = {1 \over \pi \alpha_x \alpha_y ({ {x_n^2 \over \alpha_x^2} + {y_n^2 \over \alpha_y^2} + z_n^2)^2}}, 
% \end{aligned}
%  \label{eqn:ggx}
% \end{equation}
% where the roughness values $\alpha_x$ $\alpha_y$ are two roughness parameters along the two principle axes orthogonal to the normal. In isotropic material setting the roughness value $\rho = \alpha_x = \alpha_y$. We can observe that the roughness is proportional to the variance of the NDF, which means larger roughness corresponds to a larger lobe. 

\subsection{Stage 1: Geometry Reconstruction}
\label{sec:stage1}
\begin{figure}[ht]
    \centering
    \includegraphics[width=0.45\textwidth]{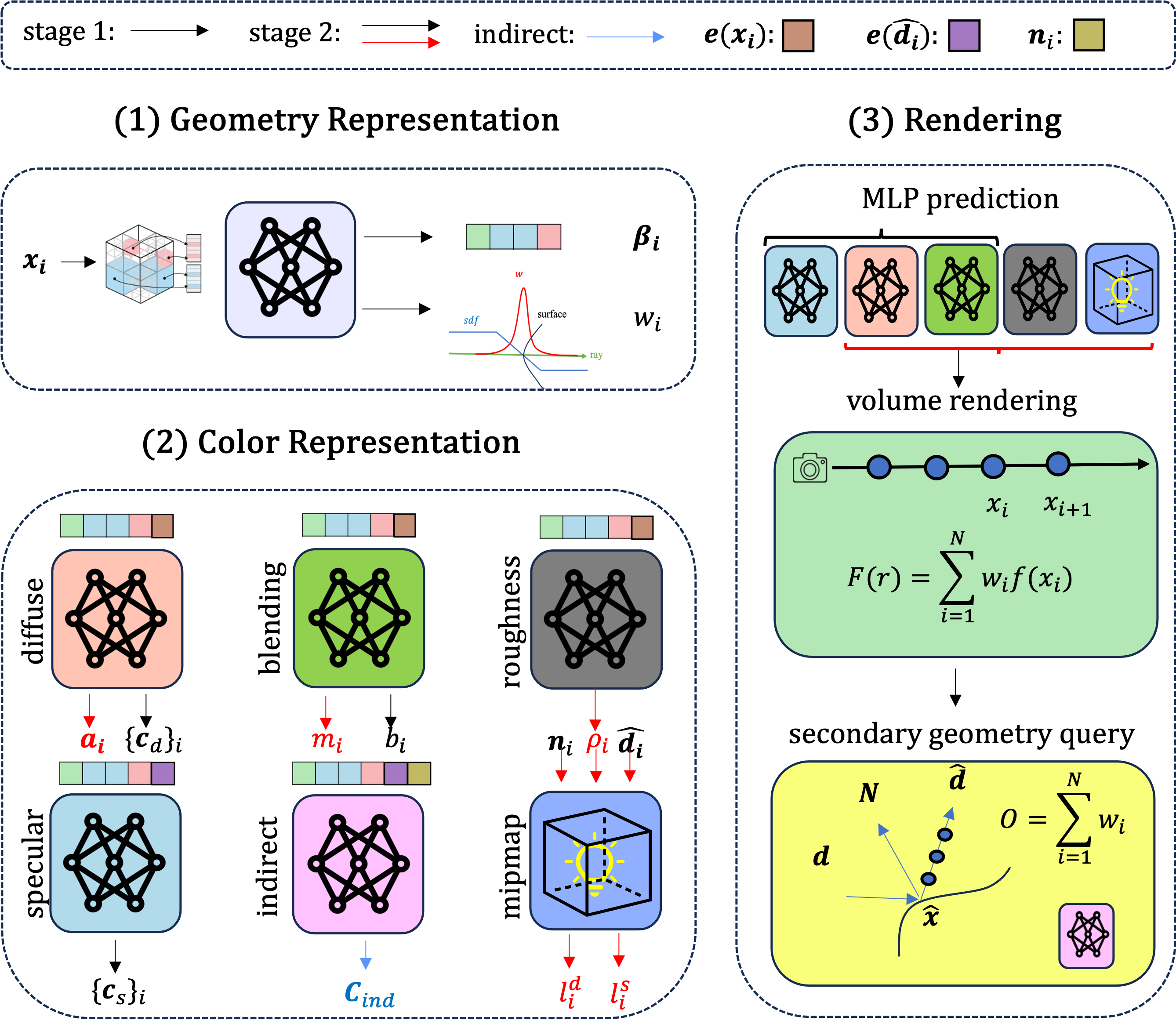}
    \caption[First Stage Pipeline]{\textbf{Our pipeline. The colors of the features in the figure indicate different feature concatenation combinations. }(1) Given the location $\boldsymbol{x}_i$, the progressive hash grid with the geometry MLP predicts the geometry feature $\beta_i$ and the corresponding volume rendering weight $w_i$ via Eq. \ref{eqn:vol_render_neus} and \ref{eqn:alpha_occ}. (2) For the color representation, separate networks predict per-sample albedo $\boldsymbol{a_i}$, metallic $m_i$, and roughness $\rho_i$. They share the information between the direct volume rendering pipeline (black arrow) and the physically based rendering pipeline (red arrow). (3) The per-sample values (all except the blue value) are rendered via volume rendering. Then we compute the expected surface intersection $\boldsymbol{\hat{x}}$ and trace another ray to compute the occlusion probability $O$. Finally, the direct and indirect colors are blended via $O$.}
    \label{fig:pipeline}
    \vspace{-2em}
\end{figure}
\textbf{Geometry Representation}.
We use instant NGP (iNGP)~\cite{muller2022instant}, which consists of a multi-level hash encoding and a small multi-layer perceptron (MLP), to predict the neural SDF.  As shown in \cref{fig:pipeline}, we use a progressive hash grid $\gamma$ followed by a geometry MLP $f_{geo}$ to predict the SDF value $s$ and a geometry feature $\mathbf{\beta}$:
\begin{align}
\label{eqn:geo_pred}
s_i, \mathbf{\beta}_i = f_{geo}(\gamma(\mathbf{x}_i)) 
\end{align}
Following \cref{eqn:vol_render_neus} and \cref{eqn:alpha_occ}, we can compute the volume rendering weight $w_i$ and normal vector $\mathbf{n}_i = { \nabla_{\mathbf{x}_i} s_i \over ||\nabla_{\mathbf{x}_i} s_i ||}$ for each sample.

\noindent \textbf{Radiance Representation}. The direct illumination is estimated using the spatial location $\mathbf{x}_i$ and reflected direction $\hat{\mathbf{d}}$ as inputs akin to~\citet{verbin2022ref}. The reflected direction is computed as:
\begin{align}
\label{eqn:reflection}
\mathbf{\hat{d}}_i = -2(\mathbf{d} \cdot \mathbf{n}_i) \mathbf{n}_i + \mathbf{d}
\end{align}
The encodings $e(\mathbf{x}_i)$ and $e(\mathbf{\hat d}_i)$ are respectively computed via positional and spherical harmonic encodings. We define three MLPs, $f_{diff}$, $f_{spec}$, $f_{blend}$, and predict the diffuse color \{$\mathbf{c}_d\}_i = f_{diff}(e(\mathbf{x}_i))$, specular color $\{\mathbf{c}_s\}_i =f_{spec}(e(\mathbf{\hat d}_i))$, and the blending weight $b_i = f_{blend}(e(\mathbf{x}_i), \mathbf{\beta}_i)$. We extend the original volume rendering equation \cref{eqn:vol_render_neus} for these values as a function of sample location $\mathbf{x}_i$: 
\begin{equation}
\begin{aligned}
    \{\mathbf{c}_d\}_i = (1 - b_i) * \{\mathbf{c}_d\}_i;\ \  \{\mathbf{c}_s\}_i = b_i * \{\mathbf{c}_s\}_i \\
   \{\mathbf{C}_d, \mathbf{C}_s, \mathbf{N}, T\} = \sum\nolimits^N_{i=1} w_i \{\mathbf{c}_d, \mathbf{c}_s, \mathbf{n}, t\}_i
\end{aligned}
\label{eqn:vol_render_extend}
\end{equation}
Note that $t_i$ are the sample values during ray marching. In this way, we get the diffuse radiance $\mathbf{C}_d$ and specular radiance $\mathbf{C}_s$ under direct illumination. %This formulation directly predicts the result of Monte-Carlo integration for the diffuse part and specular part.
More details are discussed in the supplementary.  We will describe how to model indirect illumination in \cref{sec:stage3}

% \subsection{Stage 2: Information Sharing for Material \& Light Estimation}
\vspace{-1em}
\subsection{Stage 2: Material and Light Estimation}
\label{sec:stage2}

% In stage 2, observing that monte-carlo sampling results in high noise during the novel view synthesis, we decide to use split-sum approximation as done in NVDiffrec~\cite{munkberg2022extracting} and NeRO \cite{liu2023nero}. However, benefiting from our information-share and indirect illumination prediction, our algorithm performs better than the previous split-sum-based neural inverse rendering algorithms. The physical-based rendering using split-sum approximation is represented as follows:
In Stage 2, we optimize materials and environment lighting via differentiable PBR, given the geometry and radiance field initialized in Stage 1. Note that in stage 2, geometry and the radiance field are optimized jointly with PBR parameters, contrary to~\cite{lyu2022nrtf,zhang2022invrender, liu2023nero}. The PBR appearance $\mathbf{c}^{pbr} = \mathbf{c}^{pbr}_d + \mathbf{c}^{pbr}_s$ is computed via the split-sum approximation as follows:
\begin{equation}
\begin{aligned}  
\{\mathbf{c}^{pbr}_d\}_i &= (1 - m_i) * \mathbf{a}_i * \mathbf{l}^d_i(\mathbf{n}_i) \\
\{\mathbf{c}^{pbr}_s\}_i &= (F_0(m_i, \mathbf{a}_i) * F_1(\rho_i, \mathbf{n}_i \mathbf{\hat{d}}) \\
&+ F_2(\rho_i, \mathbf{n}_i \mathbf{\hat{d}})) * \mathbf{l}^s_i(\rho_i, \mathbf{\hat{d}}_i)
\end{aligned}
\label{eqn:pbr_split-sum}
\end{equation}
where $F_1, F_2$ are integral values only dependent on the roughness and cosine between the normal and reflection direction, $\mathbf{l}^d_i, \mathbf{l}^s_i$ are light integrals approximated by a differentiable mipmap. More details and derivation of the split-sum approximation are discussed in the supplementary.  Finally, we apply volume rendering for the diffuse and specular appearance of Stage 2:
\begin{equation}
\begin{aligned}
\{\mathbf{C}^{pbr}_d, \mathbf{C}^{pbr}_s\} &= \sum\nolimits^N_{i=1} w_i \{\mathbf{c}^{pbr}_d, \mathbf{c}^{pbr}_s\}_i \nonumber
\end{aligned}
\label{eqn:stage2_volrend}
\end{equation}

\noindent \textbf{Information Sharing between Stage 1 \& 2}.
% We need to predict the albedo, metallic, and roughness as the material parameters for rendering in this equation.
Comparing \cref{eqn:pbr_split-sum} and \cref{eqn:vol_render_extend}, we note that both blending weight $b_i$ and metallic $m_i$ are used to mix the diffuse and specular colors. Besides, diffuse color $c^d_i$ and albedo $\mathbf{a}_i$ are both view-independent and low-entropy.  We thus propose to share this information by expanding the output dimension of the MLP in Stage 1: \{$\mathbf{c}^d_i, \mathbf{a}_i\} = f_{diff}(e(\mathbf{x}_i); \mathbf{\beta}_i), \{b_i, m_i\} = f_{blend}(e(\mathbf{x}_i), \mathbf{\beta}_i)$, and create a new MLP for the roughness prediction $\rho_i = f_{rough}(e(\mathbf{x}_i), \mathbf{\beta}_i)$. We use MLP to predict all parameters in both stages and only optimize the albedo and metallic parameters in stage 2. As shown in \cref{fig:pipeline}, we jointly optimize the PBR loss and the radiance field loss from Stage 1. This MLP sharing technique provides a novel way to initialize the material MLP, thus giving more stable material prediction as shown in supplementary Fig. 6.

\subsection{Modeling Indirect Illumination}
\label{sec:stage3}
\begin{figure}
    \centering
    \begin{tabular}{ccc}
    \multicolumn{3}{c}{\includegraphics[width=0.41\textwidth]{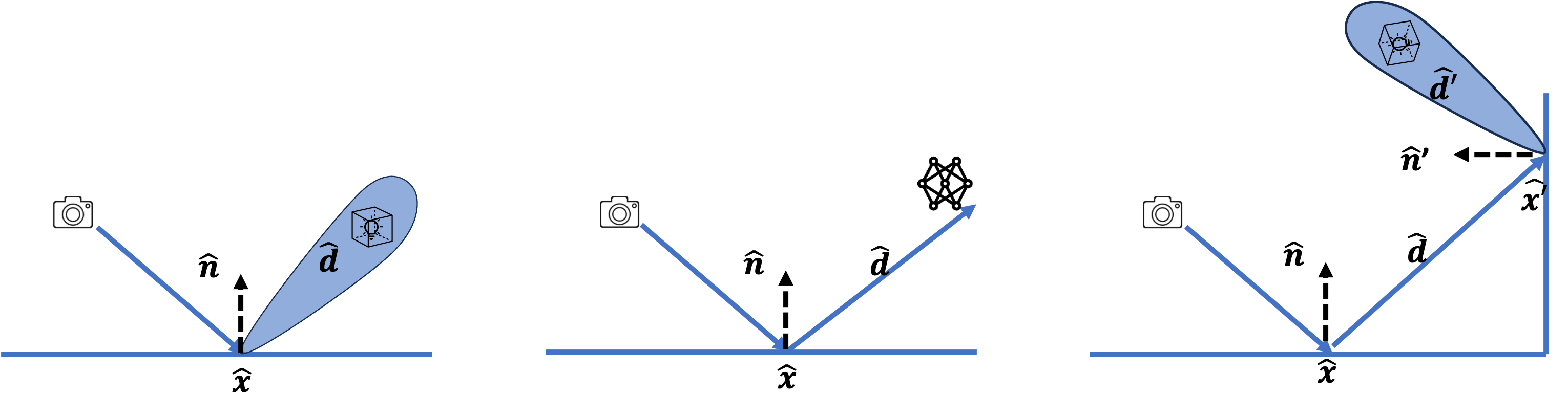}}\\
    (a) 1st split-sum & (b) indirect MLP & (c) 2nd split-sum  \\
    \end{tabular}
    \caption{\textbf{Different Illumination Inference Method}. (a) During training and relighting, we use the first split-sum to compute the direct illumination. (b) During training, we use an MLP to predict indirect illumination and blend with direct illumination using the occlusion probability. (c) During relighting, we use a second split-sum with one additional ray bounce to compute the indirect illumination. }
    \label{fig:illumination}
    \vspace{-1.5em}
\end{figure}
% To further improve the robustness of our system when indirect illumination exists, as shown in Figure \ref{fig:illumination}, we further introduce indirect illumination MLP during training and second split-sum during relighting.
So far we have excluded indirect illumination from our model, whereas in practice modeling indirect illumination is crucial for reconstructing and rendering realistic scenes, especially when the scene is highly specular. We thus introduce two novel techniques to model indirect illumination effects: indirect illumination MLP for training and second split-sum for test-time relighting.

% \boldparagraph{Indirect Illumination MLP}  Observing that indirect effect normally happens for specular part of the radiance, we add a residual term to refine the specular radiance. In Section~\ref{sec:stage1}, we compute the expected sample values $T$, this is used to compute the expected surface intersection $\mathbf{\hat{x}} = \mathbf{o} + T  \mathbf{d}$. Then, we use this surface location as input to geometry models $\{\gamma, f_{geo}\}$ to compute the corresponding feature $\mathbf{\beta_{\hat{x}}}$. As shown in Figure \ref{fig:pipeline}, to better model the indirect illumination, we use another MLP $f_{ind}$ to predict the indirect illumination $\mathbf{C}_{ind} = f_{ind}(\mathbf{N}, e(\mathbf{\hat{d}}), \mathbf{\beta}_{\hat{x}})$, where $\mathbf{\hat{d}}$ is the reflected secondary ray direction along the expected normal $\mathbf{N}$.  After that, we apply the ray marching along the secondary ray direction $\mathbf{\hat{d}}$ only to compute the volume rendering weights $w_i$ and do not query the color MLPs. We use the weights to compute the occlusion probability $O = \sum_{i=1}^{N}w_i$ and use this value to blend the specular part as a weighted sum of direct and indirect illumination:
\noindent \textbf{Indirect Illumination MLP}.  Observing that indirect illumination is often most relevant for the specular component of the incoming radiance, we add a residual term to refine the specular radiance in Stage 1 (\cref{sec:stage1}). Note that we only estimate the indirect illumination once for one primary ray for efficiency. Specifically, we compute the expected depth values $T$, which is used to compute the expected surface intersection $\mathbf{\hat{x}} = \mathbf{o} + T  \mathbf{d}$. Then, we use this surface location as input to geometry models $\{\gamma, f_{geo}\}$ to compute the corresponding feature $\mathbf{\beta_{\hat{x}}}$. As shown in \cref{fig:pipeline} and \cref{fig:illumination}(b), to better model the indirect illumination, we use another MLP $f_{ind}$ to predict the indirect illumination $\mathbf{C}_{ind} = f_{ind}(\mathbf{N}, e(\mathbf{\hat{d}}), \mathbf{\beta}_{\hat{x}})$, where $\mathbf{\hat{d}}$ is the reflected secondary ray direction along the expected normal $\mathbf{N}$. After that, we apply the ray marching along the secondary ray direction $\mathbf{\hat{d}}$ only to compute the volume rendering weights $w_i$ and do not query the color MLPs. We use the weights to compute the occlusion probability $O = \sum_{i=1}^{N}w_i$ and use this value to blend the specular part as a weighted sum of direct and indirect illumination:
\begin{equation}
\begin{aligned}
\mathbf{C} &=  \mathbf{C}_d + (1-O) * \mathbf{C}_s + O* \mathbf{C}_{ind} \\
\mathbf{C}^{pbr} &=  \mathbf{C}^{pbr}_d + (1-O) * \mathbf{C}^{pbr}_s + O * \mathbf{C}_{ind}
\end{aligned}
\label{eqn:opacity}
\end{equation}
More details are discussed in the supplementary.

% \boldparagraph{Second Split-sum} During relighting stage, since the indirect illumination also depends on the environment map, naively reusing the indirect MLP for unseen light condition results in artifacts. Therefore, we propose a novel second split-sum (Figure~\ref{fig:illumination} (c)) for relighting. Different from the ray tracing in Monte Carlo rendering, we trace secondary rays under the split-sum rendering equation. We compute the expected intersection $\mathbf{\hat{x}}$ and secondary direction $\mathbf{\hat{d}}$. Given the illumination term in Equation \ref{eqn:rendering_equation}, we further divide the incident radiance into direct and indirect illumination. As stated in \cite{liang2023envidr}, the indirect illumination has a strong effect only with a small roughness value and specular part of the light. So we set a threshold $\rho_t$ and only trace secondary rays for pixels with roughness value smaller than it:
\noindent \textbf{Second Split-sum}. During the relighting stage, we cannot reuse the learned indirect illumination MLP for unseen light conditions.  Therefore, we propose a novel second split-sum (\cref{fig:illumination} (c)) for relighting. Unlike Monte Carlo rendering, which requires tracing multiple rays, we trace a single ray and use the split-sum to approximate integration over the reflection lobe. We compute the expected intersection $\mathbf{\hat{x}}$ and secondary direction $\mathbf{\hat{d}}$. Given the illumination term in \cref{eqn:rendering_equation}, we further divide the incident radiance into direct and indirect illumination. As shown in \cite{liang2023envidr}, the indirect illumination mainly has a strong effect with a small roughness value on the specular part of the light. So we set a threshold $\rho_t$ and only trace secondary rays for pixels with smaller roughness values:
\begin{equation}
\begin{aligned}
L_i  =& \mathbbm{1}[\rho > \rho_t] L_{dir} \\
&+ \mathbbm{1}[\rho \leq \rho_t]((1-O) * L_{dir} + O * L_{ind})
\end{aligned}
\label{eqn:second-split-sum}
\end{equation}
%
% where $\mathbbm{1}[.]$ is an indicator function with output one if the condition holds and zero if the condition does not hold. We plug this equation into the split-sum equation, and derive the following adjustment equation for the light integral during relighting:
where $\mathbbm{1}[.]$ is an indicator function that returns one if the condition holds and zero otherwise. Plugging this equation into the split-sum equation (in the appendix), we derive the following adjustment equation for the light integral during relighting:
\begin{equation}
\begin{aligned}
\mathbf{l}^s_{relight}  \approx & (\mathbbm{1}[\rho > \rho_t] + \mathbbm{1}[\rho \leq \rho_t] * (1-O)) * \mathbf{l}^s \\
& + \mathbbm{1}[\rho \leq \rho_t]*O*L_{ind}(\mathbf{\hat{x}}, \mathbf{\hat{d}})
\end{aligned}
 \label{eqn:second-split-sum-simplified}
\end{equation}
%
% where $\rho$ and $\mathbf{l}^s$ are computed by volume rendering the per-sample roughness $\rho_i$ and per-sample light integral $\mathbf{l}^s_i$. Compared with the Monte Carlo sampling approach, we only sample one secondary ray with an additional split-sum approximation to compute the indirect illumination. We give the full proof in Sec. \ref{sec:appendix_secondproof}. Then we apply ray marching along the secondary ray $\{\mathbf{\hat{x}}, \mathbf{\hat{d}}\}$ and computing the expected second surface intersection $\mathbf{\hat{x'}}$. The indirect light $L_{ind}$ is computed as the following equation with Equation \ref{eqn:pbr_split-sum}.
where $\rho$ and $\mathbf{l}^s$ are computed by volume rendering the per-sample roughness $\rho_i$ and per-sample light integral $\mathbf{l}^s_i$.
%Compared with the Monte Carlo sampling approach, we only sample one secondary ray with an additional split-sum approximation to compute the indirect illumination.
We give the full derivation in supplementary. Applying ray marching along the secondary ray $\{\mathbf{\hat{x}}, \mathbf{\hat{d}}\}$ the expected second surface intersection $\mathbf{\hat{x'}}$ is computed. The indirect light $L_{ind}$ is evaluated by following \cref{eqn:pbr_split-sum}:
\begin{equation}
\begin{aligned}
L_{ind}(\mathbf{\hat{x}}, \mathbf{\hat{d}}) = \mathbf{c}^{pbr}_d(\mathbf{\hat{x'}}, \mathbf{\hat{d}}) + \mathbf{c}^{pbr}_s(\mathbf{\hat{x'}}, \mathbf{\hat{d}})
\end{aligned}
\end{equation}

\noindent \textbf{Training Losses}. To improve the geometry quality, we further apply the progressive hash grid and finite differences for SDF gradient estimation in the Eikonal \cite{li2023neuralangelo} and curvature \cite{rosu2023permutosdf} losses. Finally, we compute the photometric loss, and define the whole optimization loss function:
\begin{equation}
\begin{aligned}
&\mathcal{L}_{c} = ||\mathbf{C}-\mathbf{C}_{gt}||^2_2; \ \ \mathcal{L}_{pbr} = ||\mathbf{C}^{pbr} - \mathbf{C}_{gt}||^2_2\\
&\mathcal{L}_{stage\ 1} = \lambda_c\mathcal{L}_{c} + \lambda_{eik}\mathcal{L}_{eik} + \lambda_{curv}\mathcal{L}_{curv} \\
&\mathcal{L}_{stage\ 2} = \lambda_c\mathcal{L}_{c} + \lambda_c\mathcal{L}_{pbr} + \lambda_{eik}\mathcal{L}_{eik} + \lambda_{curv}\mathcal{L}_{curv}
\end{aligned}
\label{eqn:stage1_loss}
\end{equation}

\section{Experiments}
\begin{table*}[ht]
    \centering
    \resizebox{\linewidth}{!}{
        \begin{tabular}{@{}c c c c c c c c c c ccc c ccc c ccc c c@{}}
            \toprule
            \multirow{2}{*}{Method} & & \multirow{2}{*}{PB} & & 
            \multirow{2}{*}{E2ER} & & 
            \multicolumn{1}{c}{Normal} &  & \multicolumn{1}{c}{Roughness} & & \multicolumn{3}{c}{Albedo} & & \multicolumn{3}{c}{Novel View Synthesis} & & \multicolumn{3}{c}{Relighting} 
            & &  \multirow{2}{*}{Runtime}  \\ \cline{7-7} \cline{9-9} \cline{11-13} \cline{15-17} \cline{19-21}
            & & & & & &  MAE $\downarrow$ & & PSNR $\uparrow$ & & PSNR $\uparrow$ & SSIM $\uparrow$ & LPIPS $\downarrow$ & & PSNR $\uparrow$ & SSIM $\uparrow$ & LPIPS $\downarrow$ & & PSNR $\uparrow$ & SSIM $\uparrow$ & LPIPS $\downarrow$ \\ \hline
            
            NDR~\cite{munkberg2022extracting} & & \cmark & & \cmark & & 5.877 & & 23.452 & & 22.340 & 0.921 & 0.102 & & 29.312 & 0.939 & 0.062 & & 21.057 & 0.817 & 0.145 
            & & \cellcolor{orange} 2 hrs\\

            NDRMC~\cite{hasselgren2022shape}& & \cmark & & \cmark & & 4.767 & & 23.782 & & 22.602 & 0.935 & 0.084 & & 26.280 & 0.914 & 0.100 & & 25.149 & 0.890 & 0.100
            & & \cellcolor{orange} 2 hrs\\

            NMF~\cite{mai2023neural}& & \cmark & & \cmark & & 3.110 & & 22.099 & & 14.238 & 0.755 & 0.211 & & 30.516 & 0.950 & 0.037 & & 21.590 & 0.904 & \cellcolor{lightyellow}0.072
            & &  5 hrs\\

            ENVIDR~\cite{liang2023envidr} & & \xmark & & \xmark & & 3.330 & & - & & - & - & - & & \cellcolor{yellow}31.500 & 0.936 & 0.072 & & 25.524 & 0.907 & 0.080 
            & & 6  hrs*\\

            GShader~\cite{jiang2023gaussianshader} & & \xmark & & \cmark & & 6.179 & & - & & - & - & - & & 29.598 & 0.934 & 0.075 & & 23.718 & 0.910 & 0.081 
            & & \cellcolor{tablered}1.5  hrs\\

            NeRO~\cite{liu2023nero}& & \cmark & & \xmark & & 3.694 & & 20.237 & & 22.907 & 0.925 & 0.097 & & 30.331 & 0.958 & 0.054 & & \cellcolor{orange} 28.116 & \cellcolor{lightyellow}0.980 & \cellcolor{yellow}0.031
            & & $>$  10 hrs\\ \hline

            Ours (w.o. info-share) & & \cmark & & \cmark & &
            \cellcolor{lightyellow}2.191& & 
            \cellcolor{lightyellow}28.216 & &
             \cellcolor{yellow}25.634 & \cellcolor{yellow}0.931& \cellcolor{yellow}0.080& & \cellcolor{lightyellow}31.220 & \cellcolor{lightyellow}0.988&  \cellcolor{yellow}0.016 & &
            27.252 & \cellcolor{yellow}0.984 & 
            \cellcolor{yellow}0.031 & &
             \cellcolor{orange}2 hrs  \\

             Ours (w.o. indirect MLP) & & \cmark & & \cmark & &
           \cellcolor{yellow}1.947 & & 
            \cellcolor{yellow}28.310 & &
             \cellcolor{lightyellow}25.170 & \cellcolor{lightyellow}0.925 & \cellcolor{lightyellow}0.084 & & 30.451 & \cellcolor{yellow}0.990 &  \cellcolor{lightyellow}0.018 & &
            \cellcolor{yellow}27.300 & \cellcolor{yellow}0.984 & 
            \cellcolor{orange}0.030 & &
             \cellcolor{yellow} 2.5 hrs  \\

            Ours (with Monte-Carlo) & & \cmark & & \cmark & &
            2.540& & 
            25.865& &
            22.800 & 0.881 & 0.105 & & 29.482 & 0.925 & 0.072  & &
            23.766 & 0.876 & 0.126 & &
            6 hrs  \\

            Ours (w.o. sec. split-sum)& & \cmark & & \cmark & & -
           & & -& &- & - & - & & - &  -&  - & & 27.230 & \cellcolor{yellow}0.984 & \cellcolor{orange}0.030 & &
            -  \\

            Ours(full model)& & \cmark & & \cmark & &
            \cellcolor{orange}1.869 & & 
            \cellcolor{orange}29.460 & &
            \cellcolor{orange}25.839 & \cellcolor{orange}0.941 & \cellcolor{orange}0.078 & & \cellcolor{orange}32.050 & \cellcolor{orange}0.993&  \cellcolor{orange}0.012 & &
            \cellcolor{yellow}28.096 & \cellcolor{orange}0.985 & 
            \cellcolor{orange}0.030 & &
            \cellcolor{yellow} 2.5 hrs  \\
            
            Ours (full model, 5 hrs) & & \cmark & & \cmark & &
            \cellcolor{tablered}1.780 & & 
            \cellcolor{tablered}30.076 & &
            \cellcolor{tablered}25.958 & \cellcolor{tablered}0.943 & \cellcolor{tablered}0.052 & & \cellcolor{tablered}32.493 & \cellcolor{tablered}0.994&  \cellcolor{tablered}0.011 & &
            \cellcolor{tablered}28.215 & \cellcolor{tablered}0.986 & 
            \cellcolor{tablered}0.028 & &
             5 hrs  \\ 
            \bottomrule
        \end{tabular}
    }
    \caption{\textbf{Quantitative comparisons on the shiny inverse rendering synthetic dataset.} PB: Physically Based. E2ER: End to End Relightable. ENVIDR runtime: ENVIDR requires 2 additional hours for distillation. Our results have 
    significantly outperformed the baseline methods by reconstructing more accurate geometry, and material, thus achieving high-quality novel view synthesis and relighting results. }
    \label{tab:main_cmp}
\end{table*}
\begin{figure*}[h!]
    \centering
    \setlength\tabcolsep{1pt}
    \begin{tabular}{ccccccc}
        
        \includegraphics[height=0.12\textwidth]{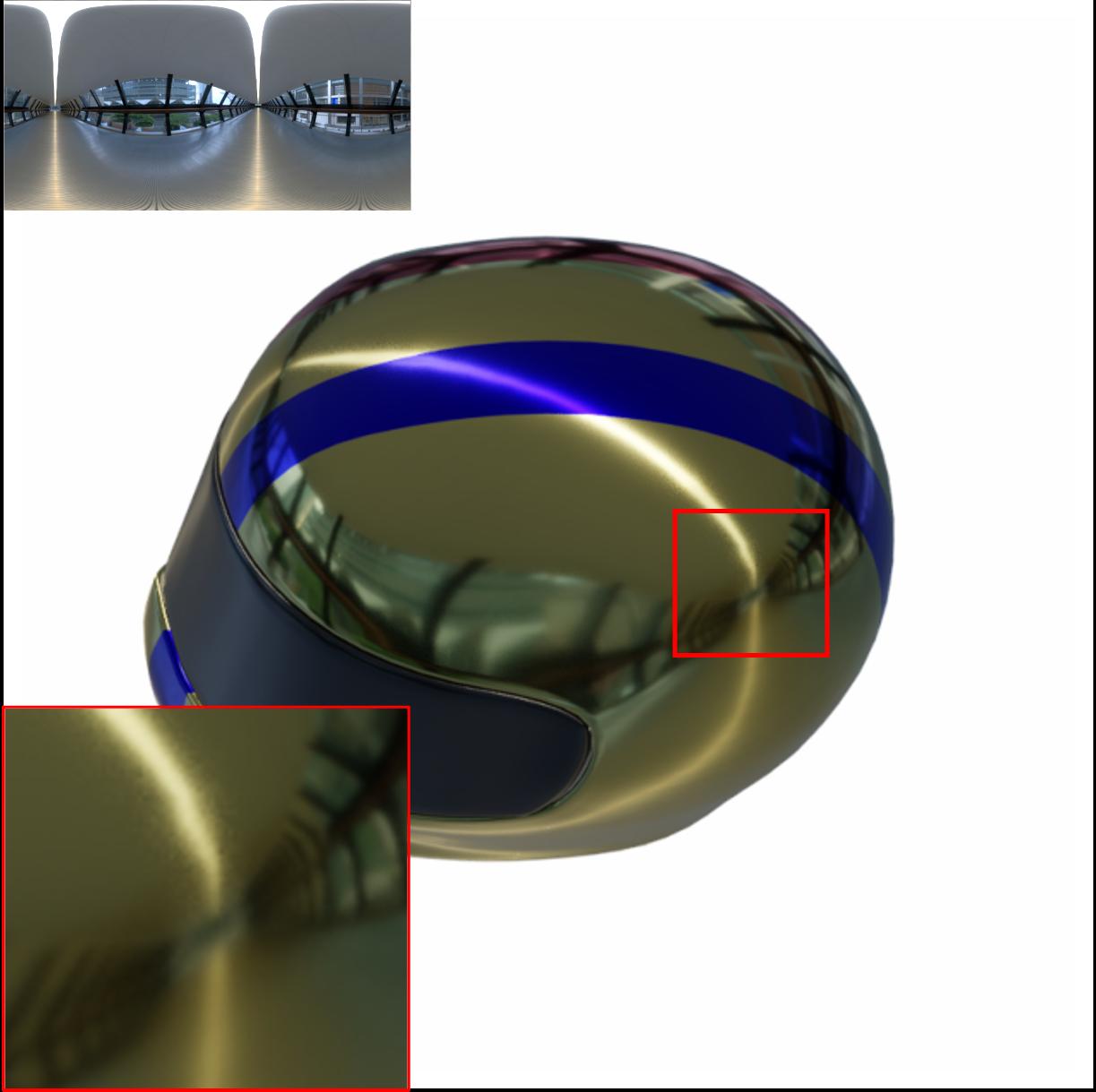} &
        \includegraphics[height=0.12\textwidth]{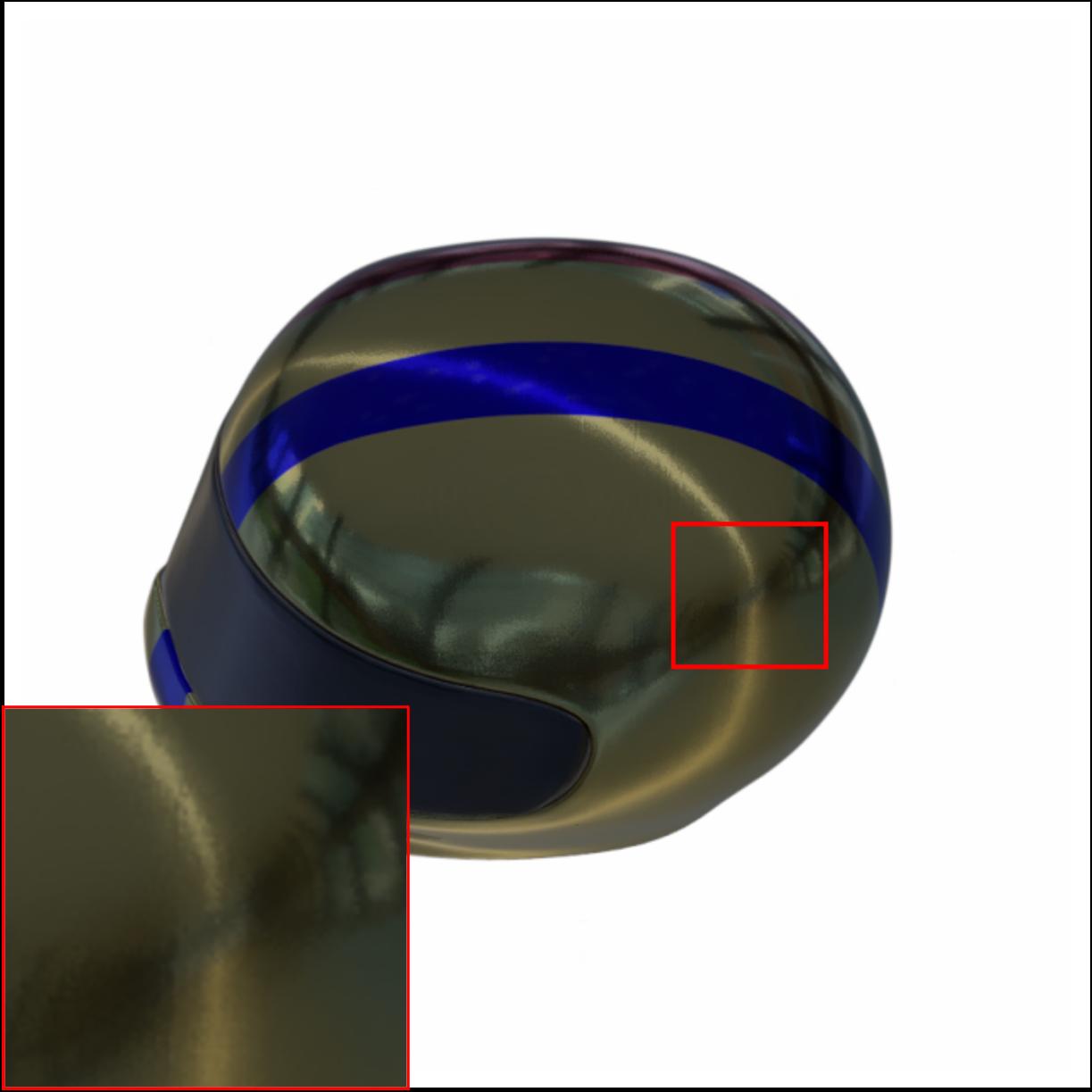} &
        \includegraphics[height=0.12\textwidth]{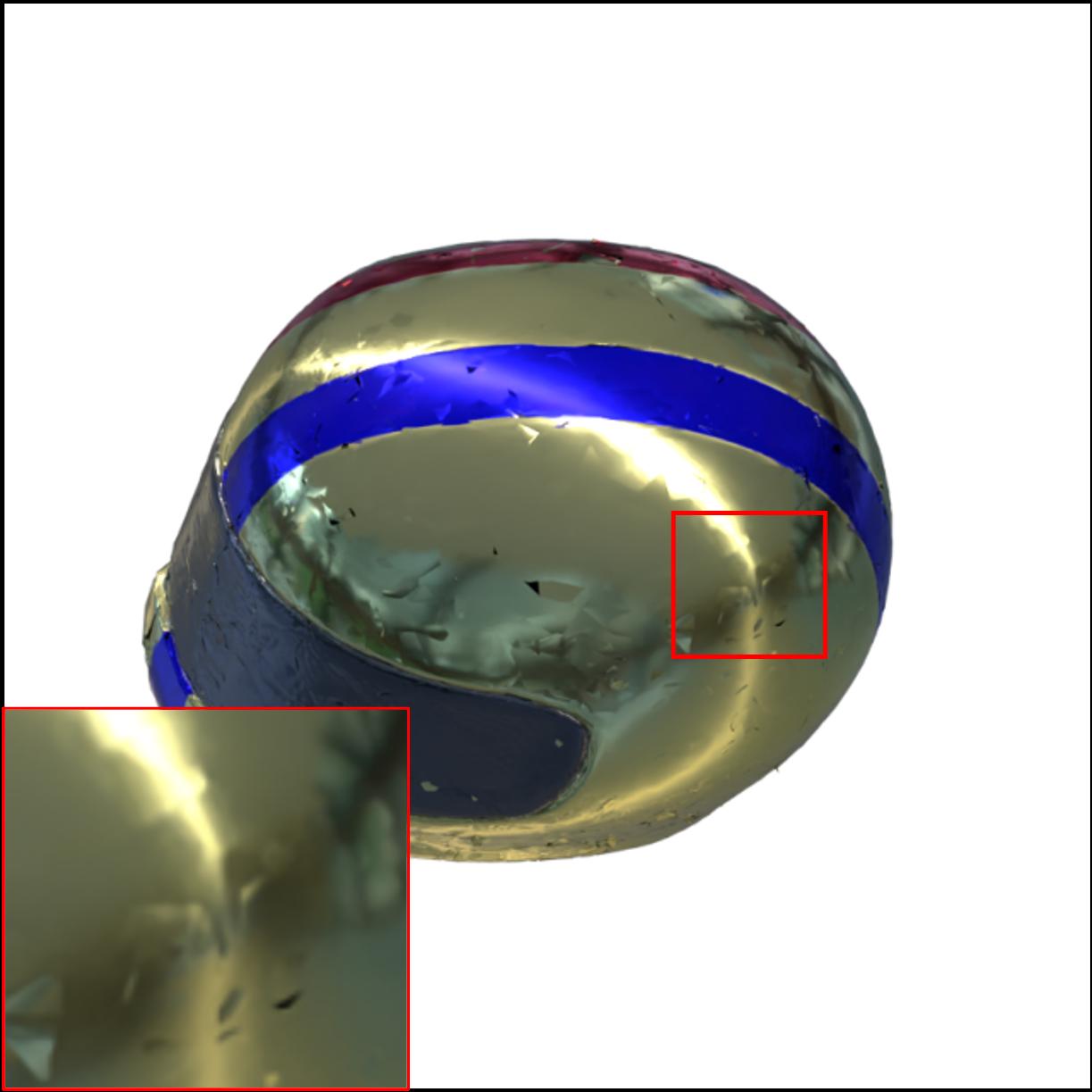} &
        \includegraphics[height=0.12\textwidth]{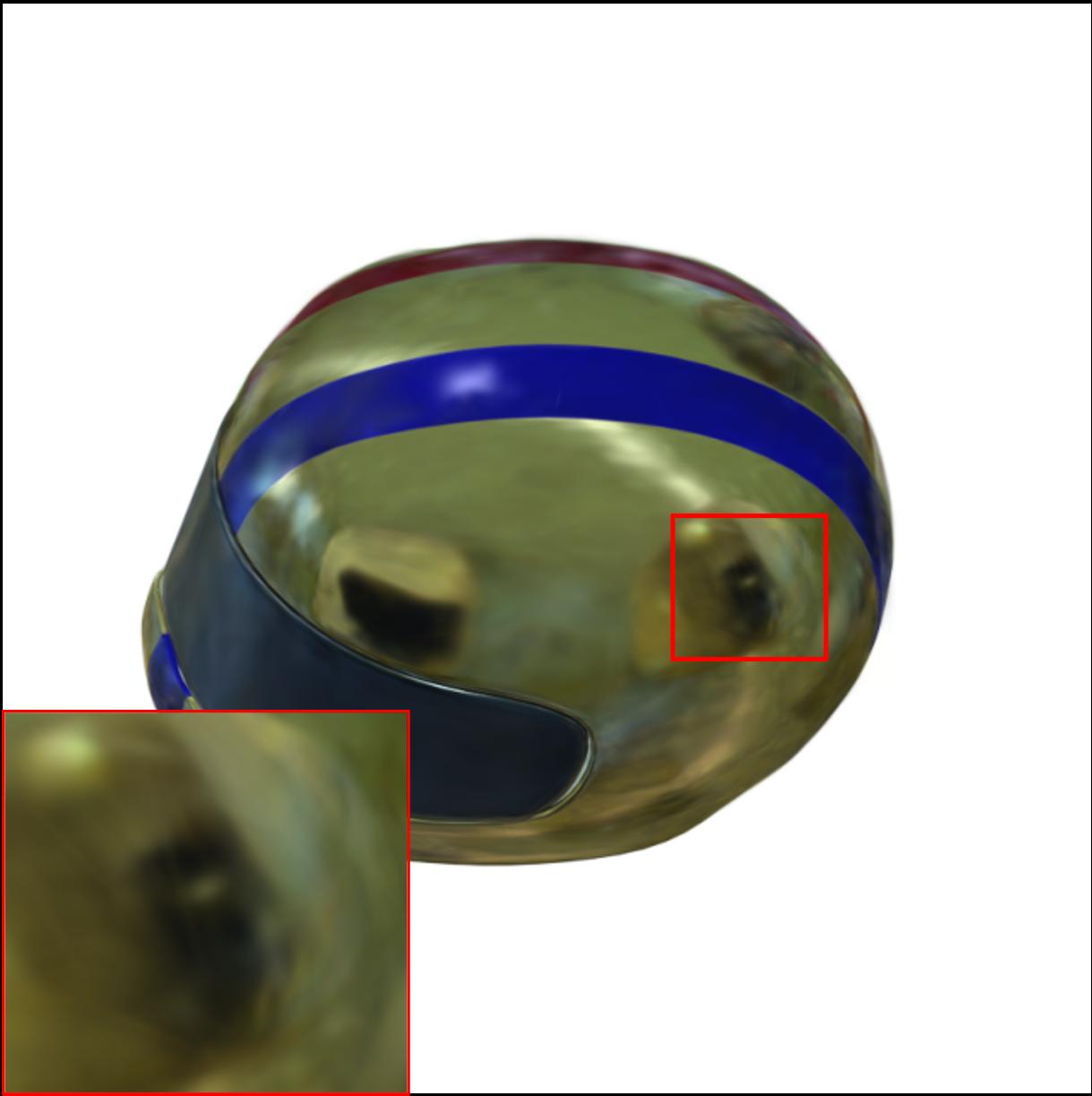} &
        \includegraphics[height=0.12\textwidth]{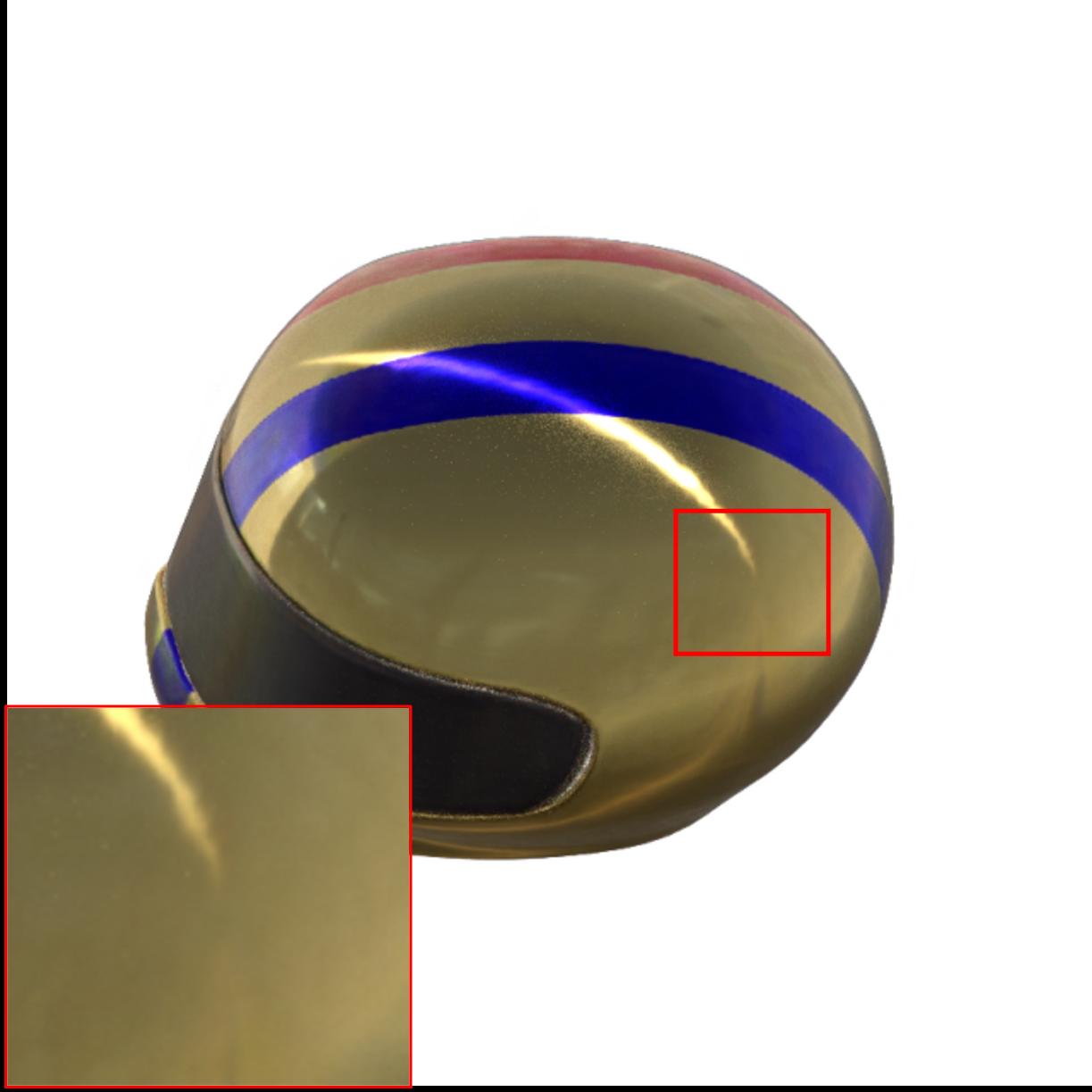} &
        \includegraphics[height=0.12\textwidth]{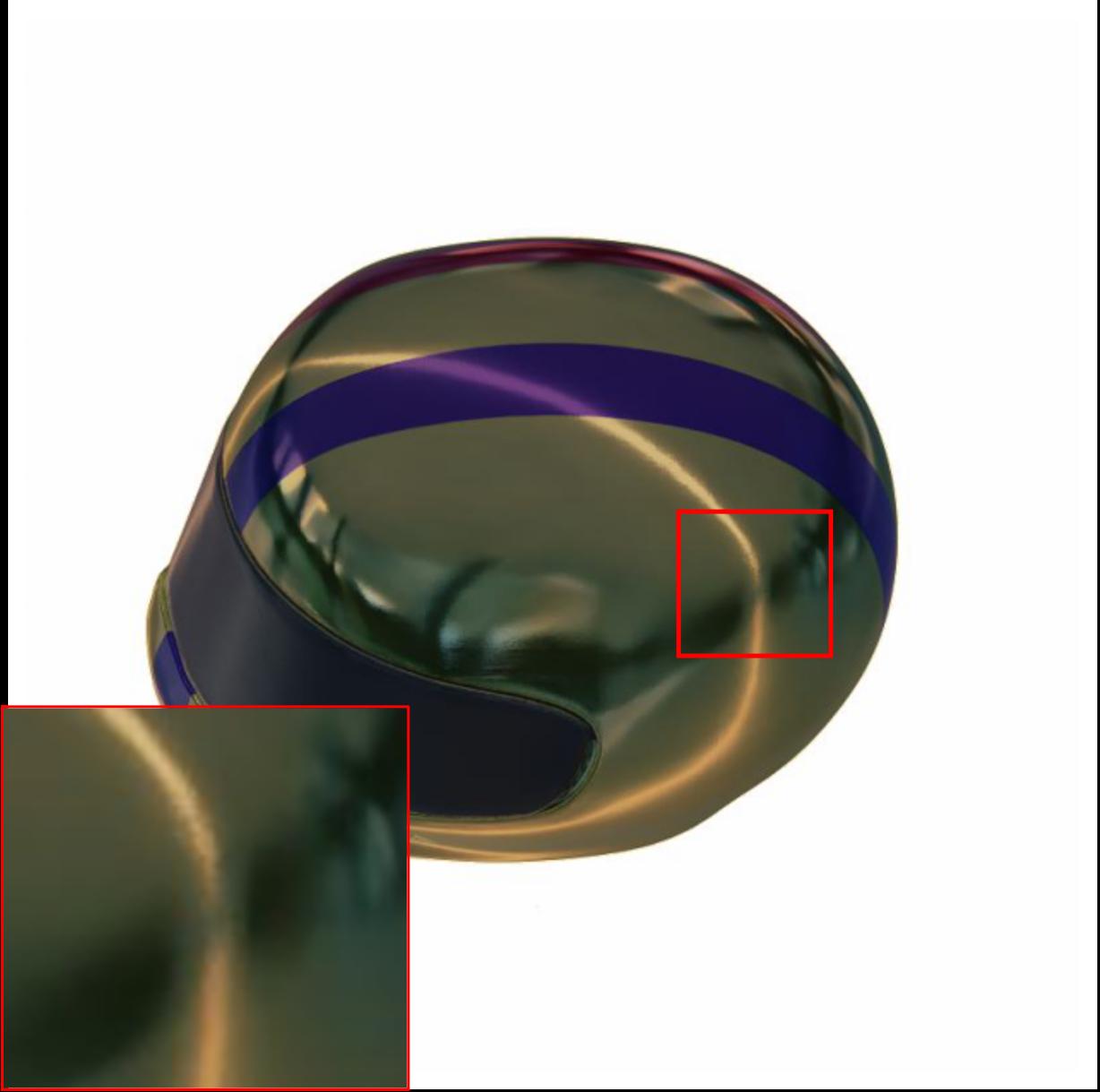}&
        \includegraphics[height=0.12\textwidth]{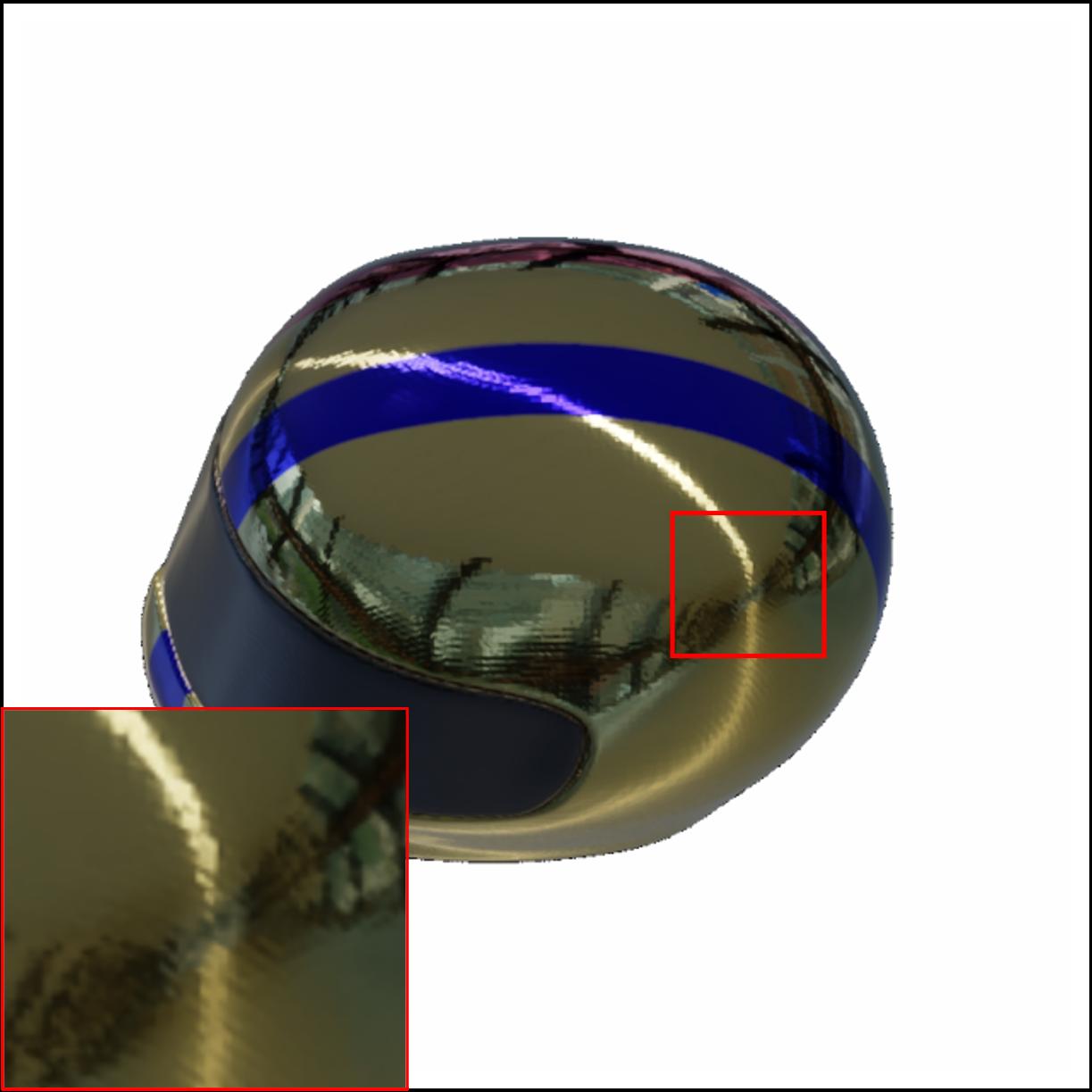}\\

         \includegraphics[height=0.12\textwidth]{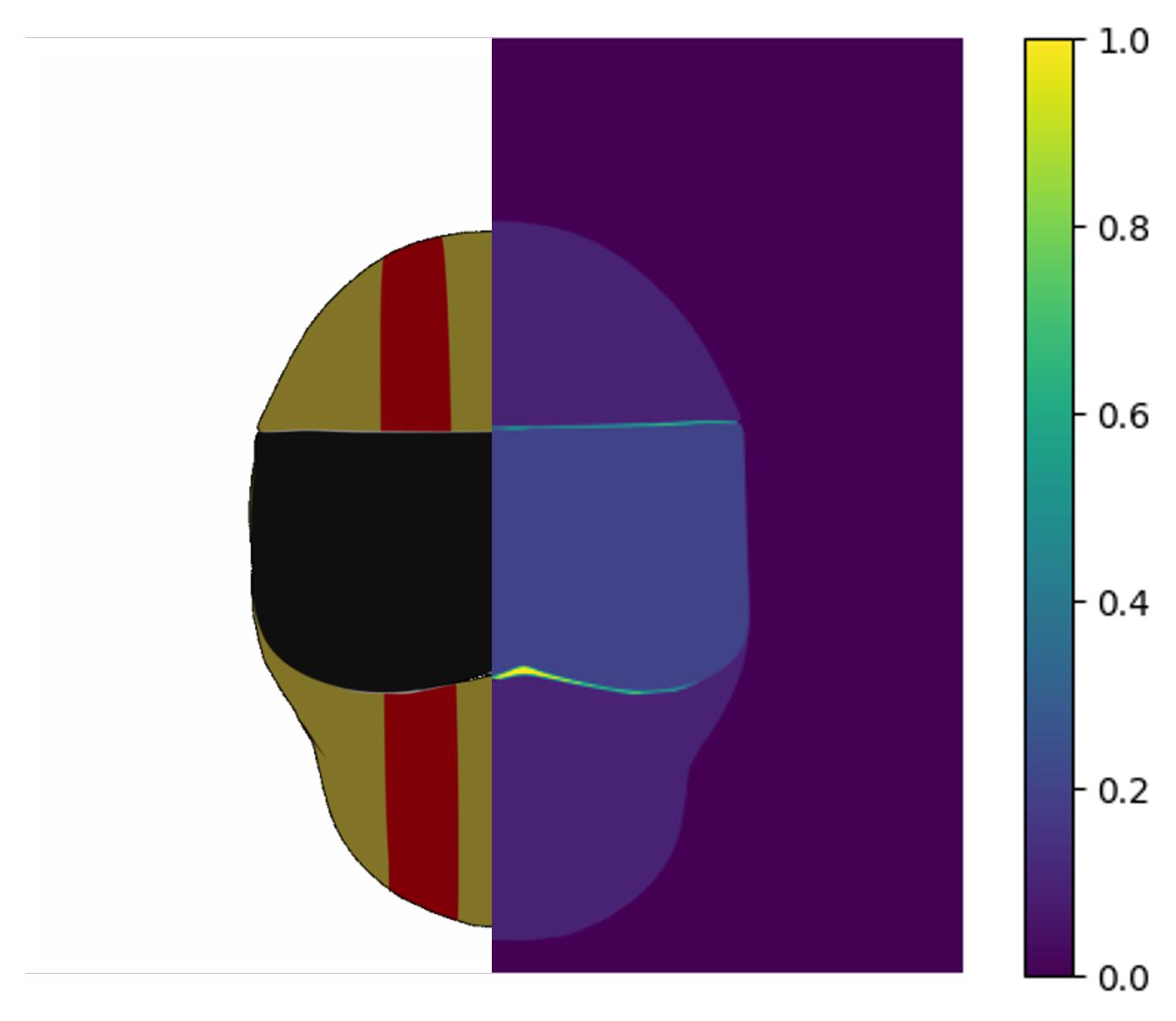} &
        \includegraphics[height=0.12\textwidth]{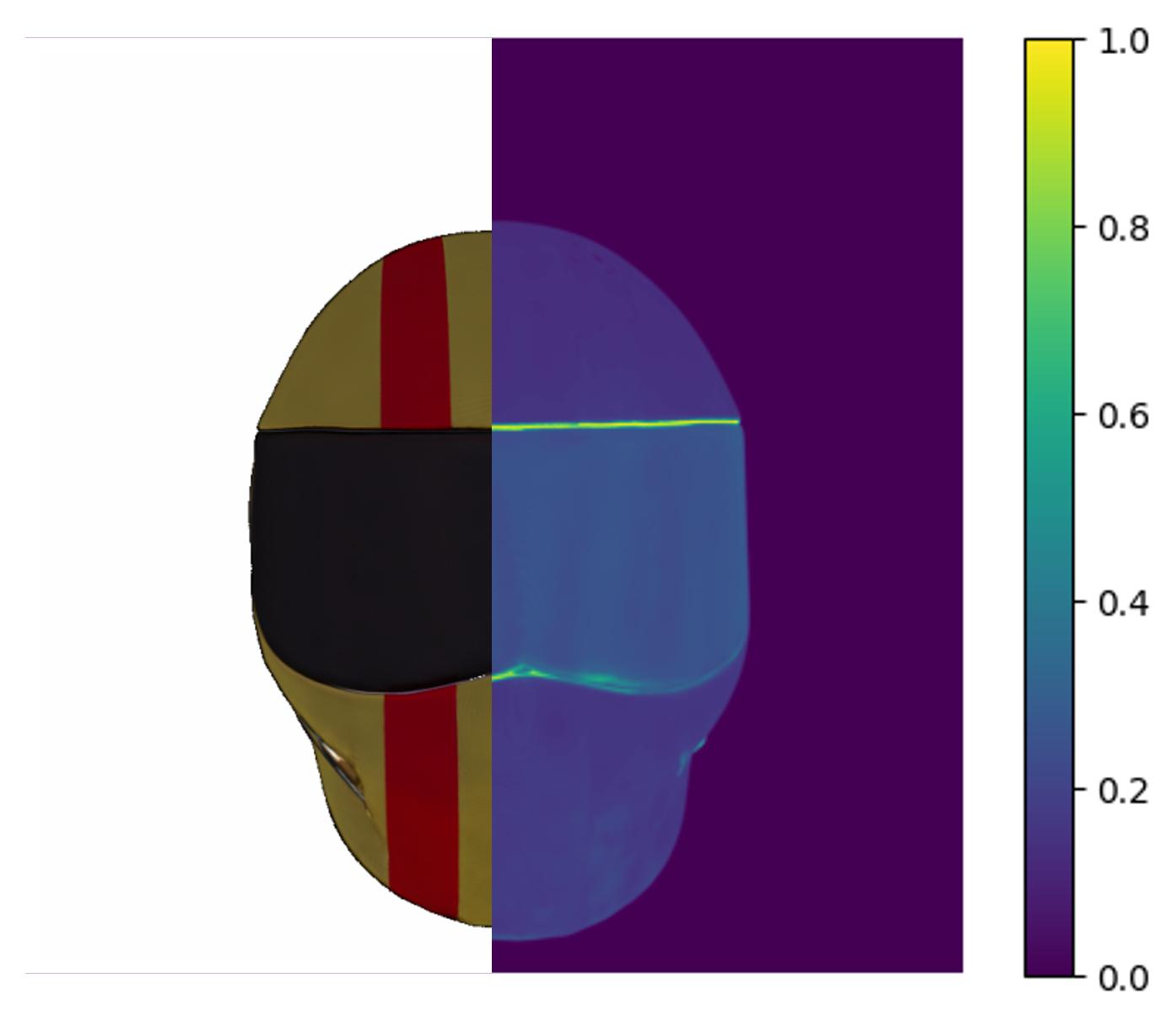} &
        \includegraphics[height=0.12\textwidth]{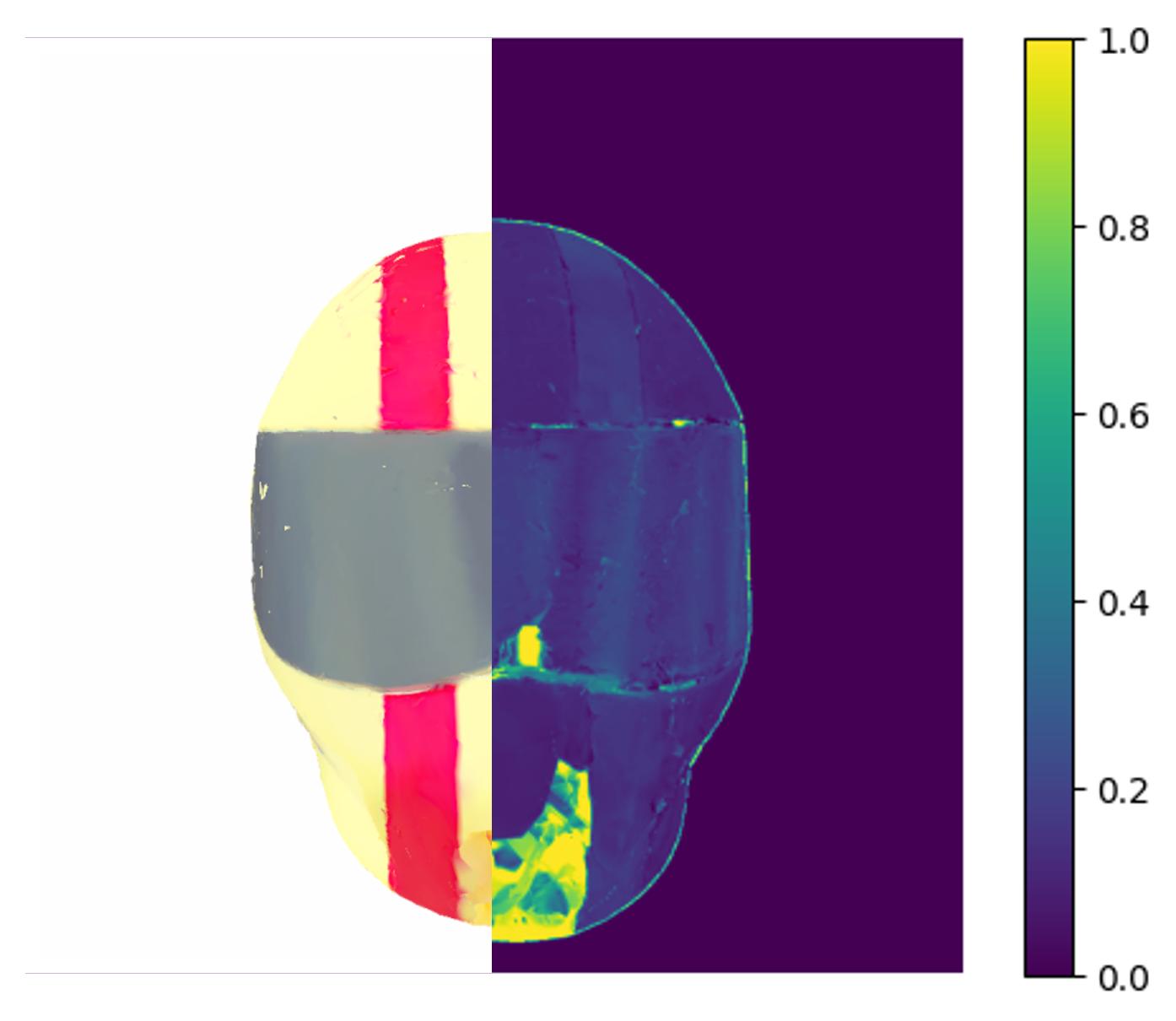} &
        \includegraphics[height=0.12\textwidth]{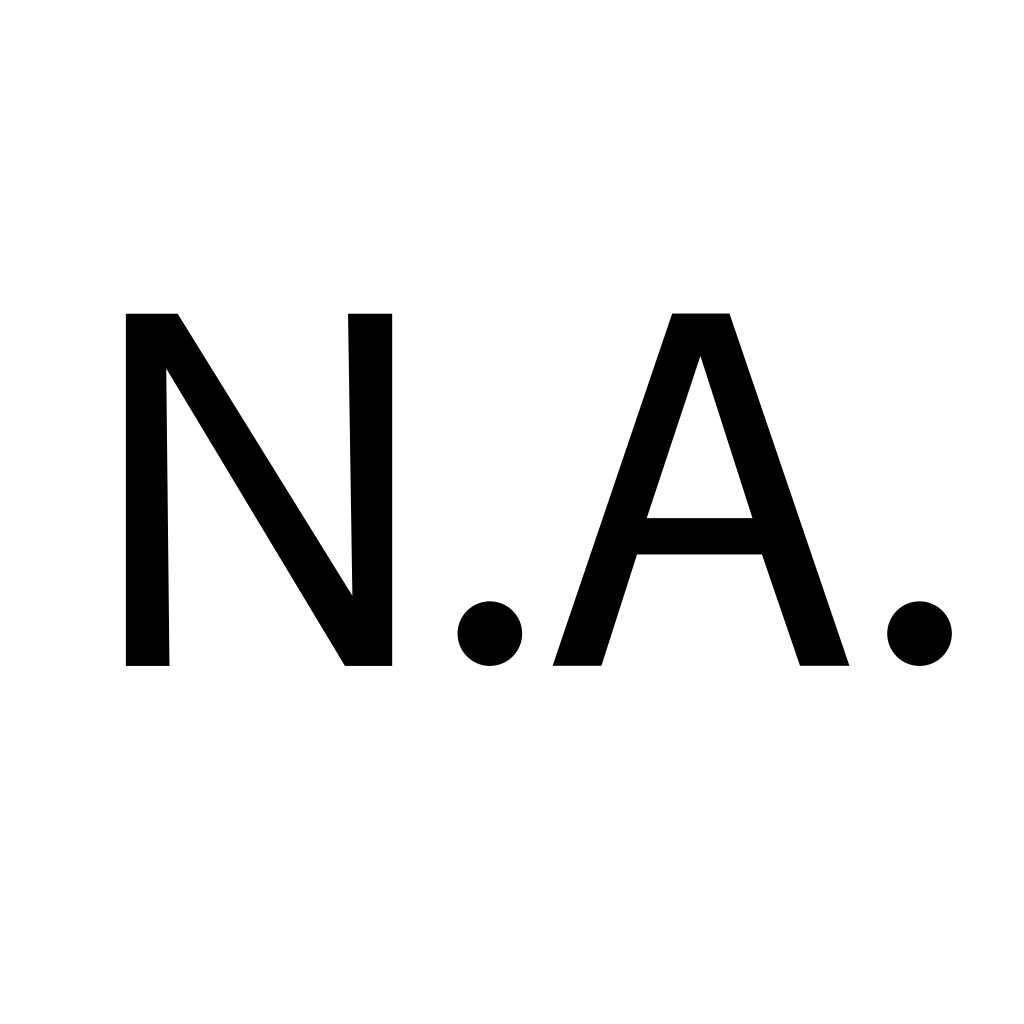} &
        \includegraphics[height=0.12\textwidth]{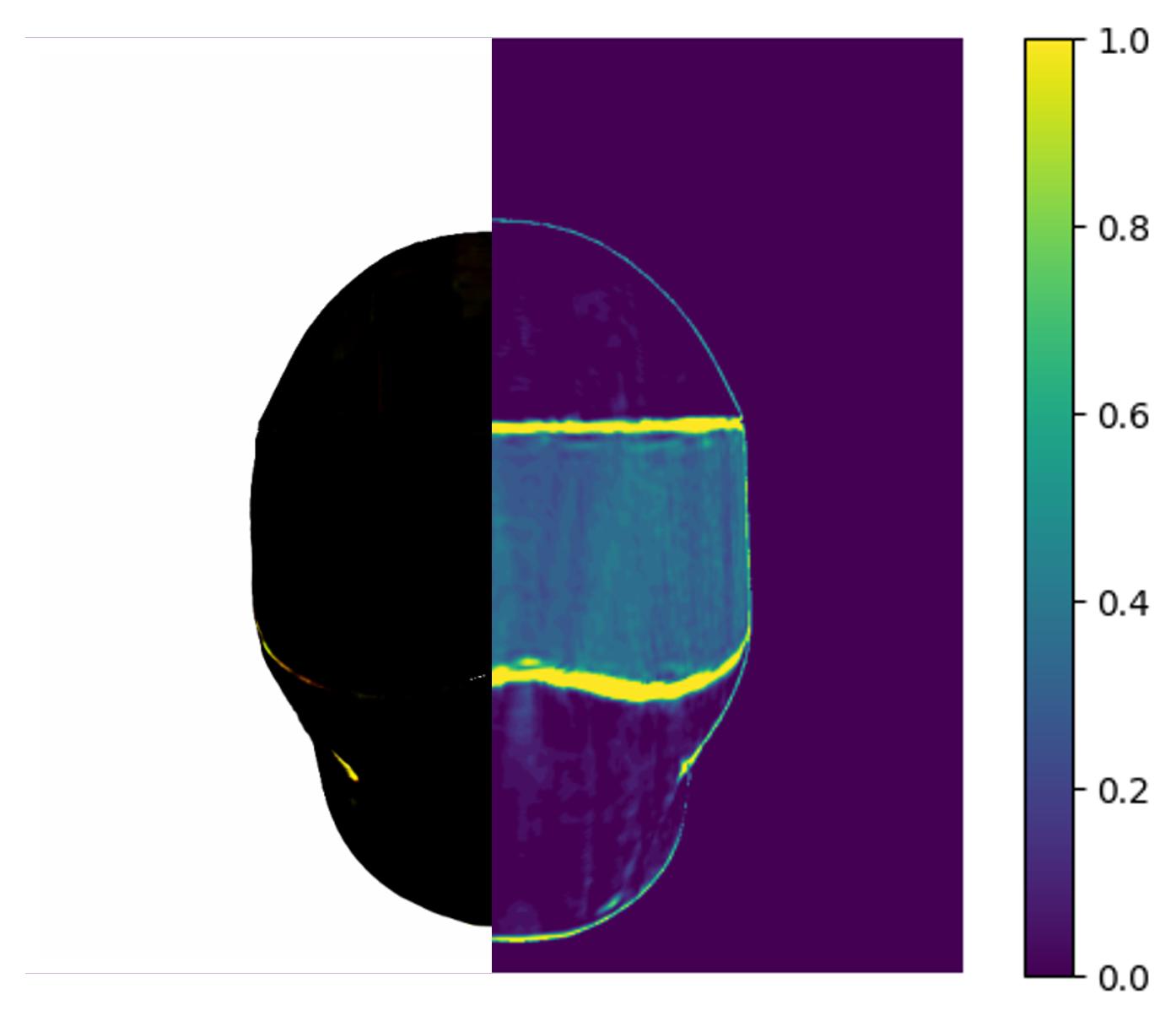} &
        \includegraphics[height=0.12\textwidth]{figures/helmet/NA_cap_icon.svg.png}  &
        \includegraphics[height=0.12\textwidth]{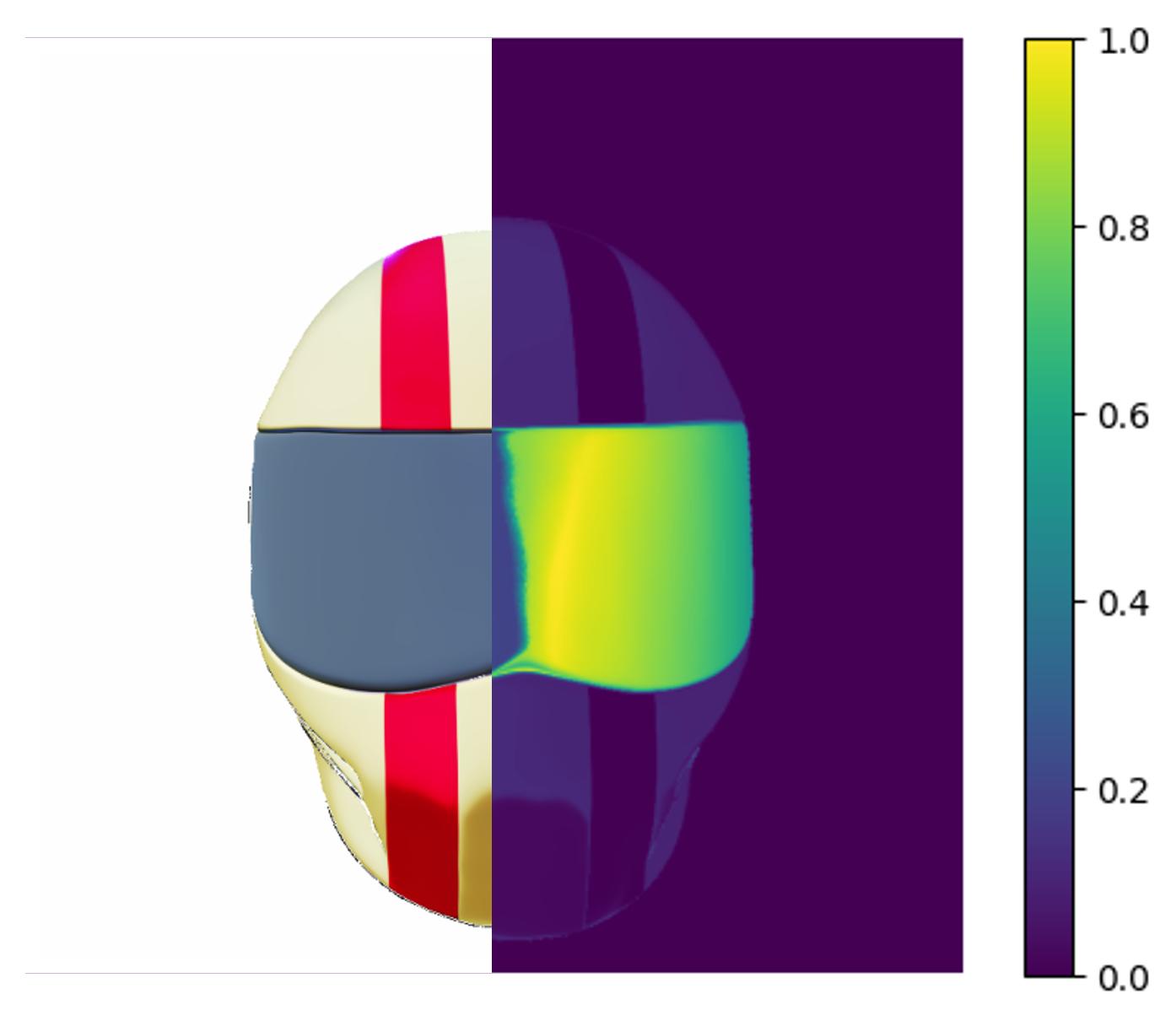}\\

        GT & Ours & NDRMC & GShader & NMF& ENVIDR & NeRO\\
    \end{tabular}
    \caption{\textbf{Qualitative comparisons on the helmet scene. From top to bottom: relighting, material.} We can observe that other methods either have blurry, color-shifted results or aliasing, noisy effect under unseen illumination. And our material estimation also outperform other baselines. }
    \label{fig:helmet}
    \vspace{-1.5em}
\end{figure*}

\subsection{Experiment Protocol}

\noindent{\textbf{Baselines}.} We compare our method with baselines focusing on inverse rendering for glossy objects. These methods include DMTet-based methods NDR~\cite{munkberg2022extracting} and NDRMC~\cite{hasselgren2022shape}, NeRF-based method NMF~\cite{mai2023neural}, neural SDF-based methods NeRO~\cite{liu2023nero}, ENVIDR~\cite{liang2023envidr}, and 3DGS-based method GaussianShader~\cite{jiang2023gaussianshader}. Note that not all of these methods are end-to-end relightable. NeRO \cite{liu2023nero} and ENVIDR \cite{liang2023envidr} use neural representation for the environment map, thus their relighting requires additional neural network training per environment map. We build our own shiny inverse rendering dataset as discussed in the supplementary material, and use NeRO~\cite{liu2023nero} real dataset to evaluate the models.

\noindent{\textbf{Evaluation Protocol}.} % We compare our method and the baselines in the following way. 
During relighting evaluation, for NeRO~\cite{liu2023nero}, NDR~\cite{munkberg2022extracting}, and NDRMC~\cite{hasselgren2022shape}, we use Blender with SPP=4096 and 4 bounces of lights. For ENVIDR~\cite{liang2023envidr}, we first pre-train and distill the environment maps to the MLP. Due to the inevitable ambiguity between the albedo and the environment map, we align each channel of the predicted albedo, roughness, and relighting with the ground truth before evaluation. We use Peak Signal-to-Noise Ratio (PSNR), Structural Similarity Index Measure (SSIM) \cite{wang2004image}, and Learned Perceptual Image Patch Similarity (LPIPS) \cite{zhang2018unreasonable} as metrics to evaluate the novel view synthesis, albedo, and relighting. Roughness is evaluated using only PSNR, and the normals are evaluated using Mean Angle Error (MAE). For all the values, we calculate the average among all the test views in five scenes. For the relighting metrics, we calculate the average under all the different environment maps. 

% \noindent{\textbf{Implementation Details}}
% Our code is built on the instant-nsr-pl codebase \cite{yuan2022instant}. We use a 512-resolution progressive hash grid with 16 levels. The geometry MLP has $2$ layers with $128$ neurons. The diffuse, specular, secondary MLP has $4$ layers with $128$ neurons. And the roughness and blending MLP has $2$ layers with $128$ neurons. The resolution of the environment map is $6\times512\times512\times3$. We start from the 4-th level of the hash grid and increase by 1 level for every 500 iterations. Regarding the hyperparameters, we use $\lambda_c = 10$, $\lambda_{eik} = 0.1$, $\lambda_{curv} = 1$. We use Adam optimizer with $\beta_1 = 0.9$, $\beta_1 = 0.999$, and $\epsilon = 10^{-12}$. The first stage of our method is trained for 10k
% iterations and the second stage for 70k iterations. All experiments are conducted on a single RTX 3090Ti GPU. 

\subsection{Results}

\noindent{\textbf{Shiny Inverse Rendering}}. We first compare the quantitative metrics on our shiny inverse rendering dataset. We categorize different methods according to whether they are physically-based models with valid BRDF modeling, and whether they are end-to-end relightable or not. We only evaluate the roughness and albedo metrics for physically-based models. The category and metrics are listed in \Cref{tab:main_cmp}. 
Our method shows superior performance compared with previous methods on all metrics. We also show the qualitative results for relighting and material estimation in \cref{fig:helmet}. More results can be found in the supplementary material. 

\textit{Relighting}. As shown in supplementary \cref{fig:relight}, we visualize the relighting results under different unseen light conditions during training. We can observe that NDRMC~\cite{hasselgren2022shape} and GShader~\cite{jiang2023gaussianshader} are limited by their low-quality geometry reconstructions. NMF \cite{mai2023neural} generates excessive noise due to the Monte Carlo Sampling. ENVIDR~\cite{liang2023envidr} generates blurry shifted colors due to the distillation of the HDR environment map to the MLP. NeRO~\cite{liu2023nero} causes aliasing artifacts when exporting to Blender, which is a required step of their model for relighting. 

\textit{Material \& geometry reconstruction}. The normal estimation is displayed in supplementary \cref{fig:normal}, while the albedo and roughness are in supplementary \cref{fig:material}. Our method has the best quality on the material estimation thanks to our information-sharing architecture. For normal reconstruction, we can observe that NDRMC~\cite{hasselgren2022shape} and GShader~\cite{jiang2023gaussianshader} produce artifacts due to limitations of surface-based-rendering/3DGS on geometry reconstruction. NMF~\cite{mai2023neural} has small bumps due to the instability of volume density field representation and Monte Carlo sampling. For SDF-based methods, ENVIDR~\cite{liang2023envidr} fails to reconstruct the geometry when indirect illumination appears, and NeRO~\cite{liu2023nero} tends to reconstruct the over-smoothed surface. 

% \noindent{\textbf{NeRO Synthetic}}

% \noindent{\textbf{TensoIR Synthetic}}

\noindent{\textbf{Real Dataset}}. Further evaluation is carried out on the Glossy Real dataset from NeRO~\cite{liu2023nero}. Following 
NeRO, we qualitatively show novel view synthesis, geometry reconstruction, and relighting results. We also quantitatively evaluate the Chamfer distance of the extracted meshes. 

\noindent  The per-scene Chamfer distance scores are listed in \Cref{tab:real}. The results of extracted meshes, novel-view-synthesis, and relighting are shown in supplementary \cref{fig:glossy_real}. We achieve comparable performance to NeRO in geometry reconstruction. Our method falls short in material estimation because the assumption of a distant environment map is violated. Please refer to the appendix for a more detailed discussion.

% Uncomment if we managed to handle background
% Further evaluation is carried out on the Glossy Real dataset from NeRO~\cite{liu2023nero}. Following NeRO, we qualitatively show novel-view-synthesis, geometry reconstructions and relighting results. We also quantitatively evaluate the Chamfer distance of the extracted meshes. An additional density-based radiance field is used to model the background. To overcome unbalanced convergence of foreground and background radiance field, we apply several strategies to stabilize the training in early stage. Please refer to section 

% \input{figures/glossy_real/glossy_real}
\begin{table}[]
    \centering
    \tiny
    \resizebox{\linewidth}{!}{
    \begin{tabular}{cccccl}
          &NDR$^*$ & NDRMC$^*$ & NeuS & NeRO  & Ours(4hrs)$^*$\\
    \midrule
    Bunny  & 0.0047  &   0.0042  &  \cellcolor{yellow}0.0022    &   \cellcolor{tablered}0.0012    & \cellcolor{orange}0.0018\\ % rabbit4 
    Coral  & 0.0025  &   \cellcolor{yellow}0.0022  &  \cellcolor{orange}0.0016    &   \cellcolor{tablered}0.0014    & \cellcolor{orange}0.0016\\ % coral
    Maneki & 0.0148  &   0.0117  &  \cellcolor{yellow}0.0091    &   \cellcolor{orange}0.0024    & \cellcolor{tablered}0.0023\\ % zhaocai5
    Bear   & 0.0104  &   0.0118  &  \cellcolor{orange}0.0074    &   \cellcolor{tablered}0.0033    & \cellcolor{yellow}0.0095\\ % bear
    Vase   &  0.0201 &   \cellcolor{yellow}0.0058  &  0.0101    &   \cellcolor{tablered}0.0011    & \cellcolor{orange}0.0012\\ % vase2
    \midrule
    Avg.   &  0.0105 & 0.0071    &   \cellcolor{yellow}0.0061   &   \cellcolor{tablered}0.0019    & \cellcolor{orange}0.0032\\ 
    \end{tabular}
    }
    \caption{\textbf{Geometry reconstruction quality in Chamfer Distance (CD$\downarrow$) on the Glossy-Real dataset.} NDR, NDRMC, and our method use ground truth object mask.}
    \label{tab:real}
    \vspace{-1.5em}
\end{table}

\subsection{Ablation Studies}

\noindent{\textbf{Information Sharing}}. 
We compared our full pipeline's material and normal estimation against a version without information sharing in supplementary \cref{fig:glossy_real}. While Stage 1 (1k iterations) in the full pipeline optimizes only the radiance field, we assessed the non-information-sharing model at both 5k iterations, matching material optimization time, and 15k iterations, aligning with geometry optimization time of the full model. Results indicate that information sharing enhances both geometry and material reconstruction, showing that extended training without information sharing still underperforms compared to the full model with less training iterations.

\begin{figure}[h]
    \centering
    \setlength\tabcolsep{1pt}
    
    \begin{tabular}{cccc}
        \small{GT} & \small{ours} & \small{w.o. indirect} & \small{w.o. 2nd split-sum}\\
        \includegraphics[height=0.11\textwidth]{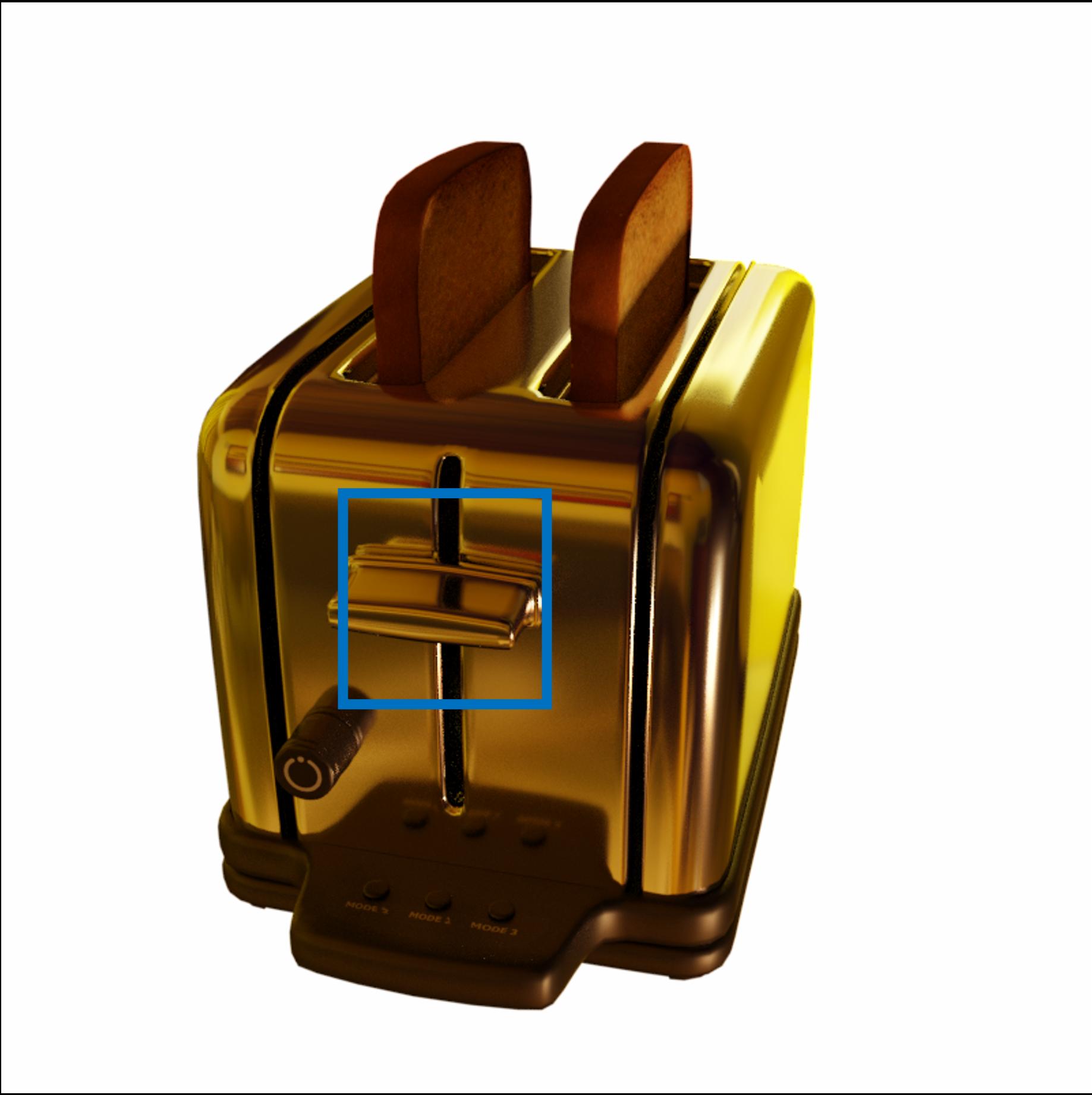} &
        \includegraphics[height=0.11\textwidth]{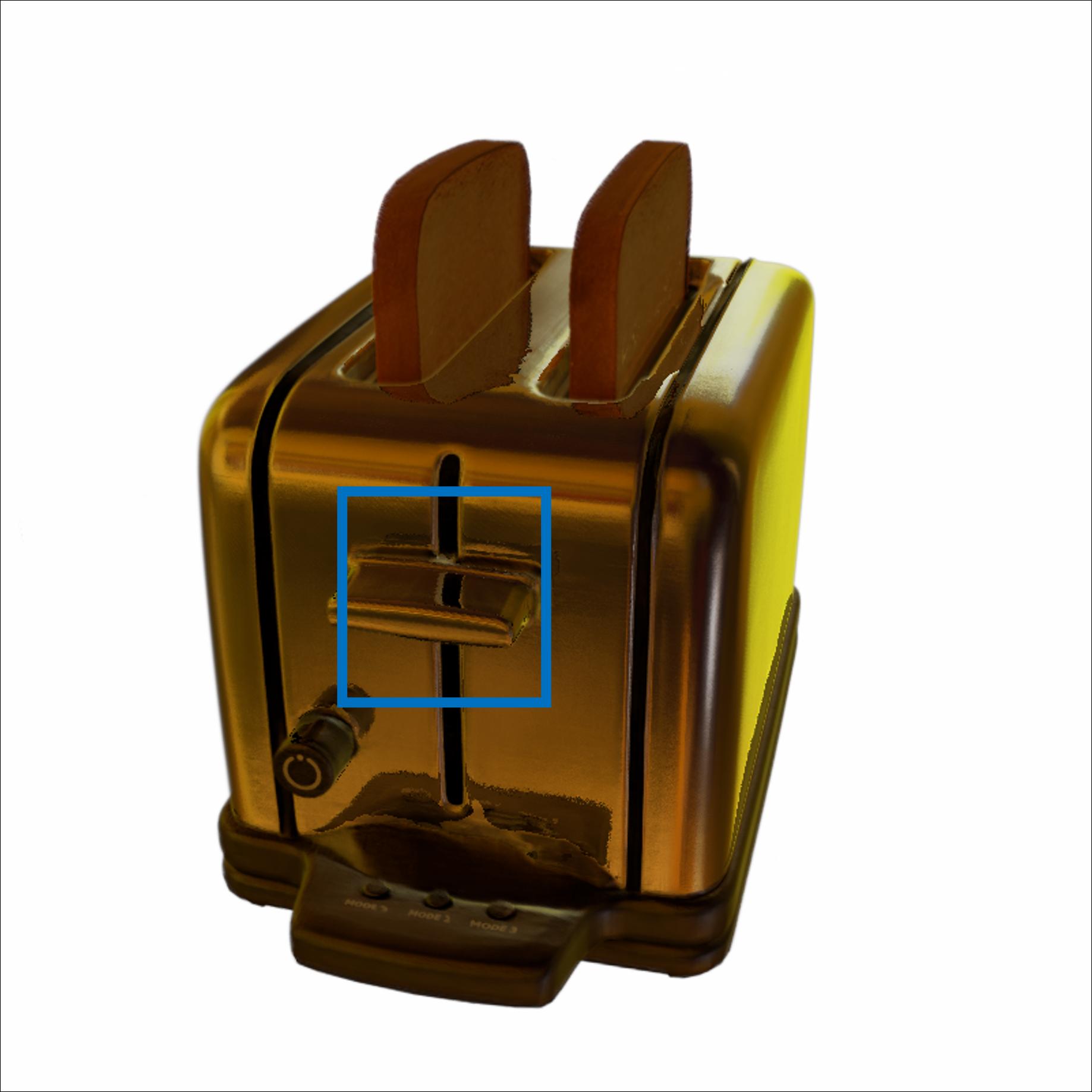} &
        \includegraphics[height=0.11\textwidth]{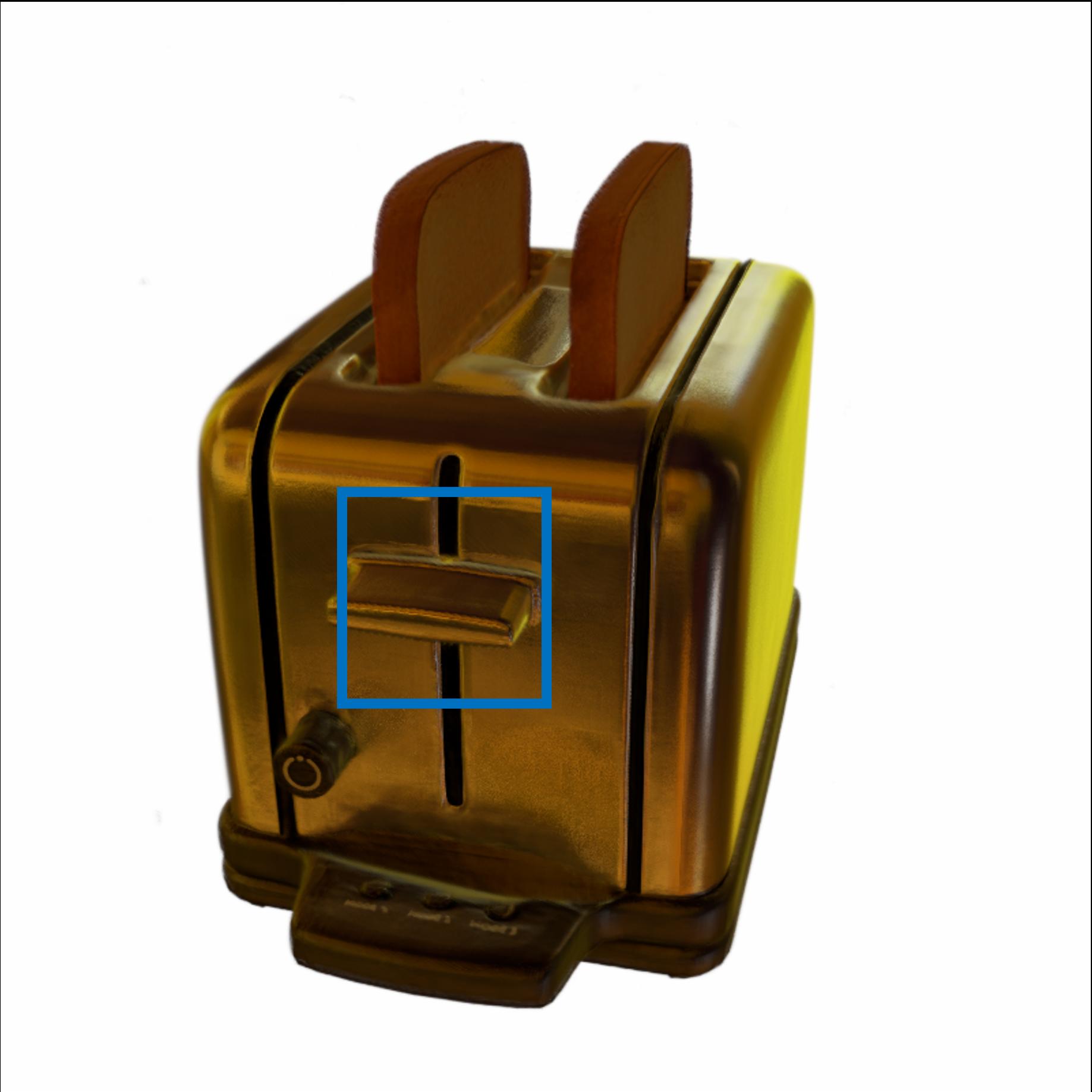} &
        \includegraphics[height=0.11\textwidth]{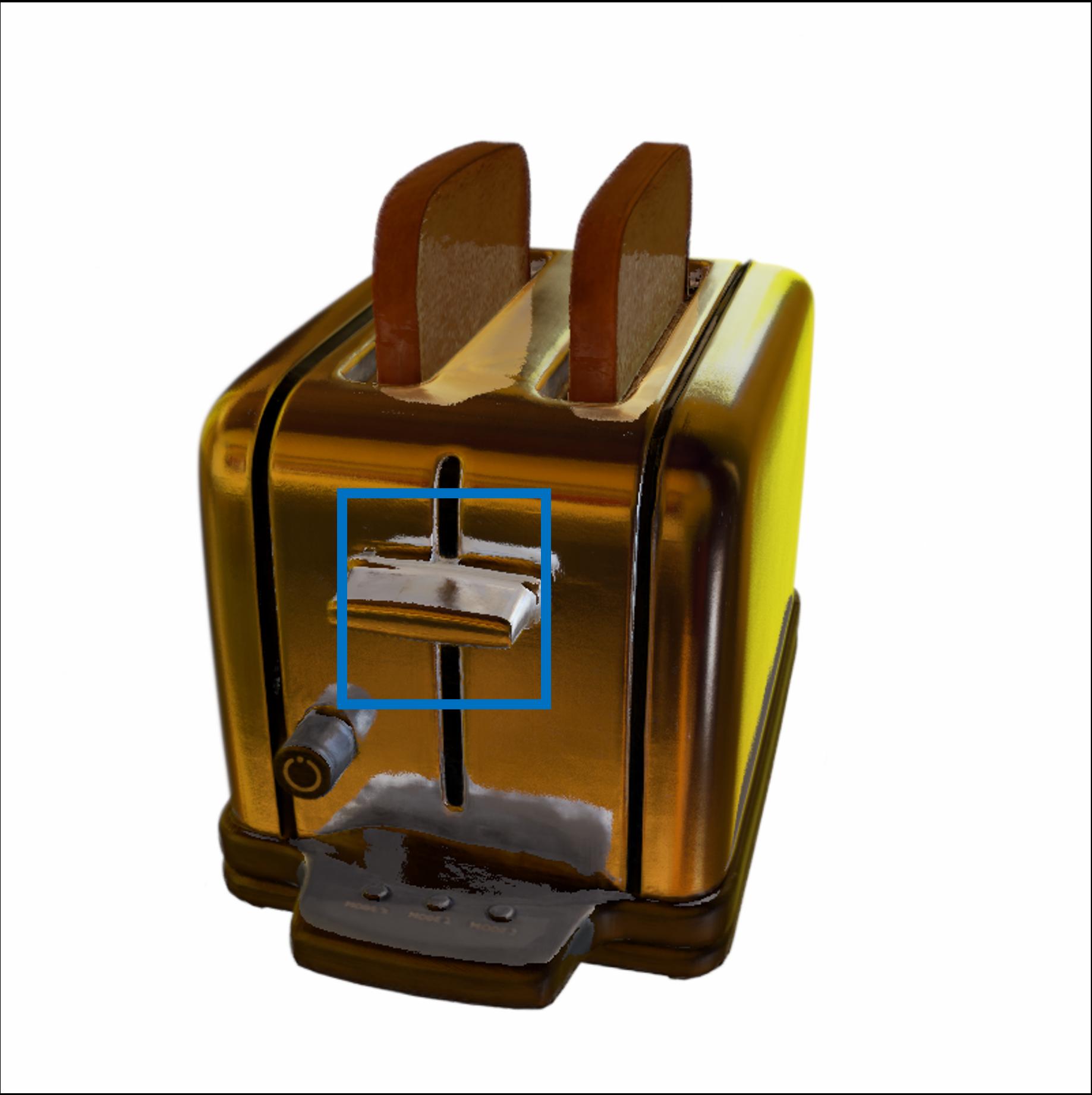} \\
        \includegraphics[height=0.11\textwidth]{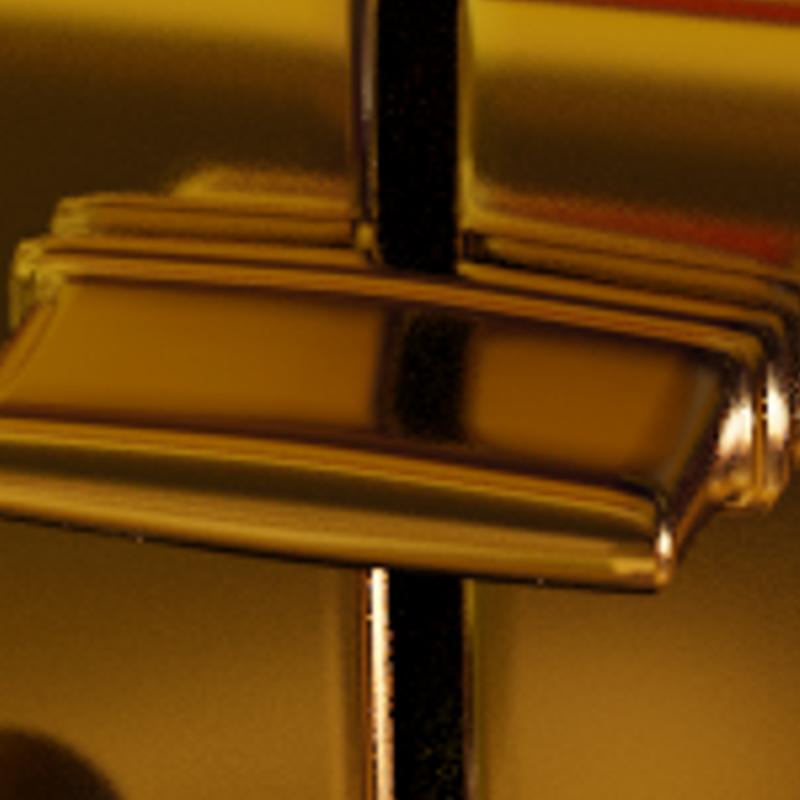} &
        \includegraphics[height=0.11\textwidth]{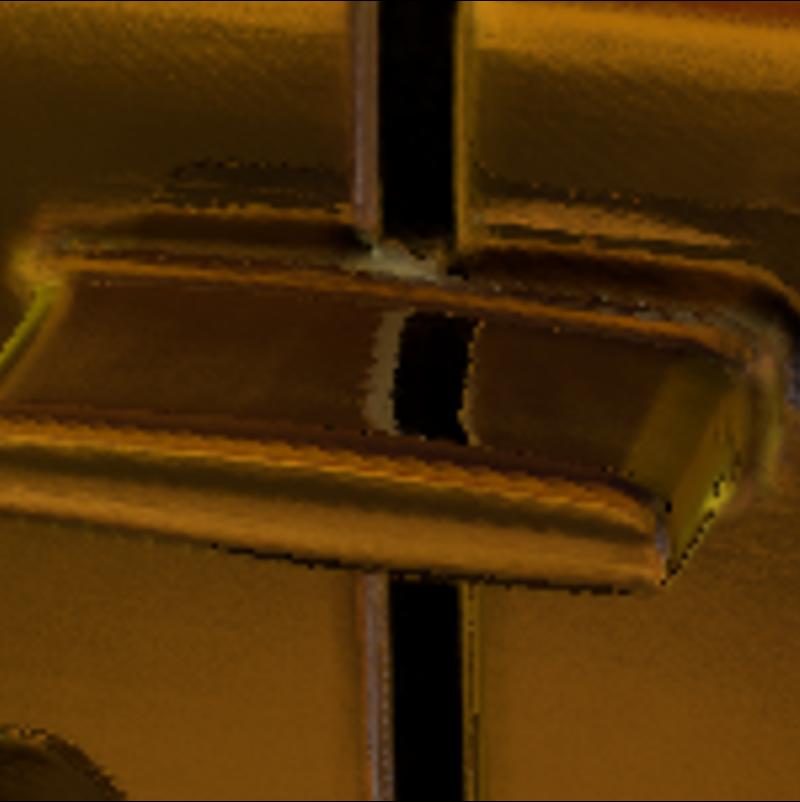} &
        \includegraphics[height=0.11\textwidth]{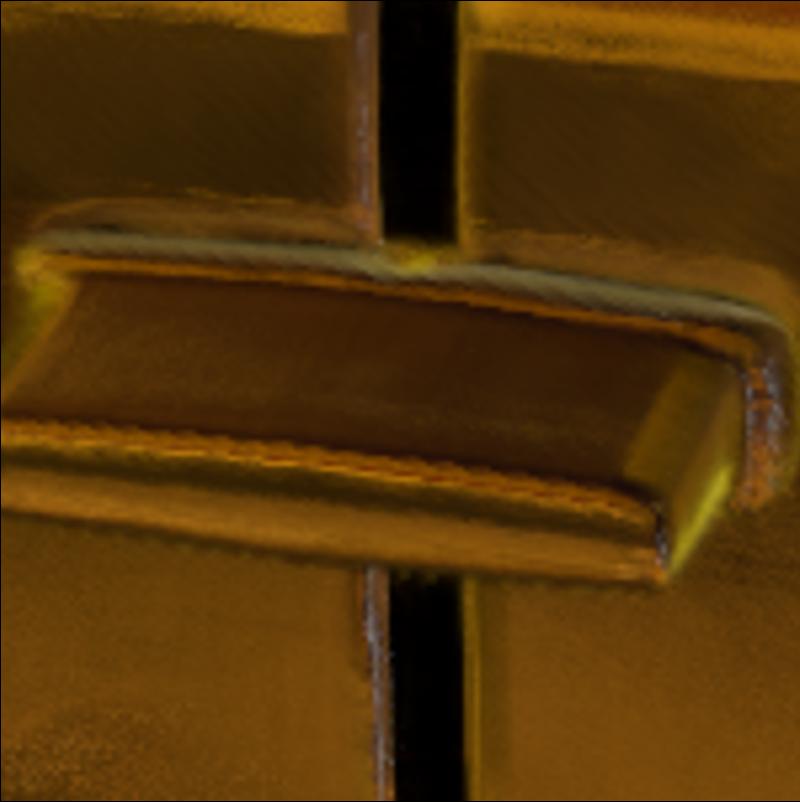} &
        \includegraphics[height=0.11\textwidth]{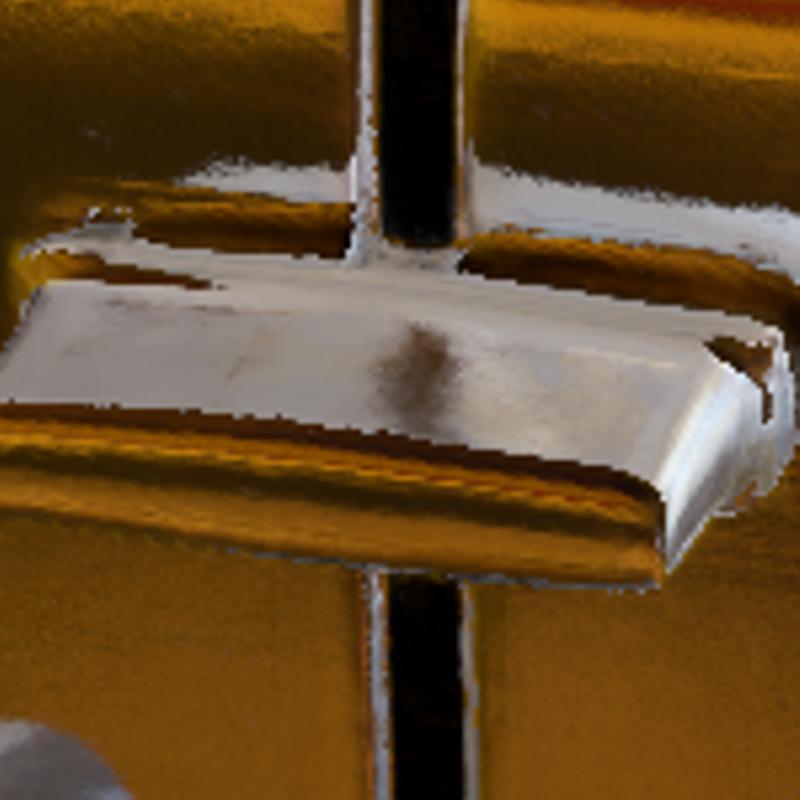} \\
        % \includegraphics[height=0.11\textwidth]{figures/indirect_ablation/helmet_gt_relight_1.jpeg} &
        % \includegraphics[height=0.11\textwidth]{figures/indirect_ablation/helmet_ours_relight_1.jpeg} &
        % \includegraphics[height=0.11\textwidth]{figures/indirect_ablation/helmet_ind_relight_1.jpeg} &
        % \includegraphics[height=0.11\textwidth]{figures/indirect_ablation/helmet_second_relight_1.jpeg} \\
        
        % GT & Ours & NDRMC & GShader & NMF& ENVIDR & NeRO\\
    \end{tabular}
    \caption{\textbf{Ablation study for indirect illumination.}}
    \label{fig:indirect_ablation}
    \vspace{-1em}
\end{figure}

\noindent{\textbf{Indirect illumination}}. In \cref{fig:indirect_ablation}, we visualize the relighting results using the fireplace environment map for the toaster scene using our full model, our model without indirect illumination prediction MLP, and our model without the second split-sum during relighting. The fully implemented model closely replicates the ground truth in terms of lighting and reflection accuracy. Models omitting indirect illumination and the second split-sum relighting show significant degradation. The former displays reduced realism in reflections, while the latter exhibits unnatural contrasts due to the indirect MLP trained in the different light conditions.

\begin{figure}[h]

\centering

\begin{tabular}{@{} c|c @{}}
  \begin{tabular}{@{} c @{}}
    GT \\[1ex]
    \includegraphics[width=2.5cm,height=2.5cm]{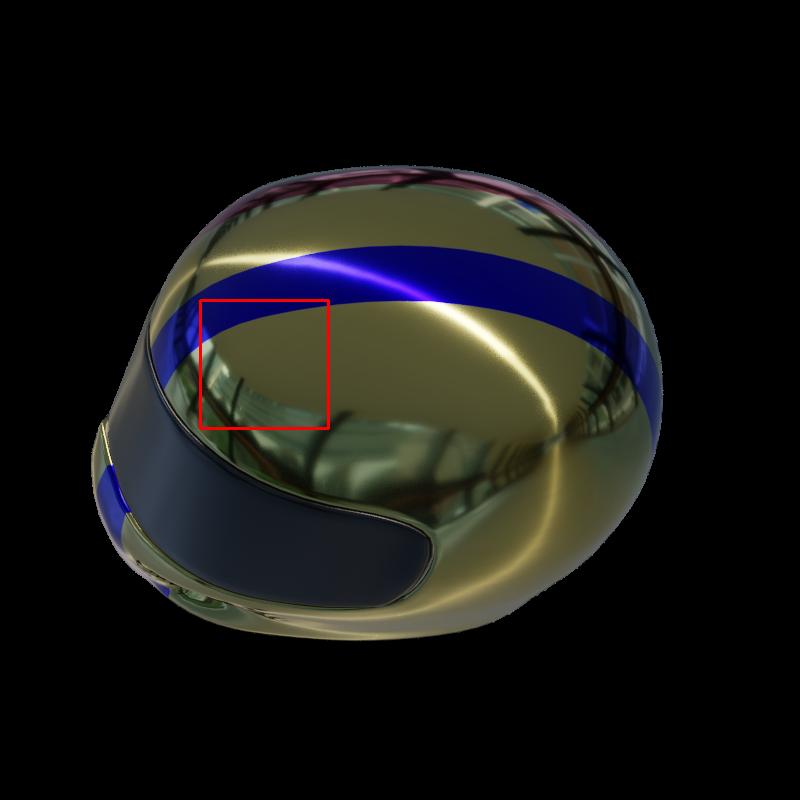} \\
    \includegraphics[width=2.5cm,height=2.5cm]{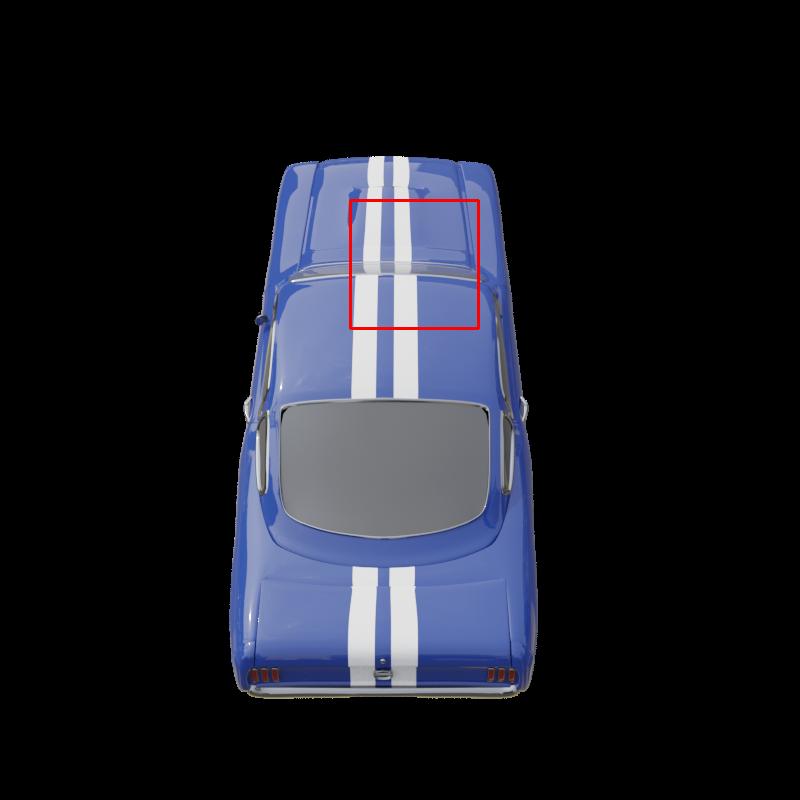}
  \end{tabular}
  &
  \begin{tabular}{ccc}
    \small{Split-sum} & \small{MC, 128spp} & \small{MC, 256spp} \\[1ex]
    \includegraphics[width=1.3cm,height=1.3cm]{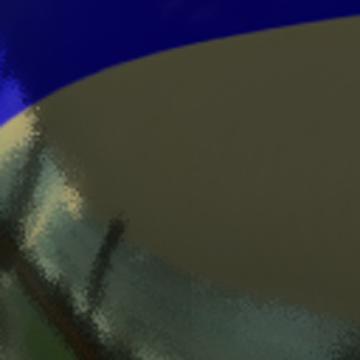} &
    \includegraphics[width=1.3cm,height=1.3cm]{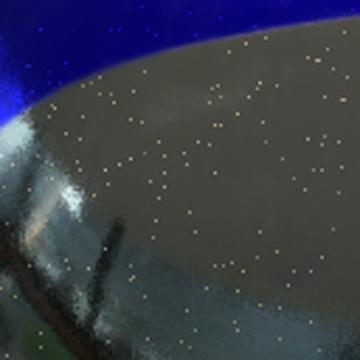} &
    \includegraphics[width=1.3cm,height=1.3cm]{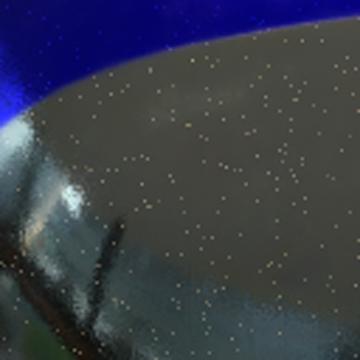}  \\[0.1cm]
    \includegraphics[width=1.3cm,height=1.3cm]{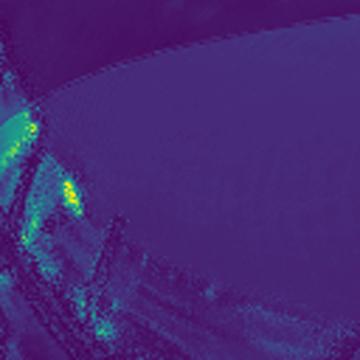} &
    \includegraphics[width=1.3cm,height=1.3cm]{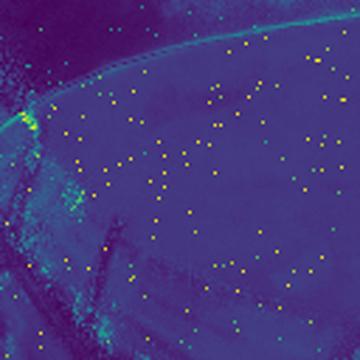} &
    \includegraphics[width=1.3cm,height=1.3cm]{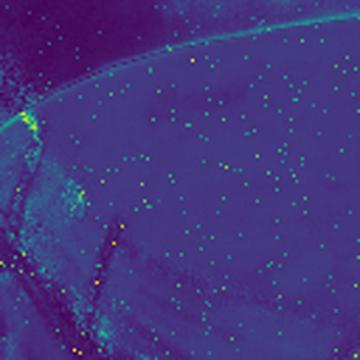}\\
    \includegraphics[width=1.3cm,height=1.3cm]{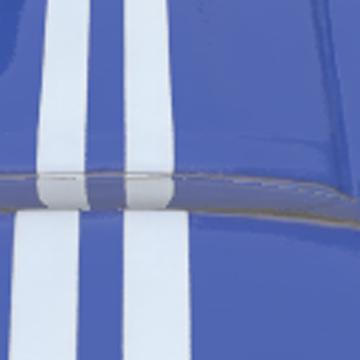} &
    \includegraphics[width=1.3cm,height=1.3cm]{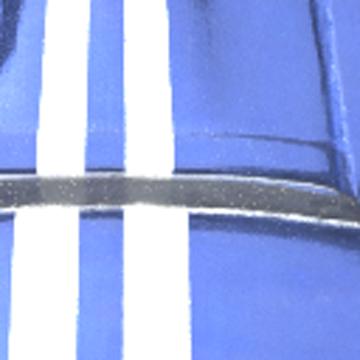} &
    \includegraphics[width=1.3cm,height=1.3cm]{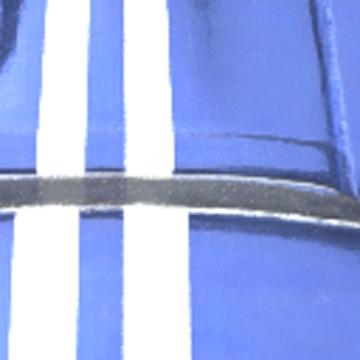}  \\[0.1cm]
    \includegraphics[width=1.3cm,height=1.3cm]{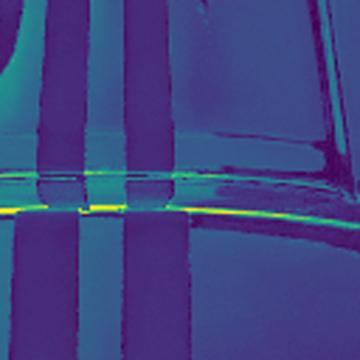} &
    \includegraphics[width=1.3cm,height=1.3cm]{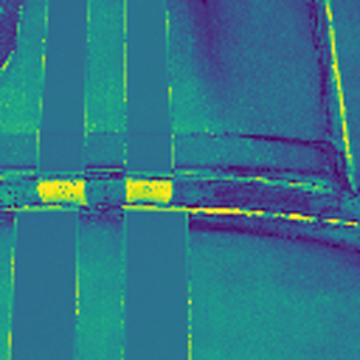} &
    \includegraphics[width=1.3cm,height=1.3cm]{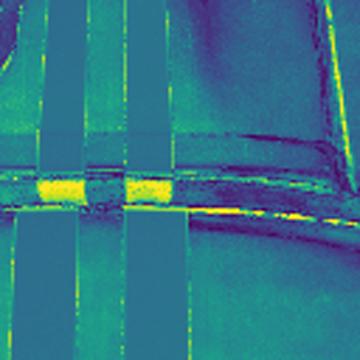}
    
  \end{tabular}

\end{tabular}

\caption{\textbf{Ablation study for Monte Carlo integration}. We visualize the relighting results for helmet and musclecar scene. Odd rows: relightings, even rows: error maps (same color map as we use for roughness maps).}
\label{fig:ablation_mc}
\vspace{-1.5em}
\end{figure}

\noindent{\textbf{Monte Carlo vs. Split-sum}}. In our pipeline, we use the split-sum approximation instead of Monte Carlo sampling to avoid sampling noise. In \cref{fig:ablation_mc}, we compare the relighting results and the corresponding error map with respect to the ground truth using the split-sum approximation, Monte Carlo rendering using multiple-importance sampling (MIS) with 128 spp and 256 spp respectively. We can observe that with split-sum our rendering results have significantly less noise compared with Monte Carlo sampling.

\subsection{Limitations}
Although we achieve superior performance on the reconstruction, inverse rendering, and relighting of shiny objects, our system still has the following limitations. Our system is built on an isotropic BRDF model, which means we cannot handle transparent or anisotropic materials. Furthermore, although our approach can handle indirect illumination for reflective surfaces, shadows for diffuse surfaces are not well solved. Thus, our model shows sub-optimal performance point light environments. Furthermore, due to the inevitable ambiguity problem between the illumination, material parameter, and geometry, the estimated material could still deviate strongly from the ground truth in some cases.

% DO NOT INCLUDE ACKNOWLEDGMENTS IN AN ANONYMOUS SUBMISSION TO SIGGRAPH 2019
%\begin{acks}
%
%The authors would like to thank Dr. Maura Turolla of Telecom
%Italia for providing specifications about the application scenario.
%
%The work is supported by the \grantsponsor{GS501100001809}{National
%  Natural Science Foundation of
%  China}{http://dx.doi.org/10.13039/501100001809} under Grant
%No.:~\grantnum{GS501100001809}{61273304\_a}
%and~\grantnum[http://www.nnsf.cn/youngscientists]{GS501100001809}{Young
%  Scientists' Support Program}.
%
%
%\end{acks}

\section{Conclusion}
We presented RISE-SDF, an end-to-end relightable inverse rendering system for glossy objects. One key idea of RISE-SDF is information-sharing between the radiance-based signed-distance field and the physically based inverse rendering field to improve the geometry and material estimation. We further introduced indirect illumination prediction during training and a second split-sum during relighting to improve the reconstruction and relighting quality on interreflections. Experiments demonstrate that our system has superior novel view synthesis, material reconstruction, geometry reconstruction, and relighting quality compared with the previous state-of-the-art.
  
{
    \small
    \bibliographystyle{ieeenat_fullname}
    \bibliography{main}
}
\clearpage
\setcounter{page}{1}
\maketitlesupplementary

\section{Implementation Details}
Our code is built on the instant-nsr-pl codebase \cite{yuan2022instant}. We use a 512-resolution progressive hash grid with 16 levels. The geometry MLP has $2$ layers with $128$ neurons. The diffuse, specular, secondary MLP has $4$ layers with $128$ neurons. And the roughness and blending MLP has $2$ layers with $128$ neurons. The resolution of the environment map is $6\times512\times512\times3$. We start from the 4-th level of the hash grid and increase by 1 level for every 500 iterations. Regarding the hyperparameters, we use $\lambda_c = 10$, $\lambda_{eik} = 0.1$, $\lambda_{curv} = 1$. We use Adam optimizer with $\beta_1 = 0.9$, $\beta_1 = 0.999$, and $\epsilon = 10^{-12}$. The first stage of our method is training for 10k
iterations and the second stage for 70k iterations. All experiments are conducted on a single RTX 3090Ti GPU. 
\section{Shiny Inverse Rendering Synthetic Dataset}
\begin{table}[]
    \centering
    \resizebox{\linewidth}{!}{
        \begin{tabular}{@{}c c c c ccccc c ccccc c ccccc c@{}}
            \toprule
            Dataset & Material & Relighting & \# Envmaps & Shiny Object \\ \midrule
            NeRF Synthetic \cite{Mildenhall2020ECCV} & \xmark & \xmark & 0 & \xmark \\
            Shiny Blender \cite{verbin2022ref} & \xmark & \xmark & 0 & \cmark \\
            NeRO Synthetic \cite{liu2023nero} & \xmark & \cmark & 3 & \cmark \\
            NeRFactor Synthetic \cite{zhang2021nerfactor} & \cmark & \cmark & 8 & \xmark \\
            TensoIR Synthetic \cite{jin2023tensoir} & \cmark & \cmark & 8 & \xmark \\
            Ours & \cmark & \cmark & 9 & \cmark \\
            \bottomrule
        \end{tabular}
    }
    \caption{\textbf{Comparison of the availability of the datasets.} We show the availability of ground truth material, relighting, number of environment maps for relighting, and availability glossy object. Ours is the first dataset with ground truth material and relighting for shiny objects. } 
    \label{tab:dataset}
    \vspace{-1.5em}
\end{table}

We built our own Shiny Inverse Rendering Dataset with an aligned BRDF model as no dataset with ground truth material and relighting results for glossy objects exists. 
We provide five scenes including \textit{teapot}, \textit{coffee}, \textit{muscle car}, \textit{toaster}, and \textit{helmet} from the Shiny Blender dataset \cite{verbin2022ref}, with ground truth albedo, roughness, and relighting under nine different environment maps. Here are the steps to create the dataset:
\begin{itemize}[leftmargin=10pt]
\item To align the BRDF during dataset generation and inverse rendering, we change the shader nodes of the five objects to the Principled BRDF with default parameters except for the metallic, roughness, and base color (albedo) in Blender \cite{blender}. 
\item We render the objects under ten different light conditions, and choose one of them as the training light, and others for relighting.
\item To export accurate ground truth albedo and roughness, we manually create the blender files using Diffuse BRDF with the albedo and roughness value in the Principled BRDF model as the base color and render the diffuse pass.
\end{itemize}
\noindent In ~\Cref{tab:dataset}, we compare our new dataset to existing ones. To the best of our knowledge, ours is the first glossy dataset with an aligned BRDF model for forward and inverse rendering with accurate material and relighting ground truth. 
\section{Discussion on MLP Predictions in Stage 1}
\label{sec:stage_one_appendix}
Using Eq.~4 and 5 in the main paper, we can divide the per-sample physically-based rendering equation into the following format:
\begin{equation}
    \begin{aligned}
         \{\mathbf{c}_d\}_i = (1 -m_i) {\mathbf{a}_i \over \pi}\int_{H^2} L_{dir}(\mathbf{x}_i, \mathbf{\omega}_j) cos\theta_i d\mathbf{\omega}_j\\
          \{\mathbf{c}_s\}_i= \int_{H^2}{D(\rho_i)F(m_i, \mathbf{a}_i)G \over 4 |\mathbf{d} \cdot \mathbf{n}_i| |\mathbf{\omega}_j \cdot \mathbf{n}_i|} L_{dir}(\mathbf{x}_i, \mathbf{\omega}_j) cos\theta_i d\mathbf{\omega}_j\\
    \end{aligned}
    \label{eqn:separate_appendix}
\end{equation}
Since the geometry is not well reconstructed in the first stage, the SDF is not converged near the surface. Therefore, the amount of volume rendering samples is large during the first stage since it is hard to prune the samples according to the SDF value. To save computation and to enhance training stability, we directly use MLPs to predict these two integration values. A blending weight is predicted to simulate the effect of metallic parameters. Note that we assume there is no indirect illumination in this equation.

\section{Split-sum Approximation}
\label{sec:appendix_split_sum}
Given the sample location $\mathbf{x}_i$, ray direction $\mathbf{d}$, and the normal direction $\mathbf{n}_i$, we write the rendering equation and the corresponding Monte Carlo integration:
\begin{equation}
\begin{aligned}
 \mathbf{c}^{pbr}_i(\mathbf{x}_i, \mathbf{d}) = & \int_{H^2}f_r(\mathbf{x}_i, \mathbf{d}, \mathbf{\omega}_j)L_i(\mathbf{x}_i, \mathbf{\omega}_j) (\mathbf{n}_i \cdot \mathbf{\omega}_j) d\mathbf{\omega}_j \\
  \approx & {1 \over N_{mc}} \sum^{N_{mc}}_{j=1} {f_r(\mathbf{x}_i, \mathbf{d}, \mathbf{\omega}_j)L_i(\mathbf{x}_i, \mathbf{\omega}_j)
 (\mathbf{n}_i \cdot \mathbf{\omega}_j) \over p(\mathbf{\omega}_j; \mathbf{\hat{d}}_i, \rho_i)} \\
 \approx & ({1 \over N_{mc}} \sum^{N_{mc}}_{j=1} {f_r(\mathbf{x}_i, \mathbf{d}, \mathbf{\omega}_j) (\mathbf{n}_i \cdot \mathbf{\omega}_j) \over p(\mathbf{\omega}_j; \mathbf{\hat{d}}_i, \rho_i)}) \\
& \cdot({1 \over N_{mc}} \sum^{N_{mc}}_{j=1} L_i(\mathbf{x}_i, \mathbf{\omega}_j) )
\end{aligned}
 \label{eqn:split-sum_A}
\end{equation}
This integration can be approximated by the multiplication of separate sums. If we write the split-sum Monte Carlo integration back to the continuous form:
\begin{equation}
\begin{aligned}
 \mathbf{c}^{pbr}_i(\mathbf{x}_i, \mathbf{d})  \approx & \int_{H^2}f_r(\mathbf{x}_i, \mathbf{d}, \mathbf{\omega}_j)(\mathbf{n}_i \cdot \mathbf{\omega}_j) d\mathbf{\omega}_j \\ &\cdot \int_{H^2} L_i(\mathbf{x}_i, \mathbf{\omega}_j) p(\mathbf{\omega}_j; \mathbf{\hat{d}}_i, \rho_i)d\mathbf{\omega}_j
\end{aligned}
 \label{eqn:split-sum2_A}
\end{equation}
The first integration is called as BSDF integral, and the second integration is called the light integral. The light integration is approximated by a multi-level mipmap. Consistent with \cite{munkberg2022extracting}, we use cube maps (with resolution $6 \times 512 \times 512$). The base level corresponds to the smallest roughness value, and increases among the mip-levels. For each level, the mipmap is computed by average pooling the base level followed by a convolution using the GGX distribution with the corresponding roughness as the kernel. The mipmap is implemented as a differentiable function with respect to $\mathbf{\hat{d}}_i$ and roughness $\rho$:
\begin{equation}
\begin{aligned}
\int_{H^2} L_i(\mathbf{x}_i, \mathbf{\omega}_j) p(\mathbf{\omega}_j; \mathbf{\hat{d}}_i, \rho)d\mathbf{\omega}_j \approx Mipmap(\mathbf{\hat{d}}_i, \rho)
\end{aligned}
 \label{eqn:mipmap_A}
\end{equation}
In our paper, the BRDF is defined as a simplified version of the Disney BSDF as in Eq.~5 in the main paper. The fresnel term $F$ is defined as:
\begin{equation}
\begin{aligned}
 F = F_0 + (1 - F_0)(1 - \mathbf{\omega}_j \cdot \mathbf{h})^5, \\
\end{aligned}
 \label{eqn:fresnel}
\end{equation}
where $F_0 = 0.04 * (1-m) + m * \mathbf{a}$ is the simplified basic reflection ratio, and $\mathbf{h}$ is the half-way vector between $-\mathbf{d}$ and $\mathbf{\omega}_j$. Since we have both the diffuse and the specular parts in the BSDF, we can separate the rendering equation into the diffuse and specular parts. For the diffuse part, we directly extract the albedo outside of the integrand, and use the normal direction and largest roughness to query the mipmap:
\begin{equation}
\begin{aligned}
  \mathbf{l}^{d}_i & = Mipmap(\mathbf{n}_i, \rho_{max}) \\
 \{\mathbf{c}^{pbr}_d\}_i & = ((1 - m_i)\mathbf{a}_i \int_{H^2}{(\mathbf{n}_i \cdot \mathbf{\omega}_j) \over \pi} d\mathbf{\omega}_j )\mathbf{l}^{d}_i \\
 & = (1 - m_i) * \mathbf{a}_i  * \mathbf{l}^d_i
\end{aligned}
 \label{eqn:diffuse_integral_A}
\end{equation}
For the specular part of BSDF integration, if we substitute the Fresnel in the BSDF:
\begin{equation}
\begin{aligned}  
& \int_{H^2}f_r(\mathbf{x}_i, \mathbf{d}, \mathbf{\omega}_j)(\mathbf{n}_i \cdot \mathbf{\omega}_j) d\mathbf{\omega}_j\\
 = & F_0 \int_{H^2}{f_r(\mathbf{x}_i, \mathbf{d}, \mathbf{\omega}_j) \over F}(1 - (1 - \mathbf{\omega}_j \cdot \mathbf{h})^5)(\mathbf{n}_i \cdot \mathbf{\omega}_j) d\mathbf{\omega}_j\\
& + \int_{H^2}{f_r(\mathbf{x}_i, \mathbf{d}, \mathbf{\omega}_j) \over F}(1 - \mathbf{\omega}_j \cdot \mathbf{h})^5(\mathbf{n}_i \cdot \mathbf{\omega}_j) d\mathbf{\omega}_j \\
\end{aligned}
 \label{eqn:specular_integral_A}
\end{equation}
This leaves two integrations only dependent on $\rho$ and $\mathbf{n}_i \cdot \mathbf{\omega}_j$, then we can precompute the result and store it to a 2D LUT: 
\begin{equation}
\begin{aligned}  
 &\int_{H^2}f_r(\mathbf{x}_i, \mathbf{d}, \mathbf{\omega}_j)(\mathbf{n}_i \cdot \mathbf{\omega}_j) d\mathbf{\omega}_j \\ 
 &= F_0 * F_1(\rho, \mathbf{n}_i \cdot \mathbf{d}),
 + F_2(\rho, \mathbf{n}_i \cdot \mathbf{d}), \\
 &\mathbf{l}^{s}_i =  Mipmap(\mathbf{\hat{d}}_i, \rho)\\
&\{\mathbf{c}^{pbr}_s\}_i =  (F_0 * F_1 + F_2) * \mathbf{l}^{s}_i. 
\end{aligned}
 \label{eqn:lut_A}
\end{equation}

\section{Discussion on Indirect Illumination}
\label{sec:appendix_indirect}

\begin{figure*}[ht]
    \centering
    \setlength{\tabcolsep}{0.0130\linewidth}
    \includegraphics[width=1\linewidth]{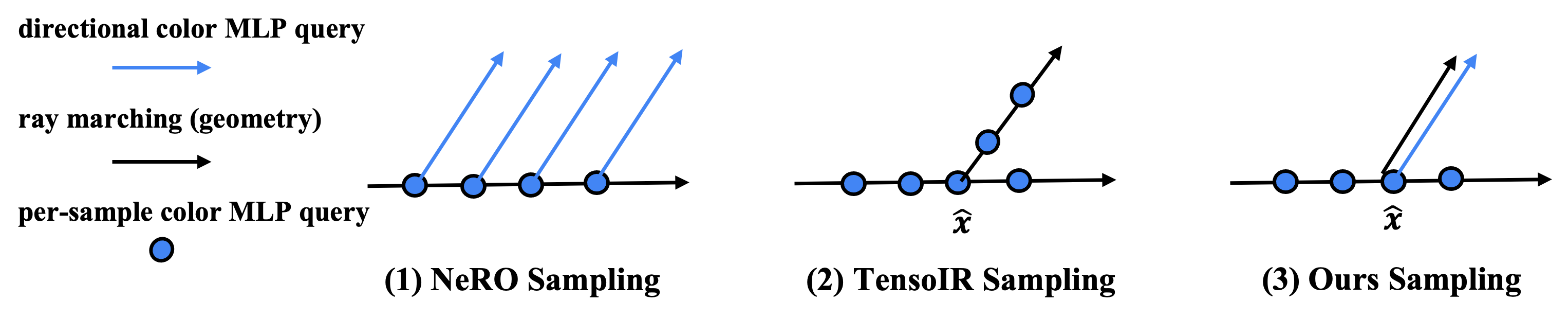}
    \caption[Indirect Sampling Method]{Indirect sampling method. Each blue arrow represents an MLP query using the secondary ray direction. Each black arrow represents a ray marching with multiple geometry MLP queries. Each blue dot represents color MLP queries for each sample. (1) For each sample along the primary ray, NeRO \cite{liu2023nero} queries an opacity MLP and an indirect color MLP to estimate the per-sample indirect illumination. (2) TensoIR \cite{jin2023tensoir} and ENVIDR \cite{liang2023envidr} compute the expected surface intersection $\mathbf{\hat{x}}$ and apply a secondary ray marching. This is not efficient when there are multiple color MLPs in our case. (3) Our indirect sampling only queries the geometry MLP to compute the opacity and uses one color MLP only once for the expected surface intersection. 
      \label{fig:indirect_sampling}}
\end{figure*}
To model the indirect illumination, we first assume that the indirect effect is only apparent for the specular part of the radiance. Therefore, we only modify the specular part $\{\mathbf{c}_s\}_i$ to $\{\mathbf{c}'_s\}_i$ in \cref{eqn:separate_appendix}: 
\begin{equation}
\begin{aligned}  
f_r^s(\mathbf{x}_i, \mathbf{d}, \mathbf{\omega}_j) = & {DFG \over 4 |\mathbf{d} \cdot \mathbf{n}_i| |\mathbf{\omega}_j \cdot \mathbf{n}_i|}\\
  \{\mathbf{c}'_s\}_i =& \int_{H^2} f_r^s(\mathbf{x}_i, \mathbf{d}, \mathbf{\omega}_j)(\mathbf{n}_i \cdot \mathbf{\omega}_j) \\
  &((1-O) L_d(\mathbf{x}_i, \mathbf{\omega}_j) + O L_{ind}(\mathbf{x}_i, \mathbf{\omega}_j) ) d\mathbf{\omega}_j\\
  = & (1-O) \int_{H^2} f_r^s(\mathbf{x}_i, \mathbf{d}, \mathbf{\omega}_j)(\mathbf{n}_i \cdot \mathbf{\omega}_j) L_d(\mathbf{x}_i, \mathbf{\omega}_j)d\mathbf{\omega}_j \\ 
  &+ O \int_{H^2} f_r^s(\mathbf{x}_i, \mathbf{d}, \mathbf{\omega}_j)(\mathbf{n}_i \cdot \mathbf{\omega}_j) L_{ind}(\mathbf{x}_i, \mathbf{\omega}_j)d\mathbf{\omega}_j \\
  =& (1-O) \{\mathbf{c}_s\}_i \\
  &+ O \int_{H^2} f_r^s(\mathbf{x}_i, \mathbf{d}, \mathbf{\omega}_j)(\mathbf{n}_i \cdot \mathbf{\omega}_j) L_{ind}(\mathbf{x}_i, \mathbf{\omega}_j)d\mathbf{\omega}_j 
\end{aligned}
 \label{eqn:indirect_appendix}
\end{equation}
After applying volume rendering to the equation above, we have:
\begin{equation}
\begin{aligned}  
 C_{ind} &= \sum^N_{i=1}w_i\int_{H^2} f_r^s(\mathbf{x}_i, \mathbf{d}, \mathbf{\omega}_j)(\mathbf{n}_i \cdot \mathbf{\omega}_j) L_{ind}(\mathbf{x}_i, \mathbf{\omega}_j)d\mathbf{\omega}_j\\
 \mathbf{C}'_s &= \sum^N_{i=1} w_i \{\mathbf{c}'_s\}_i =  (1-O) \sum^N_{i=1} w_i\{\mathbf{c}_s\}_i + O C_{ind}\\
 &= (1-O)C_s + OC_{ind}
\end{aligned}
\end{equation}
\noindent Compared with NeRO~\cite{liu2023nero}, we use an MLP $f_{ind}$ to directly predict $C_{ind}$ instead of predicting the indirect illumination for each volume sample. More specifically, we use the normal vector $\mathbf{N}$, secondary ray direction $\mathbf{\hat{d}}$, and the geometry feature $\mathbf{\beta}_{\mathbf{\hat{x}}}$ at the expected intersection point. As shown in \cref{fig:indirect_sampling}, our algorithm only has one secondary color MLP query. Assume there are $N$ samples on the primary ray and $M$ samples on the secondary ray ($M << N$). For the indirect sampling in NeRO \cite{liu2023nero}, since there is one MLP for opacity prediction and one MLP for indirect color for each sample on the primary ray, the total number of secondary MLP queries is $2N$. For TensoIR \cite{jin2023tensoir} and ENVIDR \cite{liang2023envidr}, since they compute the surface intersection first and then use secondary ray marching to compute the radiance field, the number of MLP queries is $4M$ (in our case there is one density MLP and three color MLPs). In our algorithm, we only query the geometry MLP for indirect illumination, and query indirect color once for the expected intersection $\mathbf{\hat{x}}$, thus the total query number is $M + 1$. 

\section{Derivation of Second Split-sum}
\label{sec:appendix_secondproof}
We define the illumination as:
\begin{equation}
\begin{aligned}
L_i  =& \mathbbm{1}[\rho > \rho_t] L_{dir}\\
&+ \mathbbm{1}[\rho \leq \rho_t]((1-O) * L_{dir} + O * L_{ind}), 
\end{aligned}
 \label{eqn:second-split-sum-A}
\end{equation}
which means we only consider the indirect light when the roughness is smaller than the threshold. We plug this equation into the light integral in \cref{eqn:split-sum2_A}, and the relighting light integral for the specular part becomes:
\begin{equation}
\begin{aligned}
\mathbf{l}^s_{relight} =& (\mathbbm{1}[\rho > \rho_t] + \mathbbm{1}[\rho \leq \rho_t] * (1-O)) * \mathbf{l}^s \\
& + \mathbbm{1}[\rho \leq \rho_t]*O*\int_{H^2} L_{ind}(\mathbf{\hat{x}}, \mathbf{\omega}_j) p(\mathbf{\omega}_j; \mathbf{\hat{d}}, \rho)d\mathbf{\omega}_j 
\end{aligned}
 \label{eqn:second-split-sum-B}
\end{equation}

\noindent The light integral $\mathbf{l}^s$ in the equation can be computed by volume rendering the per-sample light integral $\mathbf{l}_i^s$. For the second part, with a small enough $\rho_t$, the GGX distribution can be approximated as a delta function with an infinity value at the reflected direction $\mathbf{\hat{d}}$. So we simplify the equation above into the following form:
\begin{equation}
\begin{aligned}
\mathbf{l}^s_{relight} \approx & (\mathbbm{1}[\rho > \rho_t] + \mathbbm{1}[\rho \leq \rho_t] * (1-O)) * \mathbf{l}^s \\
& + \mathbbm{1}[\rho \leq \rho_t]*O*L_{ind}(\mathbf{x}_i, \mathbf{\hat{d}}).
\end{aligned}
 \label{eqn:second-split-sum-simplified-A}
\end{equation}

\section{Relighting Runtime Cost}
\begin{table}[h]
    \centering
    \begin{tabular}{ccl}
         & NeRO  & Ours$^*$\\
          \hline 
    teapot & 0.481 & 0.461 \\
    musclecar & 0.292 & 0.253\\
    coffee & 0.203 & 0.193\\
    toaster & 0.199 & 0.133\\
    helmet & 0.282 & 0.148 \\
    \hline 
    Avg. & 0.292 & 0.238 \\
    \end{tabular}
    \caption{\textbf{Relighting Runtime (fps) of NeRO and our pipeline.}}
    \label{tab:runtime}
\end{table}

\noindent In \cref{tab:runtime}, we show the per-scene relighting runtime (fps) of NeRO (Blender)~\cite{liu2023nero} and our pipeline. We can find that both methods cannot achieve interactive frame rates, while our method achieves higher quality with comparable runtime.

\section{Diffuse Synthetic Scene Result}
\begin{table*}[t!]
    \centering
    \resizebox{\linewidth}{!}{
        \begin{tabular}{@{} c c c c c c ccc c ccc c ccc@{}}
            \toprule
            \multirow{2}{*}{Scene} & &
            \multirow{2}{*}{Method} & & 
            \multicolumn{1}{c}{Normal} & & \multicolumn{3}{c}{Albedo} & & \multicolumn{3}{c}{Novel View Synthesis} & & \multicolumn{3}{c}{Relighting}  \\ \cline{5-5} \cline{5-9} \cline{11-13} \cline{15-17}
            & & & & MAE $\downarrow$ & & PSNR $\uparrow$ & SSIM $\uparrow$ & LPIPS $\downarrow$ & & PSNR $\uparrow$ & SSIM $\uparrow$ & LPIPS $\downarrow$ & & PSNR $\uparrow$ & SSIM $\uparrow$ & LPIPS $\downarrow$ \\ \hline

            % Lego
            \multirow{4}{*}{Lego} & & NeRFactor & & 9.767 & & 25.444 & 0.937 & 0.112 & & 26.076 & 0.881 & 0.151 & & 23.246 & 0.865 & 0.156 \\
            
            & & InvRender & & 9.980 & & 21.435 & 0.882 & 0.160 & & 24.391 & 0.883 & 0.151 & & 20.117 & 0.832 & 0.171 \\ 
                        
            & & TensoIR & & 5.980 & & 25.240 & 0.900 & 0.145 & & 34.700 & 0.968 & 0.037 & & 27.596 & 0.922 & 0.095 \\
            
            & & Ours& & 9.247 & & 20.457 & 0.890 & 0.113 & & 31.657 & 0.995 & 0.009 & & 25.599 & 0.980 & 0.028 \\ \hline

            % Hotdog
            \multirow{4}{*}{Hotdog} & & NeRFactor & & 5.579 & & 24.654 & 0.950 & 0.142 & & 24.498 & 0.940 & 0.141 & & 22.713 & 0.914 & 0.159 \\
            
            & & InvRender & & 3.708 & & 27.028 & 0.950 & 0.094 & & 31.832 & 0.952 & 0.089 & & 27.630 & 0.928 & 0.089 \\ 
            
            & & TensoIR & & 4.050 & & 30.370 & 0.947 & 0.093 & & 36.820 & 0.976 & 0.045 & & 27.927 & 0.933 & 0.115 \\
            
            & & Ours & & 4.515 & & 22.756 & 0.961 & 0.075 & & 37.866 & 0.997 & 0.007 & & 26.665 & 0.977 & 0.038 \\ \hline

            % Armadillo
            \multirow{4}{*}{Armadillo} & & NeRFactor & & 3.467 & & 28.001 & 0.946 & 0.096 & & 26.479 & 0.947 & 0.095 & & 26.887 & 0.944 & 0.102 \\
            
            & & InvRender & & 1.723 & & 35.573 & 0.959 & 0.076 & & 31.116 & 0.968 & 0.057 & & 27.814 & 0.949 & 0.069 \\ 
                        
            & & TensoIR & & 1.950 & & 34.360 & 0.989 & 0.059 & & 39.050 & 0.986 & 0.039 & & 34.504 & 0.975 & 0.045 \\
            
            & & Ours & & 3.098 & & 42.440 & 0.959 & 0.032 & & 42.290 & 0.999 & 0.001 & & 32.150 & 0.992 & 0.009 \\ \hline
            
            % Ficus
            \multirow{4}{*}{Ficus} & & NeRFactor & & 6.442 & & 22.402 & 0.928 & 0.085 & & 21.664 & 0.919 & 0.095 & & 20.684 & 0.907 & 0.107 \\
            
            & & InvRender & & 4.884 & & 25.335 & 0.942 & 0.072 & & 22.131 & 0.934 & 0.057 & & 20.330 & 0.895 & 0.073 \\ 
            
            & & TensoIR & & 4.420 & & 27.130 & 0.964 & 0.044 & & 29.780 & 0.973 & 0.041 & & 24.296 & 0.947 & 0.068 \\
            
            & & Ours & & 6.409 & & 31.889 & 0.909 & 0.147 & & 27.794 & 0.965 & 0.043 & & 24.501 & 0.943 & 0.077 \\ \hline

            % Average
            \multirow{4}{*}{Avg.} & & NeRFactor & & 6.314 & & 25.125 & \cellcolor{orange} 0.940 & 0.109 & & 24.679 & 0.922 & 0.120 & & 23.383 &  \cellcolor{yellow} 0.908 & 0.131 \\
            
            & & InvRender & & \cellcolor{orange} 5.074 & & \cellcolor{yellow} 27.341 & \cellcolor{yellow} 0.933 & \cellcolor{yellow} 0.100 & & \cellcolor{yellow} 27.367 & \cellcolor{yellow} 0.934 & \cellcolor{yellow} 0.089 & & \cellcolor{yellow} 23.973 & 0.901 & \cellcolor{yellow} 0.101 \\ 
            
            & & TensoIR & & \cellcolor{tablered} 4.100 & & \cellcolor{orange} 29.275  & \cellcolor{tablered} 0.950 & \cellcolor{tablered} 0.085 & & \cellcolor{tablered} 35.088 & \cellcolor{orange} 0.976 & \cellcolor{orange} 0.040 & & \cellcolor{tablered} 28.580 & \cellcolor{orange} 0.944 & \cellcolor{orange} 0.081 \\
            
            & & Ours & & \cellcolor{yellow} 5.817 & & \cellcolor{tablered} 29.386 & 0.930 & \cellcolor{orange} 0.091 & & \cellcolor{orange} 34.902 & \cellcolor{tablered} 0.989 & \cellcolor{tablered} 0.015 & & \cellcolor{orange} 27.229 & \cellcolor{tablered} 0.973 & \cellcolor{tablered} 0.038 \\ 
            
            \bottomrule
        \end{tabular}
    }
    % \caption{Our average results compares with NeRFactor and InvRender on four synthetic datasets}
    % \caption{Quantitative comparisons on the synthetic dataset. Our (single-light) results have 
    % significantly outperformed  the baseline methods by producing more accurate normals and albedo, 
    % thus achieving more realistic novel view synthesis and relighting results. 
    % Our method can further take images captured multiple rotated lighting conditions that 
    % are achieved by rotating the object, and boost the performance in inverse rendering. (All albedeo results are scaled to aligned with ground-truth as done in PhySG \cite{zhang2021physg}. )
    \caption{Per-scene results on the TensoIR synthetic datasets.}
    \label{tab:diffuse_synthetic_results}
\end{table*}
\noindent In \cref{tab:diffuse_synthetic_results}, we compare the qualitative metrics on TensoIR dataset~\cite{jin2023tensoir}, which contains diffuse objects. Since shadows are not explicitly considered in our pipeline, performance on more diffuse datasets is comparable to TensoIR but does not achieve state-of-the-art in every metric. One of the potential solutions to improve the quality is to add an MLP or a spherical harmonic grid to cache the shadow after stage one as done in GS-IR~\cite{liang2024gsir}.

\section{Discussions on Glossy Real Dataset}
\label{sec:appendix_real}
We generate ground truth object masks by projecting the ground truth mesh to the camera planes. We use the generated masks to train our model in company with mask loss as in NeuS\cite{wang2021neus}. Our model gives sub-optimal performance on this real dataset because it wrongly estimate near-field indirect illumination. The plate holding the objects is masked out and the indirect illumination MLP $f_{ind}$ could not explain secondary shading effects. Further, we do not consider explicitly the reflections of the photographer on the object. For objects with a large part of the reflection of the photographer, our method would struggle to estimate the correct geometry, material, and environment light. We believe the results can be ameliorated by taking objects' surroundings into account and modeling explicitly the reflections of the photographer. 

\section{Discussion on Limitations}

\begin{figure*}[h]
    \centering
    \includegraphics[width=0.7\linewidth]{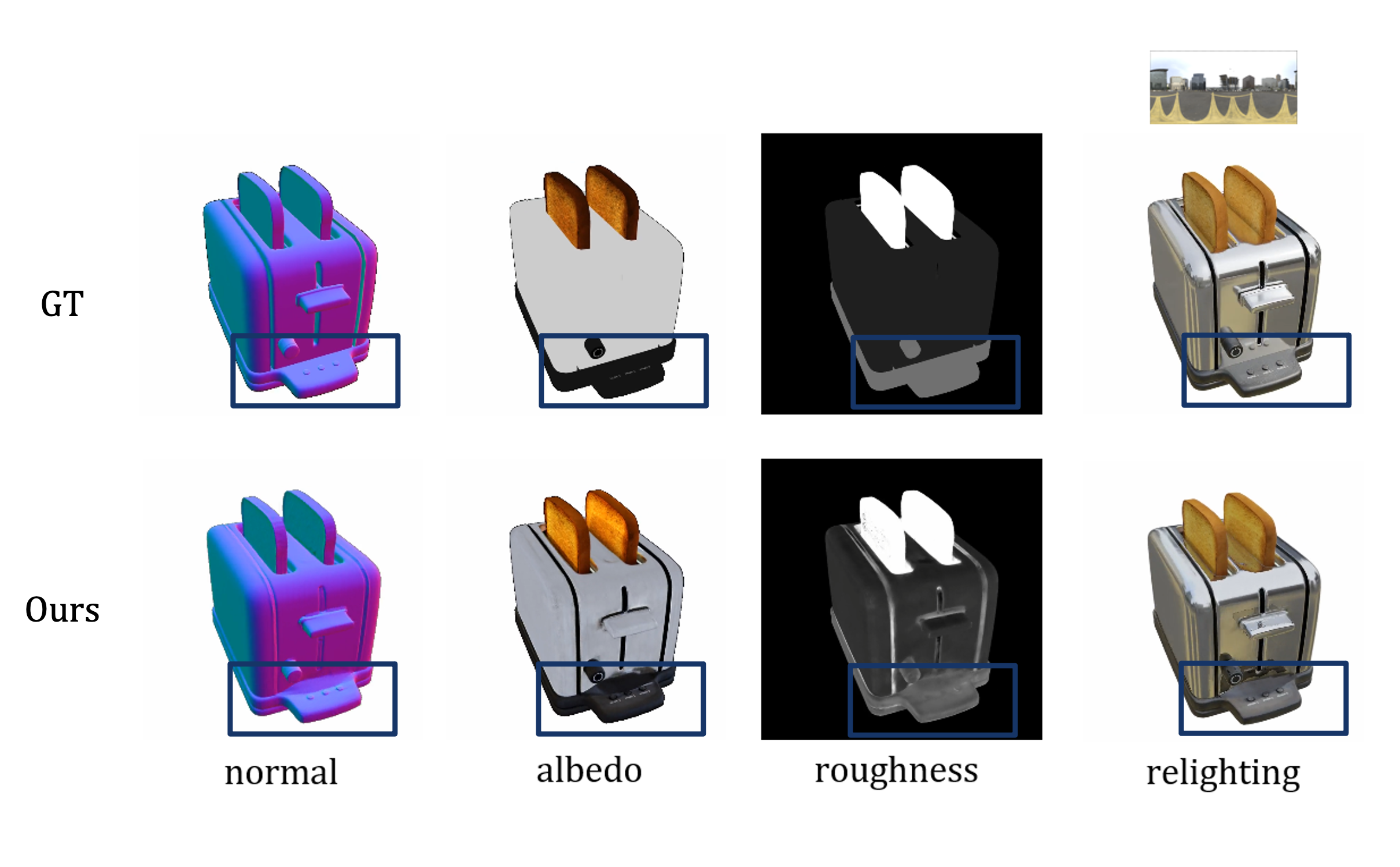}
    \vspace{-1.5em}
    \caption[An example of our Limitations]{\textbf{An example of our Limitations.}}
    \label{fig:limitation}
\end{figure*}

In \cref{fig:limitation}, we show an example of our limitations. In this case, since shadows are not well represented in our model, it causes ambiguity between the shadow and the geometry, thus generating some artifacts. 

\section{Per-Scene Results on the Synthetic Dataset}
\begin{table*}[t!]
    \centering
    \resizebox{\linewidth}{!}{
        \begin{tabular}{@{} c c c c c c c c ccc c ccc c ccc@{}}
            \toprule
            \multirow{2}{*}{Scene} & &
            \multirow{2}{*}{Method} & & 
            \multicolumn{1}{c}{Normal} & &
            \multicolumn{1}{c}{Roughness} & &
            \multicolumn{3}{c}{Albedo} & & \multicolumn{3}{c}{Novel View Synthesis} & & \multicolumn{3}{c}{Relighting}  \\ \cline{5-5} \cline{7-7} \cline{9-11} \cline{13-15} \cline{17-19}
            & & & & MAE $\downarrow$ & & PSNR $\uparrow$ & & PSNR $\uparrow$ & SSIM $\uparrow$ & LPIPS $\downarrow$ & & PSNR $\uparrow$ & SSIM $\uparrow$ & LPIPS $\downarrow$ & & PSNR $\uparrow$ & SSIM $\uparrow$ & LPIPS $\downarrow$ \\ \hline

            % Teapot
            \multirow{4}{*}{Teapot} & & NDR & & 5.43 & & 28.32 & & 31.47 & 0.98 & 0.02 & & 33.17 & 0.98 & 0.02 & & 26.78 & 0.93 & 0.03 \\
            
            & & NDRMC & & 4.63 & & 27.69 & & 30.55 & 0.99 & 0.02 & & 27.98 & 0.96 & 0.04 & & 27.20 & 0.93 & 0.04 \\
            
            & & NMF & & 5.18 & & 26.25 & & 19.42& 0.76 & 0.21 & & 33.39 & 0.98 & 0.02 & & 24.41 & 0.96 & 0.03 \\
                        
            & & ENVIDR & & 1.61 & & - & &	- & - & - & & 37.16& 0.98&0.02 && 29.22&0.97&0.03\\
            
            & & GShader & & 7.24 & & - & & -&-&-&& 34.84 & 0.97 &0.02 &&
            26.41 & 0.96 & 0.03 \\
            
            & & NeRO & & 1.23 & & 17.50 & & 29.42 & 0.98 & 0.02 & & 35.99 & 0.99 & 0.01& & 32.48 & 0.98 & 0.02 \\ 
            
            & & Ours (full model) & & 0.78 && 38.86 && 34.26 &	0.99 & 0.01 & & 38.55 & 0.99 & 0.01 & & 32.55 & 0.99 & 0.02\\
            
            & & Ours (full model, 5 hrs) & & 0.76 & & 38.61 & & 34.41 & 0.99 & 0.01 & & 39.15 & 0.99 & 0.01 & &  32.46 & 0.99 & 0.03 \\ 
            
            \hline 
            
             % Coffee
            \multirow{4}{*}{Coffee} & & NDR & & 10.44 & & 23.59 & & 20.99 & 0.91 & 0.13 & & 25.98 & 0.90 & 0.10 & & 22.25&0.82&0.15\\
            
            & & NDRMC & & 8.44 & & 23.60 & & 22.43 & 0.95 & 0.08 & & 23.93&0.87& 0.15 & & 23.34 & 0.86 & 0.14\\
            
            & & NMF & & 4.04 & & 23.13 & & 13.79 & 0.80 & 0.22 & & 26.52 & 0.91 & 0.08 & & 21.38 & 0.89 & 0.10 \\
                        
            & & ENVIDR & & 7.58 & & - & & - & - & - & & 26.13 & 0.89 & 0.13 & & 24.72 & 0.87 & 0.12\\
            
            & & GShader & & 7.06 & & -  & & - & - & -& &26.83 & 0.91 & 0.11 & &  23.59 & 0.90 & 0.10\\
            
            & & NeRO & & 3.13 & & 23.22 & & 19.01 & 0.92 & 0.12 & & 27.10 & 0.94 & 0.09 & & 25.88 & 0.92 & 0.09 \\ 
            
            & & Ours (full model) & & 3.43 & & 29.24 & & 16.91 & 0.89 & 0.12 & & 27.04 & 0.99 & 0.02 & & 25.03 & 0.98 & 0.03\\
            
            & & Ours (full model, 5 hrs) & & 3.36 & & 29.24 & & 16.92 & 0.89 & 0.11 & & 26.93 & 0.99 & 0.02 & & 25.49 & 0.98 & 0.03 \\ 
            
            \hline

            % % Car
            \multirow{4}{*}{Car} & & NDR & & 5.37 & & 22.02 & & 20.19 & 0.92 & 0.11 & & 29.73 & 0.95 & 0.04 & & 22.46 & 0.86 & 0.09\\
            
            & & NDRMC & & 5.12 & & 21.61 & & 20.07 & 0.92 & 0.11 & & 27.45 & 0.93 & 0.06 & & 26.40 & 0.90 & 0.06\\
            
            & & NMF & & 2.90 & & 26.09 & & 14.36 & 0.85 & 0.13 & & 31.31 & 0.96 & 0.02 & & 23.31 & 0.93 & 0.04\\
                        
            & & ENVIDR & & 3.32 & & - & & - & - & - & & 31.55 & 0.95 & 0.05 & & 26.43 & 0.93 & 0.06\\
            
            & & GShader & & 5.71 & & - & & - & - & - & & 31.62 & 0.95 & 0.05 & & 25.74 & 0.93 & 0.05\\
            
            & & NeRO & & 5.95 & & 23.70& &22.48&0.92&0.10 & & 26.98&0.94&0.06 & & 26.37 & 0.93 & 0.06 \\ 
            
            & & Ours (full model) & & 2.43 & & 25.65 & & 25.14 & 0.95 & 0.16 & & 32.06 & 0.99 & 0.01 & & 28.20 & 0.99 & 0.03\\
            
            & & Ours (full model, 5 hrs) & & 2.22 & & 27.74 & & 25.39 & 0.95 & 0.04 & & 32.65 & 0.99 & 0.01 & & 28.32 & 0.99 & 0.03\\ 
            
            \hline 
            
            % % Toaster
            \multirow{4}{*}{Toaster} & & NDR & & 6.05 & & 18.50 & & 14.99 & 0.87 & 0.16 & & 28.27 & 0.93 & 0.07 & &15.82 & 0.71 & 0.25\\
            
            & & NDRMC & & 4.47 & & 18.15 & & 16.27 & 0.89 & 0.12 & & 25.29 & 0.89 & 0.14 & & 22.13 & 0.85 & 0.15\\
            
            & & NMF & & 2.62 & & 14.44 & & 9.59 & 0.69 & 0.30 & & 29.82 & 0.94 & 0.04 & & 17.97 & 0.86 & 0.11\\
                        
            & & ENVIDR & & 3.26 & & - & & - & - & - & & 28.64 & 0.91 & 0.10 & & 22.50 & 0.86 & 0.12\\
            
            & & GShader & & 4.88 & & - & & - & - & -& & 28.47 & 0.92 & 0.10 & & 21.30 & 0.88 & 0.11\\
            
            & & NeRO & & 2.16 & & 14.99 & & 19.43 & 0.88 & 0.19 & & 29.27 & 0.94 & 0.08 & & 25.70 & 0.91 & 0.09 \\ 
            
            & & Ours (full model) & & 2.19 & & 20.48 & & 20.96 & 0.89 & 0.09 & & 30.36 & 0.99 & 0.01 & & 25.27 & 0.98 & 0.03\\
            
            & & Ours (full model, 5 hrs) & & 2.08 & & 20.48 & & 19.15&0.90&0.09&&30.86&0.99&0.01 & &25.38 & 0.98 & 0.03\\ 
            
            \hline 
            % Helmet
            \multirow{4}{*}{Helmet} & & NDR & & 2.09&&24.83&&24.07&0.92&0.09&&29.41&0.94&0.08&&17.98&0.76&0.20\\
            
            & & NDRMC & & 1.18&&27.95&&23.70&0.92&0.09&&26.77&0.92&0.12&&26.68&0.90&0.10\\
            
            & & NMF & & 0.78&&20.57&&14.03&0.69&0.19&&31.54&0.96&0.03&&20.88&0.89&0.08\\
                        
            & & ENVIDR & & 0.89&&-&&-&-&-&&34.02&0.95&0.06&&24.75&0.91&0.07\\
            
            & & GShader & & 6.00&&-&&-&-&-&&26.23&0.91&0.09&&21.55&0.88&0.11\\
            
            & & NeRO & & 6.00 & & 21.77&&24.19&0.92&0.06&&32.33&0.98&0.03&&30.14&0.95&0.06 \\ 
            
            & & Ours (full model) & & 0.52&&33.07&&33.92&0.97&0.01&&32.25&0.99&0.01&&29.43&0.99&0.03\\
            
            & & Ours (full model, 5 hrs) & &0.48&&34.31&&33.92&0.98&0.01&&32.87&0.99&0.01 & & 29.43 & 0.99&0.02\\ 
            
            \bottomrule
        \end{tabular}
    }
    % \caption{Our average results compare with baselines on five synthetic datasets}
    \caption{Per-scene results on the synthetic datasets.}
    \label{tab:synthetic_results}
\end{table*}
In \cref{tab:synthetic_results}, we provide the results for individual synthetic scenes mentioned in Sec. 4 of the main paper. Our method
outperforms both baselines in all four scenes.

\section{More Results}
More results including relighting, material, normal reconstruction, information sharing ablation, and real dataset are shown in Fig. 8-12.

\begin{figure*}[p]
    \centering
    \setlength\tabcolsep{1pt}
    \begin{tabular}{ccccccc}
        
        \includegraphics[height=0.14\textwidth]{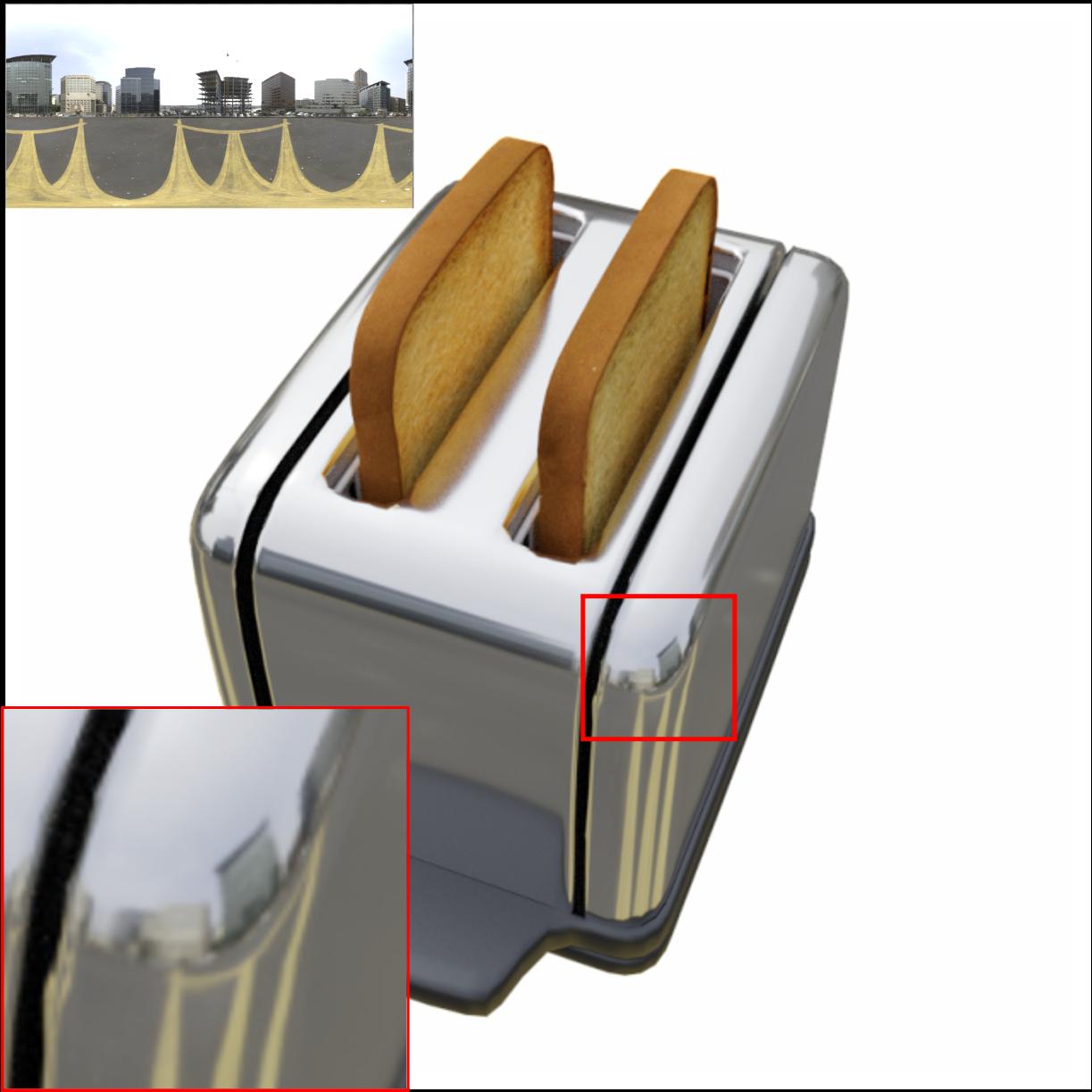} &
        \includegraphics[height=0.14\textwidth]{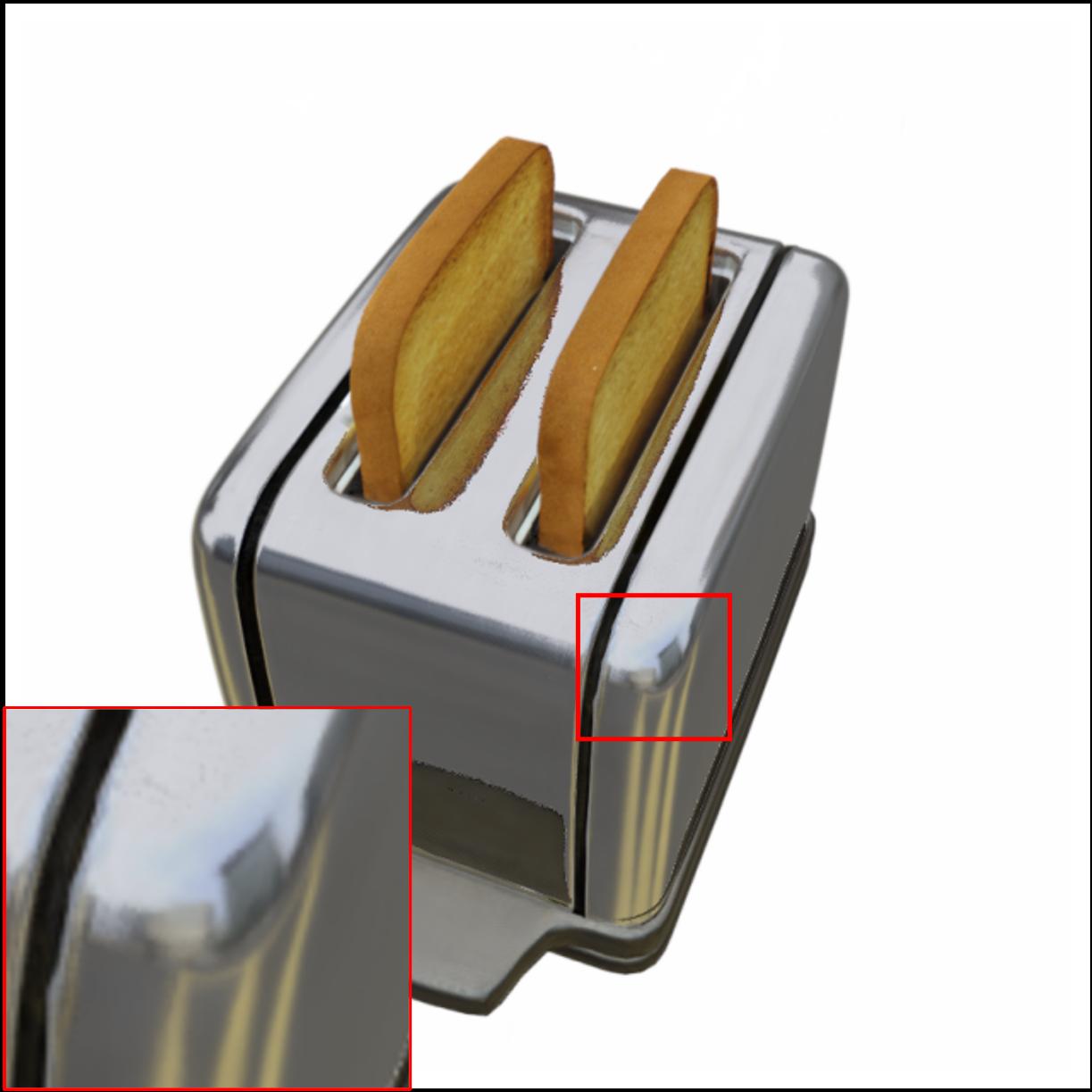} &
        \includegraphics[height=0.14\textwidth]{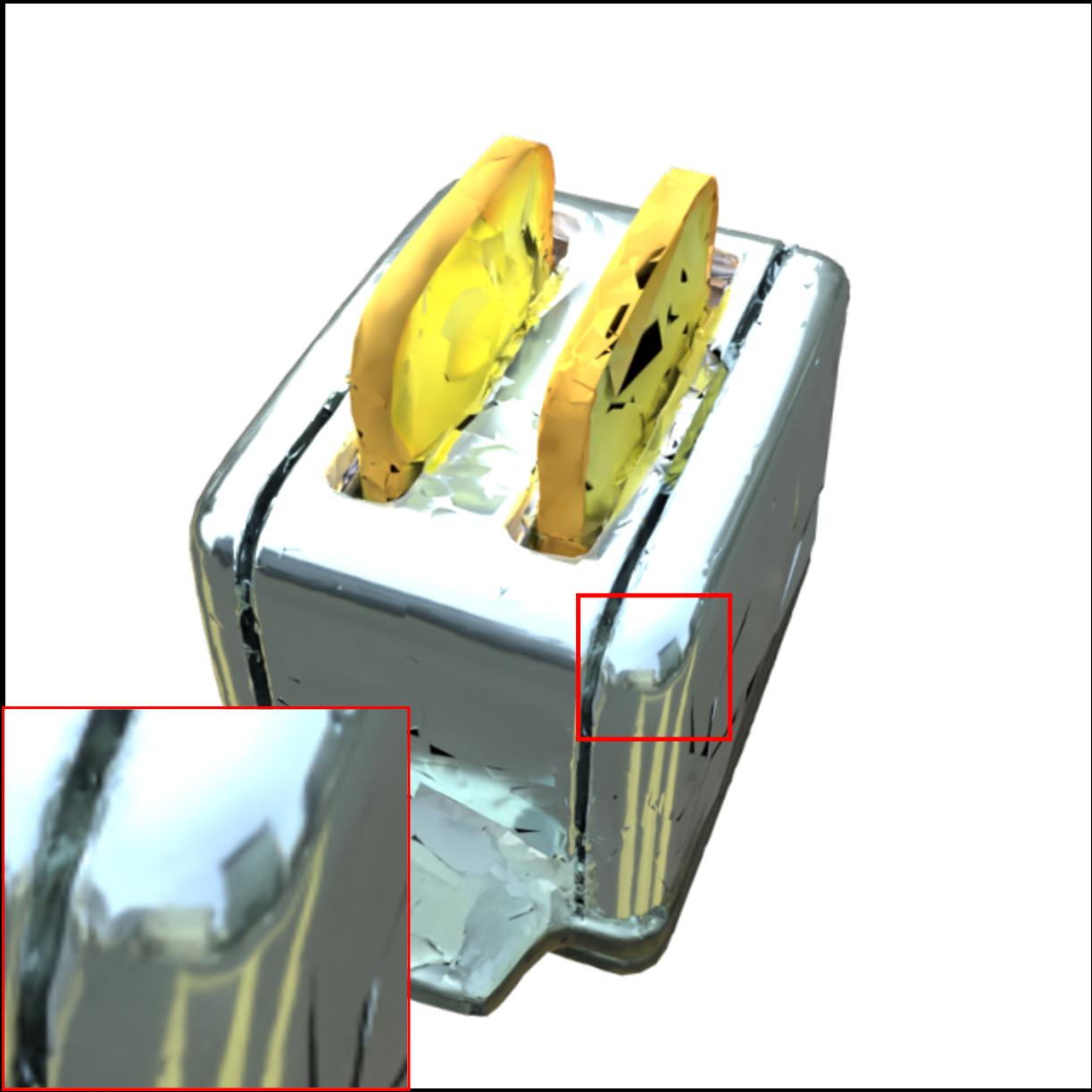} &
        \includegraphics[height=0.14\textwidth]{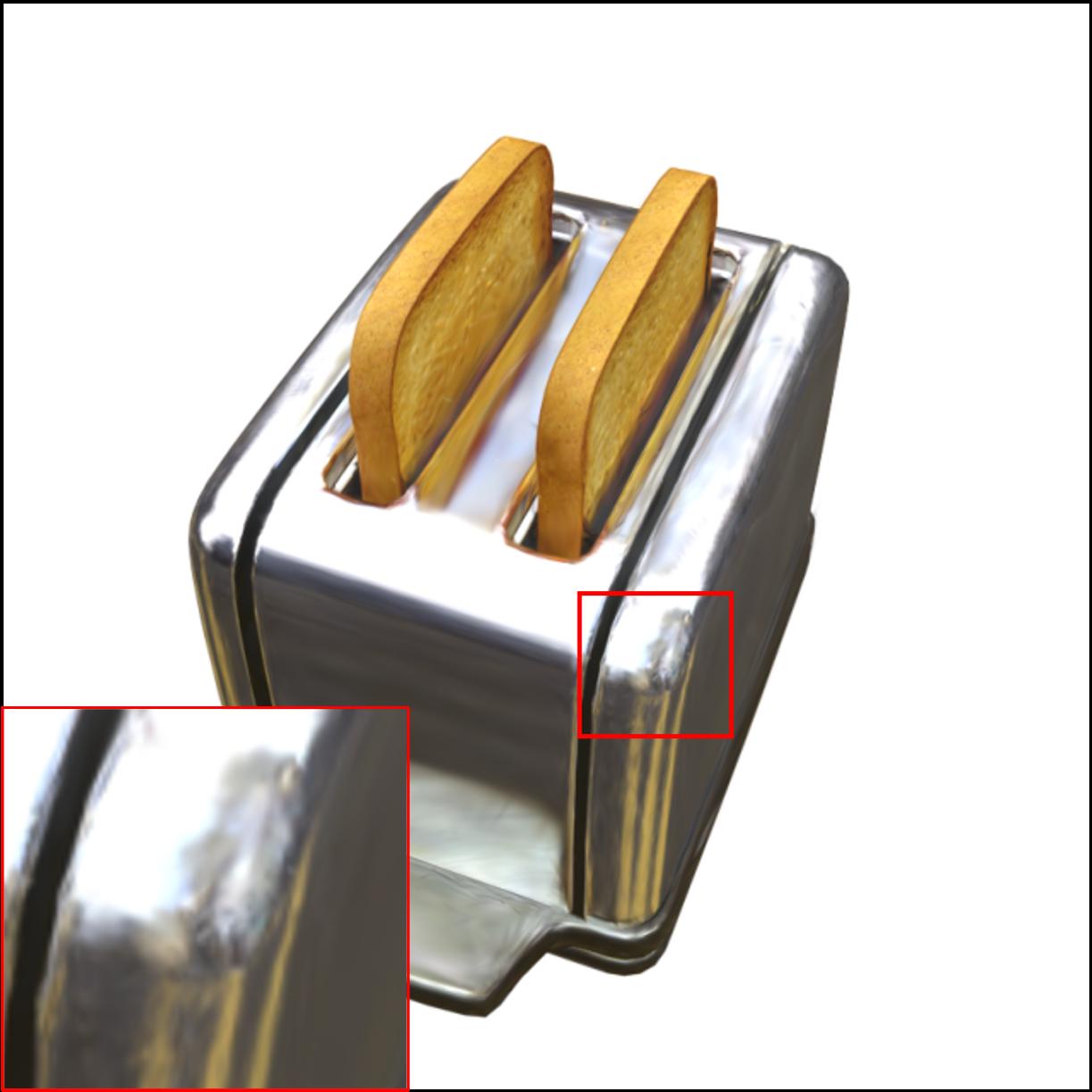} &
        \includegraphics[height=0.14\textwidth]{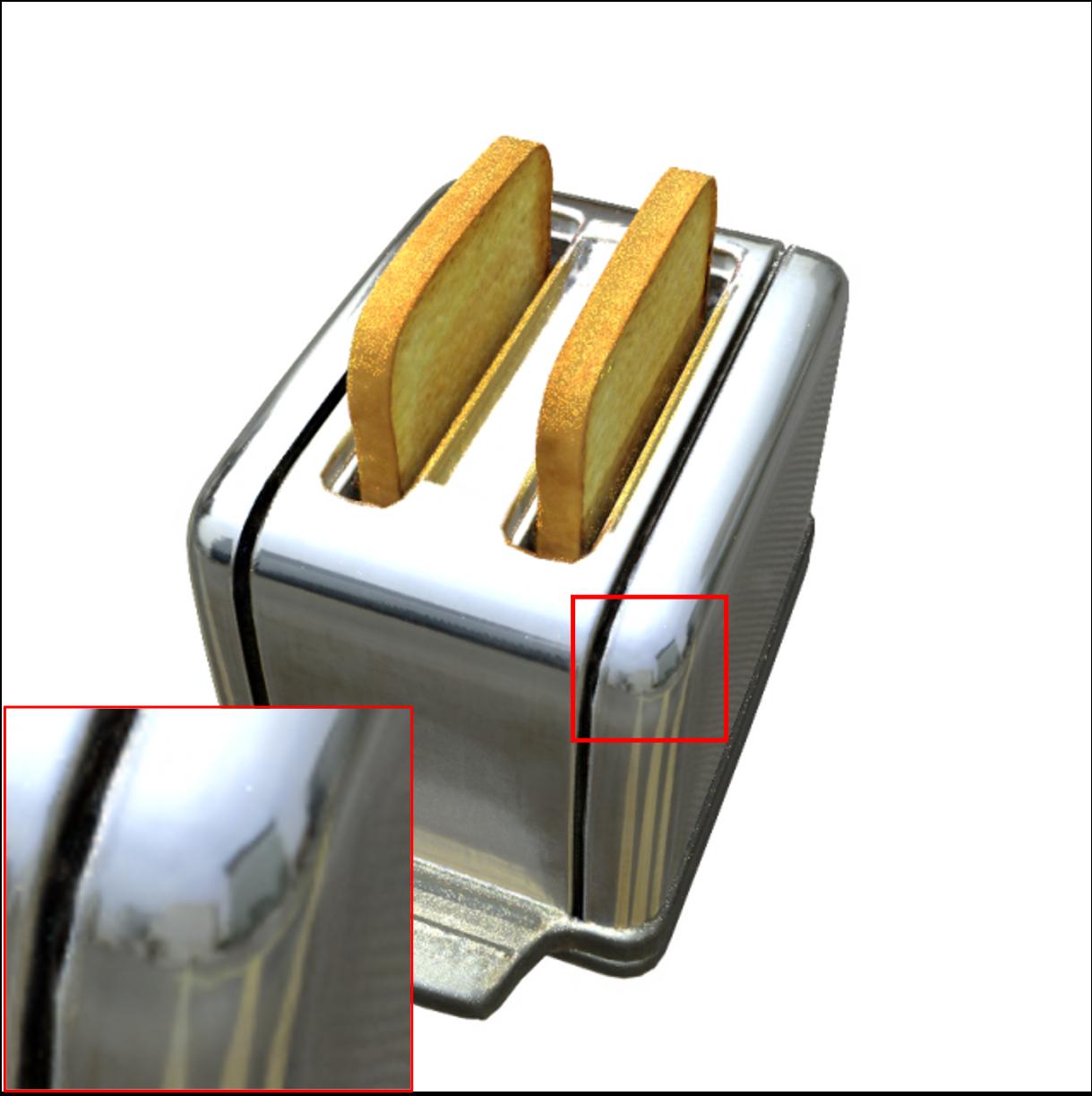} &
        \includegraphics[height=0.14\textwidth]{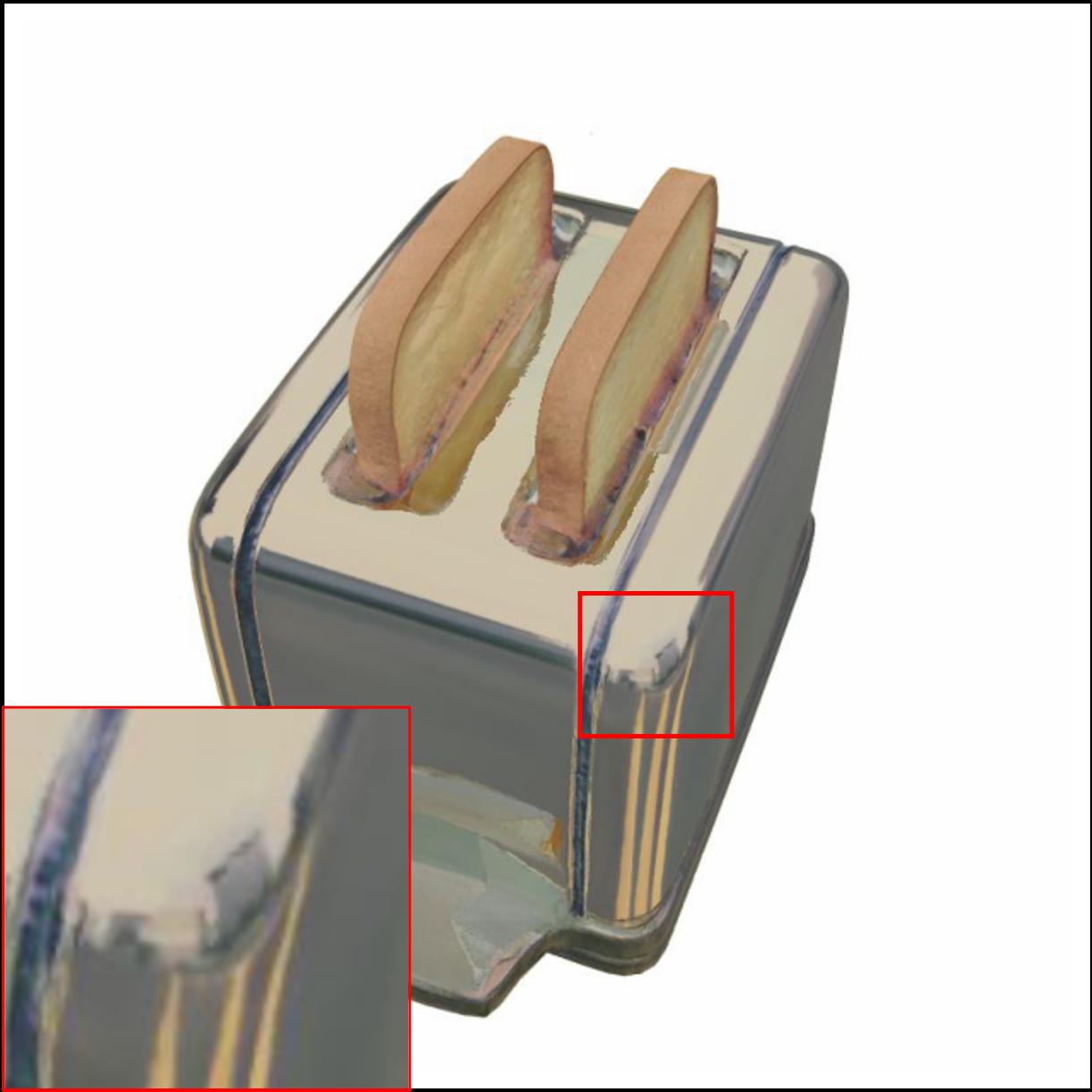}&
        \includegraphics[height=0.14\textwidth]{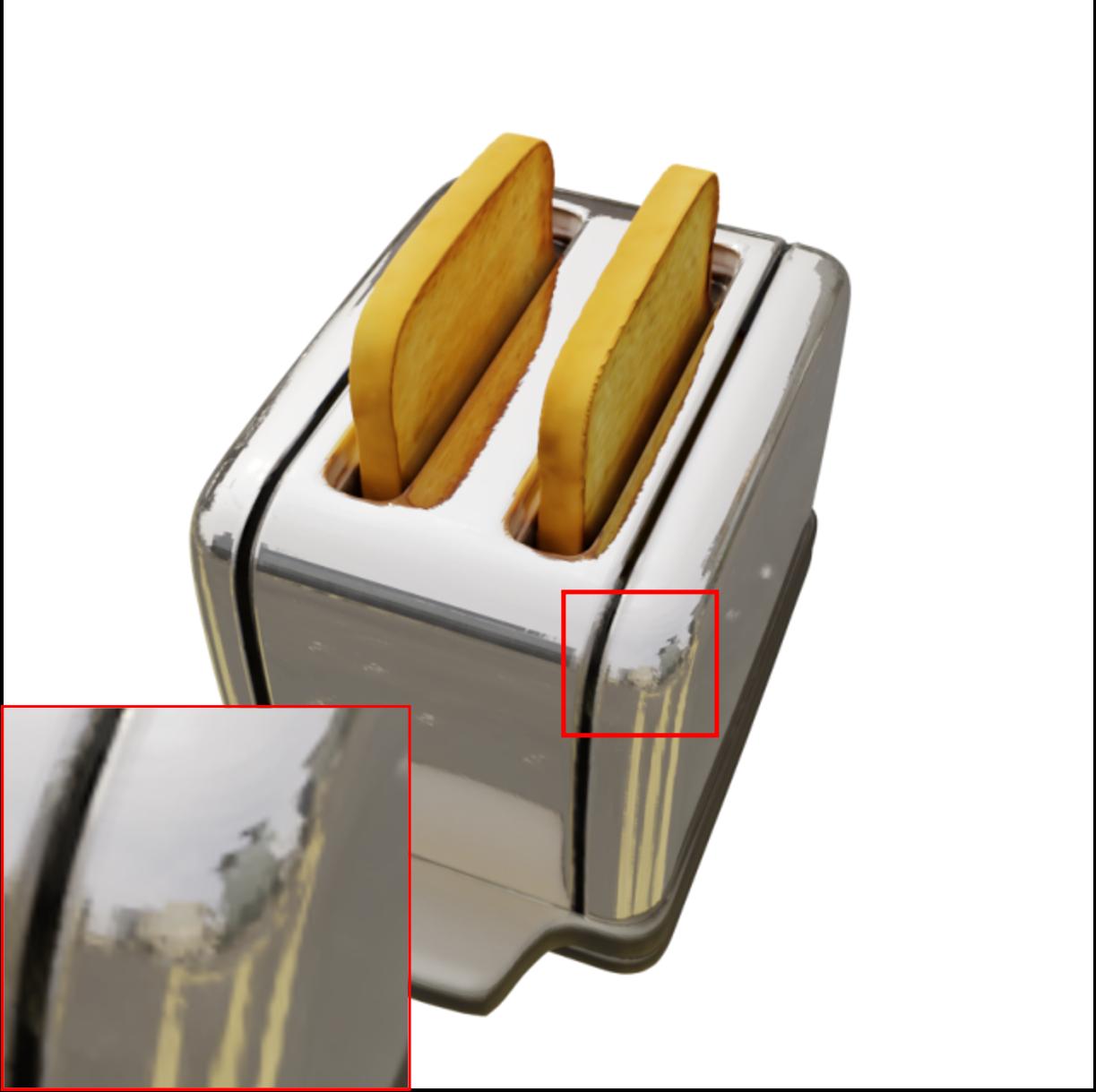}\\

        \includegraphics[height=0.14\textwidth]{figures/figure_only/relight/helmet_gt_relight_0.jpeg} &
        \includegraphics[height=0.14\textwidth]{figures/figure_only/relight/helmet_ours_relight_0.jpeg} &
        \includegraphics[height=0.14\textwidth]{figures/figure_only/relight/helmet_ndrmc_relight_0.jpeg} &
        \includegraphics[height=0.14\textwidth]{figures/figure_only/relight/helmet_gs_relight_0.jpeg} &
        \includegraphics[height=0.14\textwidth]{figures/figure_only/relight/helmet_nmf_relight_0.jpeg} &
        \includegraphics[height=0.14\textwidth]{figures/figure_only/relight/helmet_envidr_relight_0.jpeg}&
        \includegraphics[height=0.14\textwidth]{figures/figure_only/relight/helmet_nero_relight_0.jpeg}\\

        \includegraphics[height=0.14\textwidth]{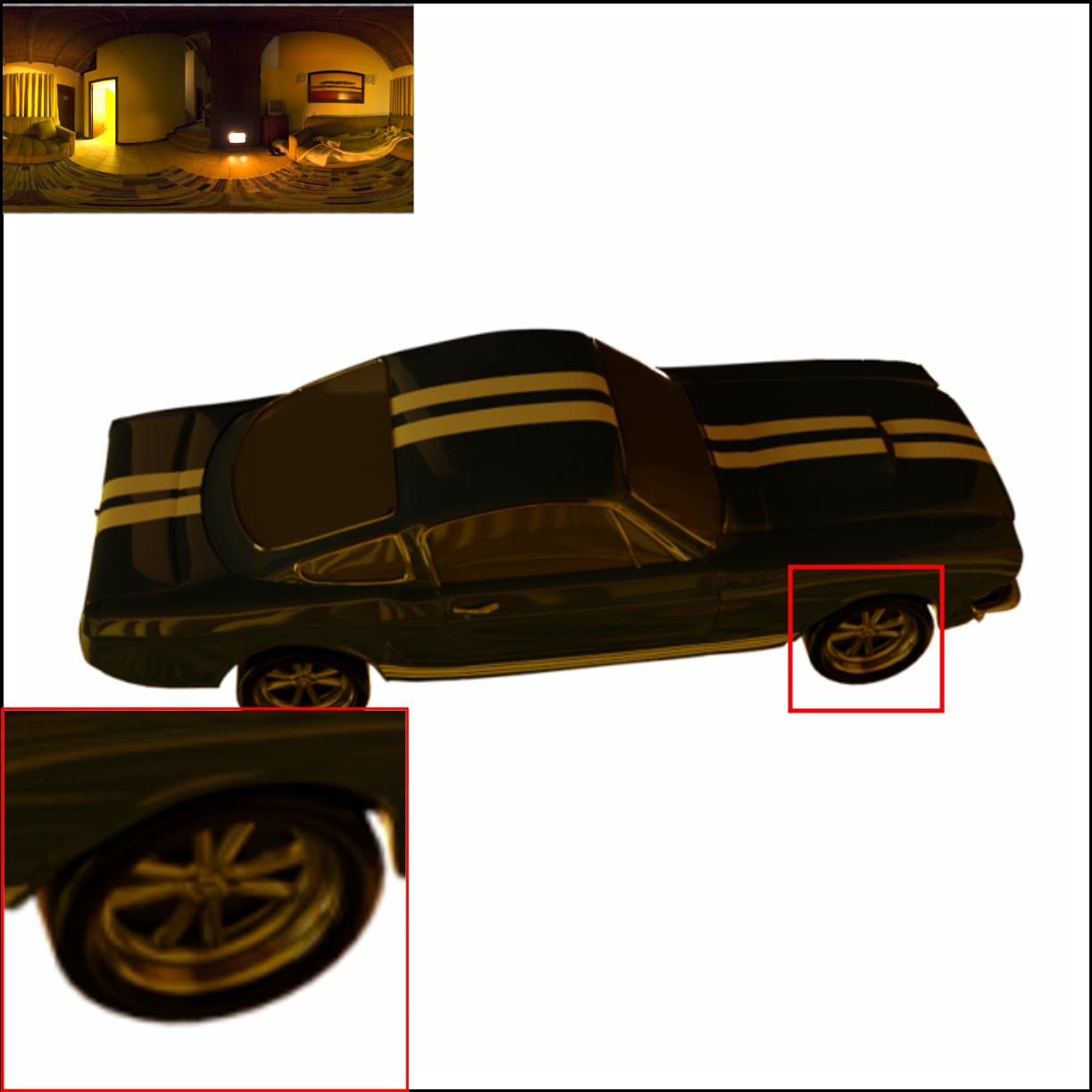} &
        \includegraphics[height=0.14\textwidth]{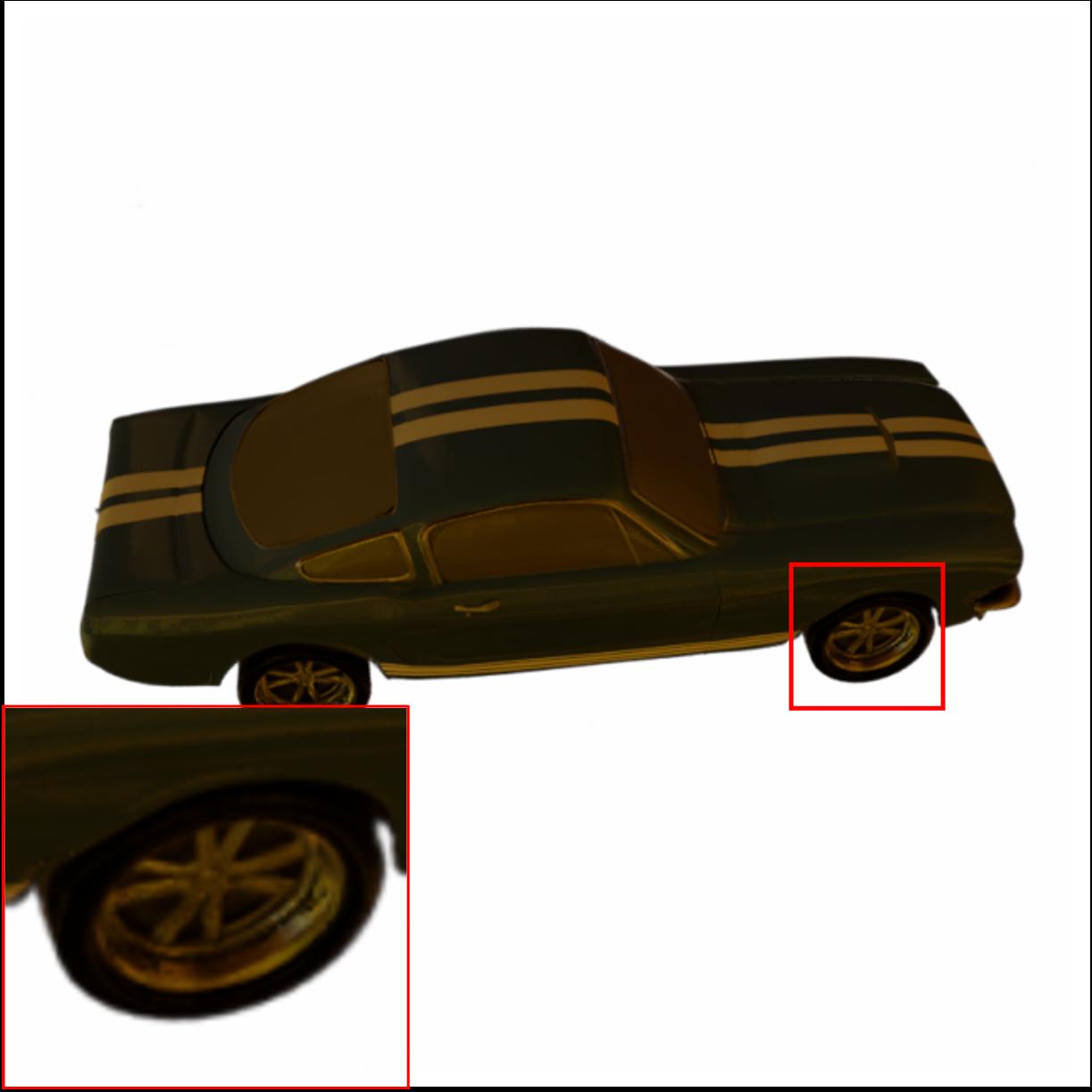} &
        \includegraphics[height=0.14\textwidth]{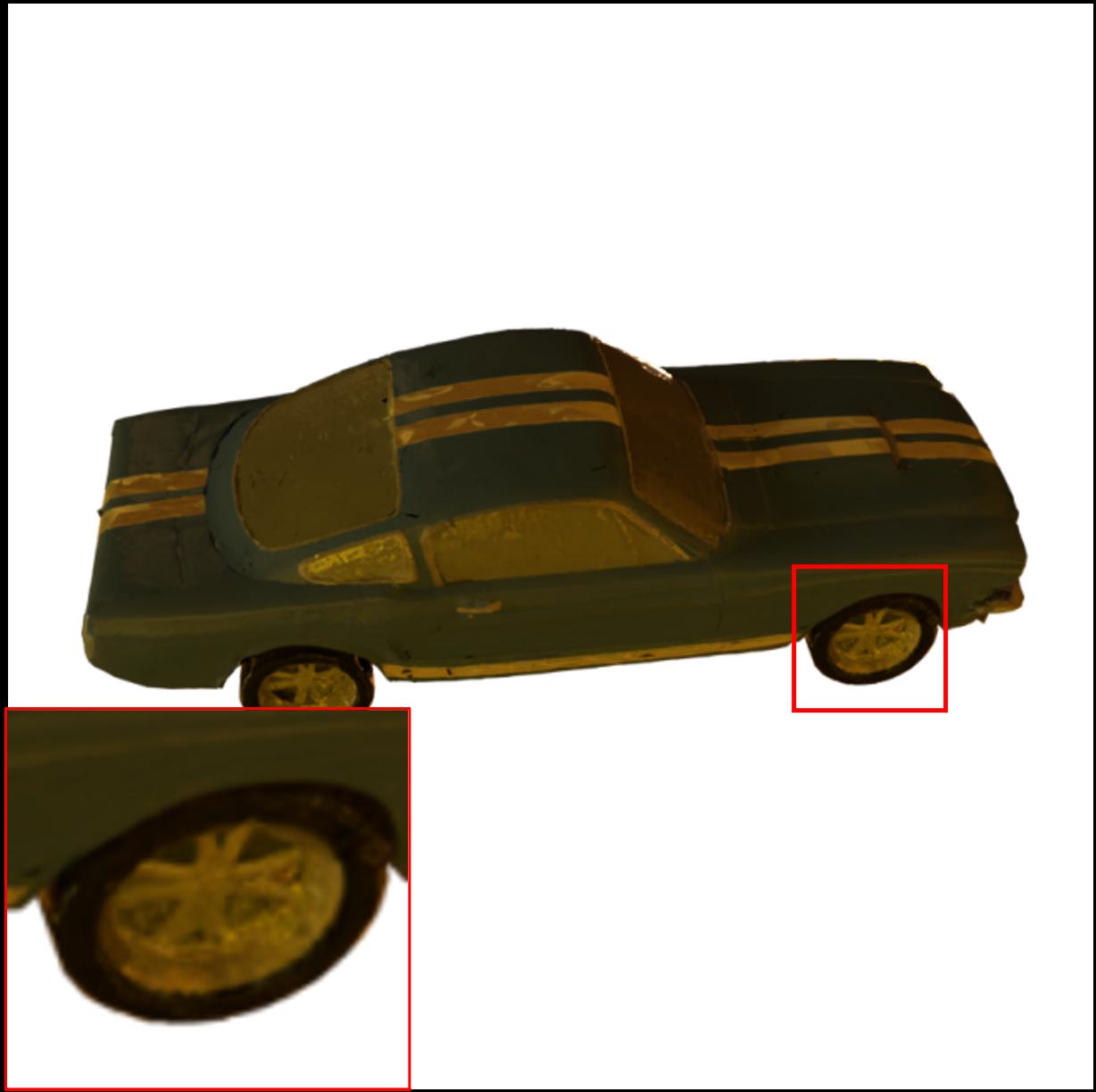} &
        \includegraphics[height=0.14\textwidth]{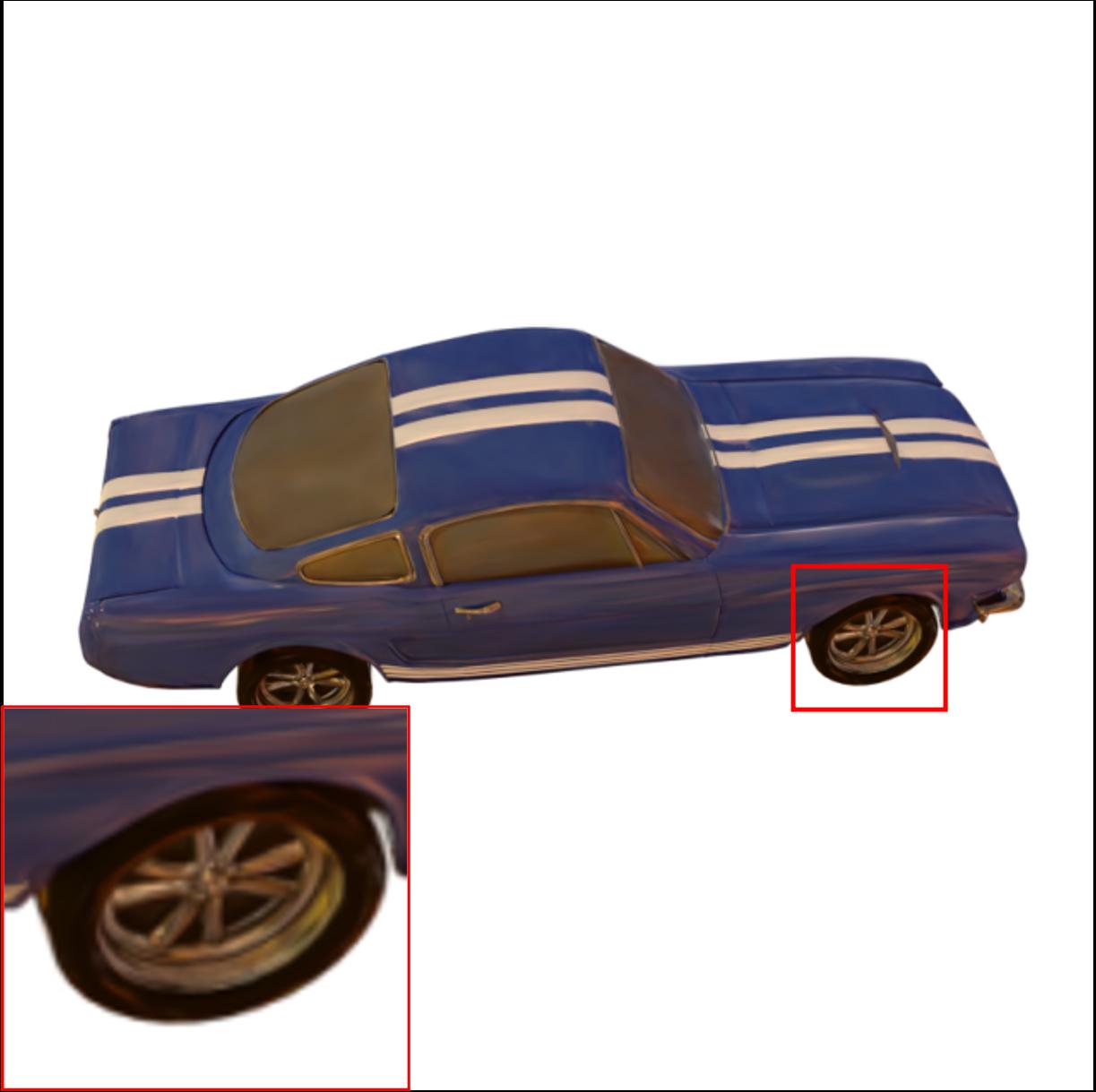} &
        \includegraphics[height=0.14\textwidth]{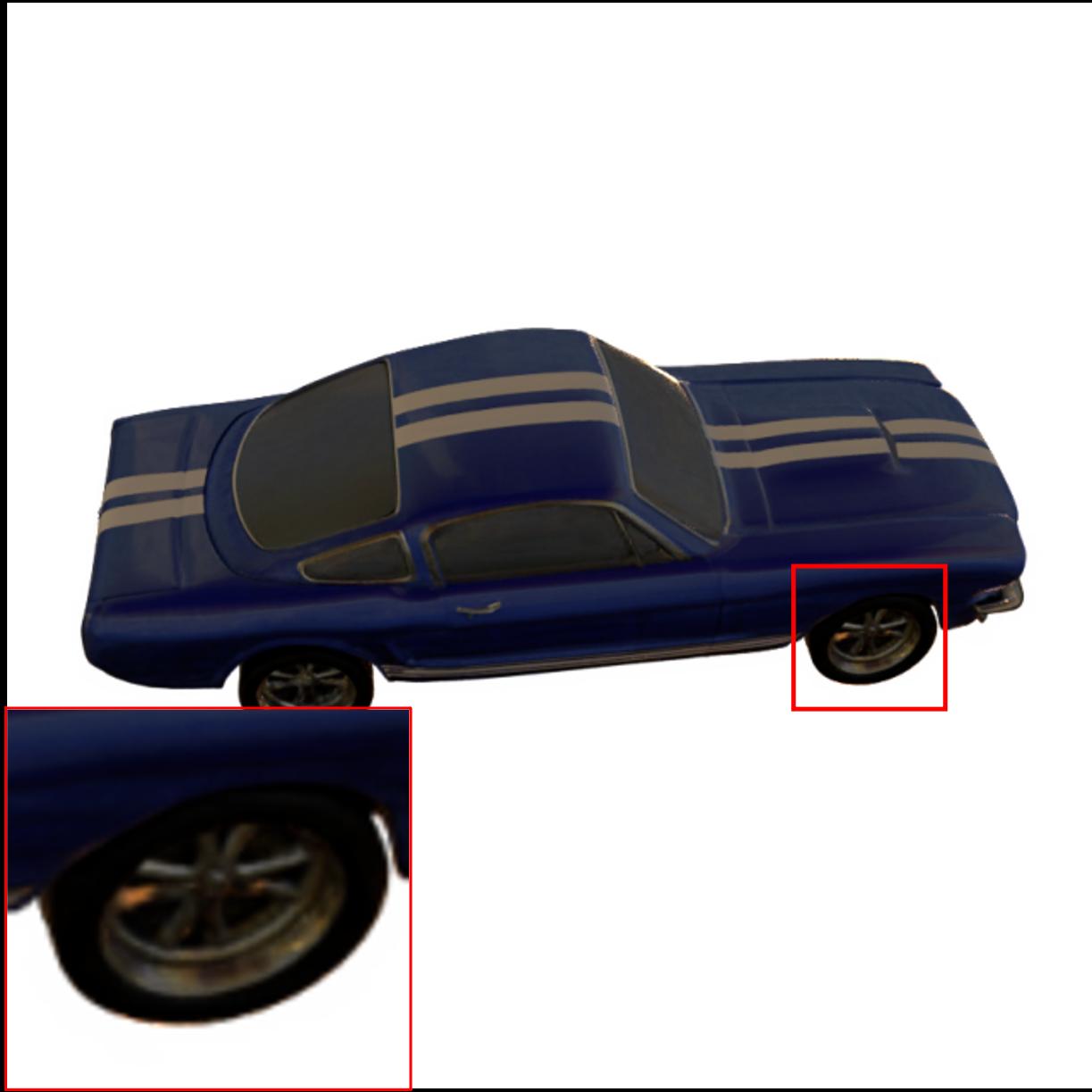} &
        \includegraphics[height=0.14\textwidth]{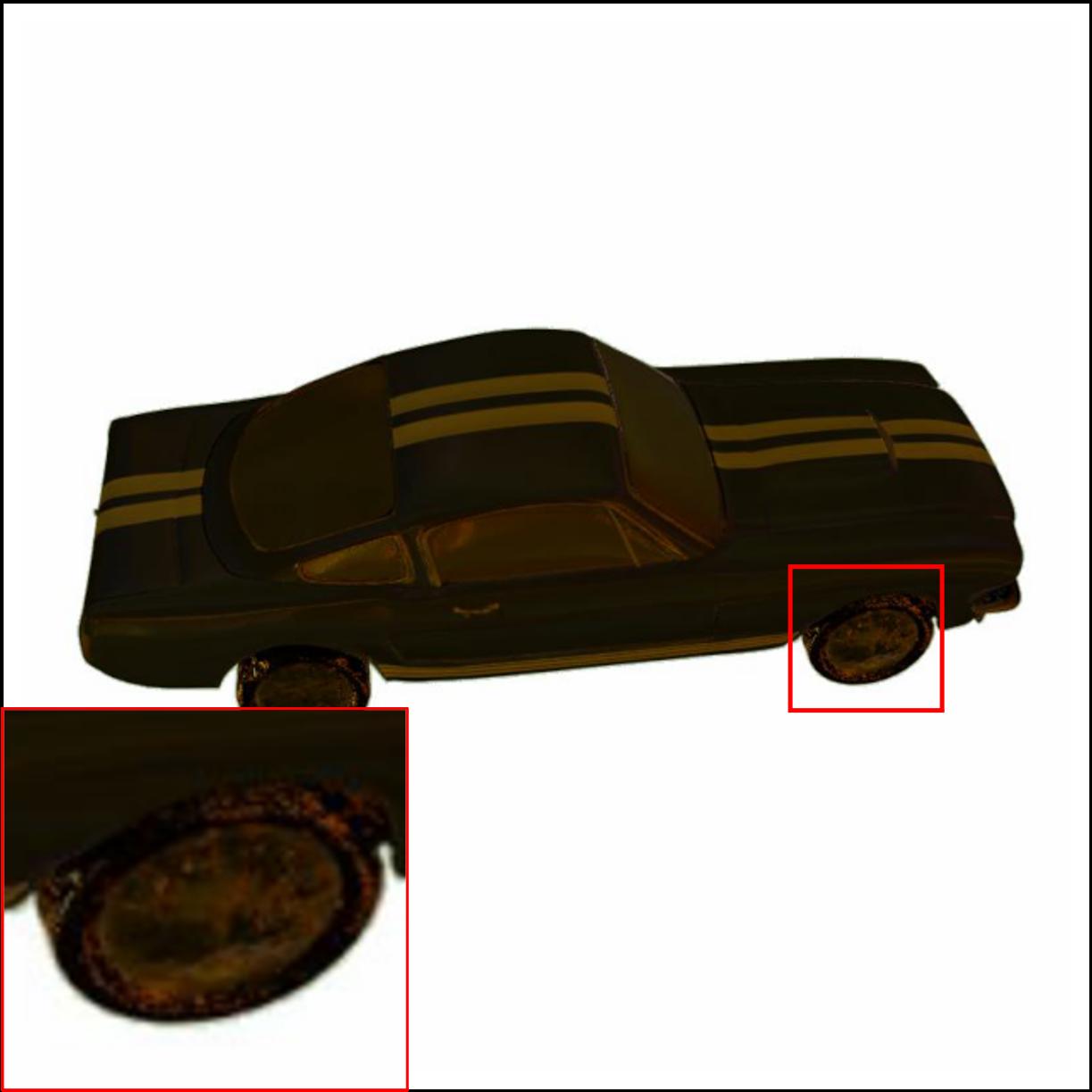}&
        \includegraphics[height=0.14\textwidth]{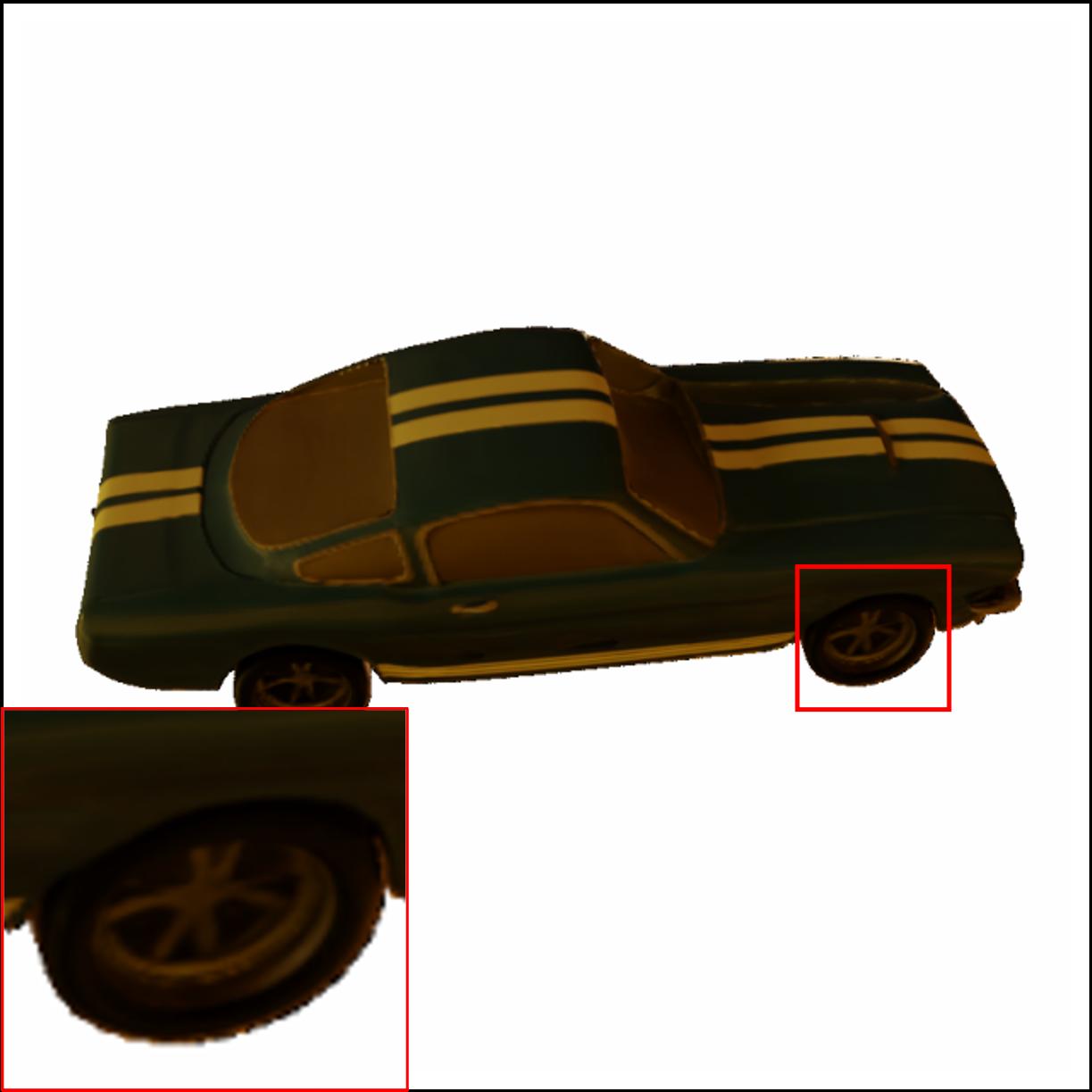}\\
        
        GT & Ours & NDRMC & GShader & NMF& ENVIDR & NeRO\\
    \end{tabular}
    \caption{\textbf{Qualitative comparisons on relighted synthetic scenes. From top to bottom: toaster, helmet, car.} We can observe that other methods either have blurry, color-shifted results or aliasing, noisy effect under unseen illumination. }
    \label{fig:relight}
\end{figure*}

\begin{figure*}[p]
    \centering
    \setlength\tabcolsep{1pt}
    \begin{tabular}{ccccccc}
        \includegraphics[height=0.14\textwidth]{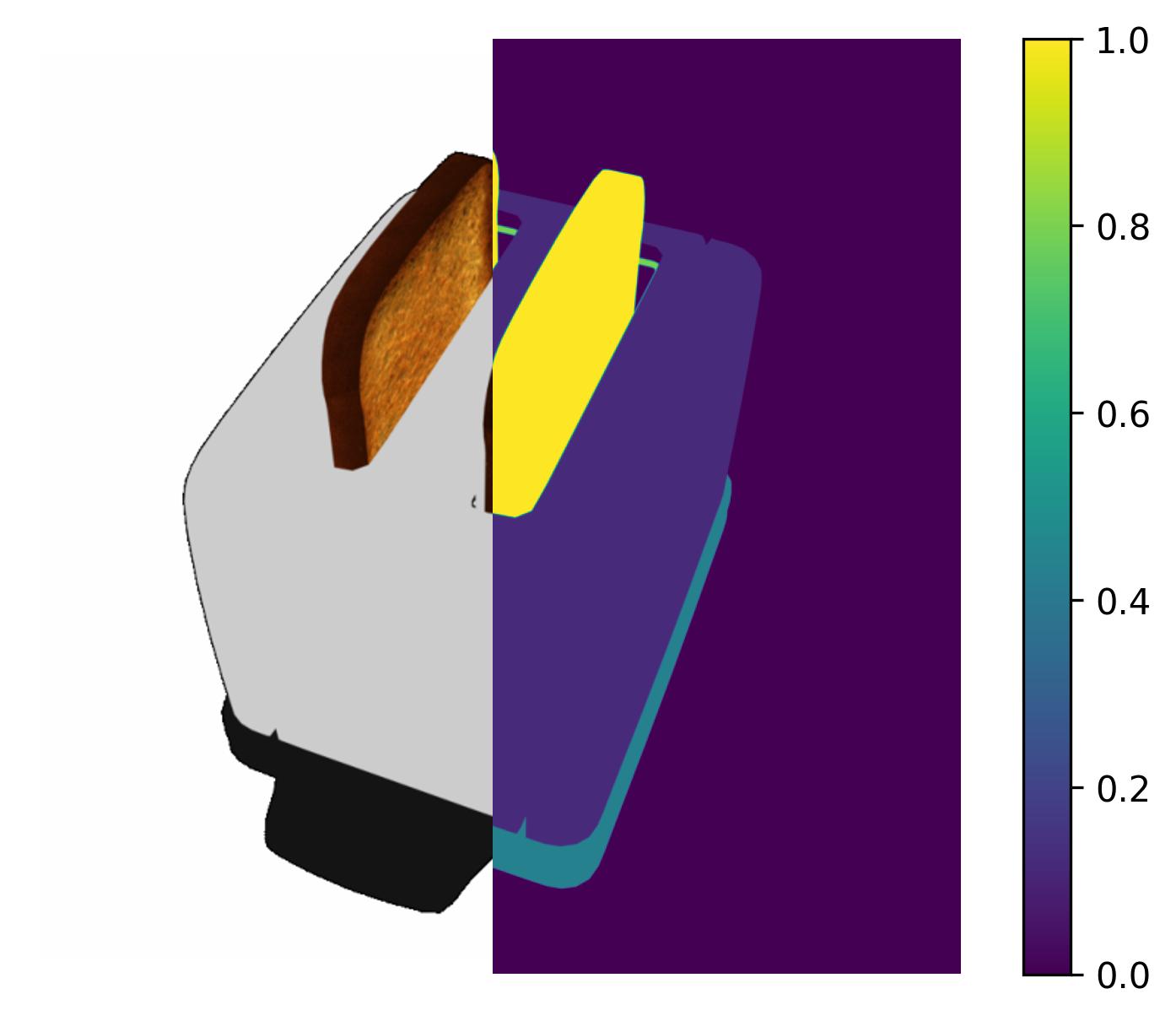} &
        \includegraphics[height=0.14\textwidth]{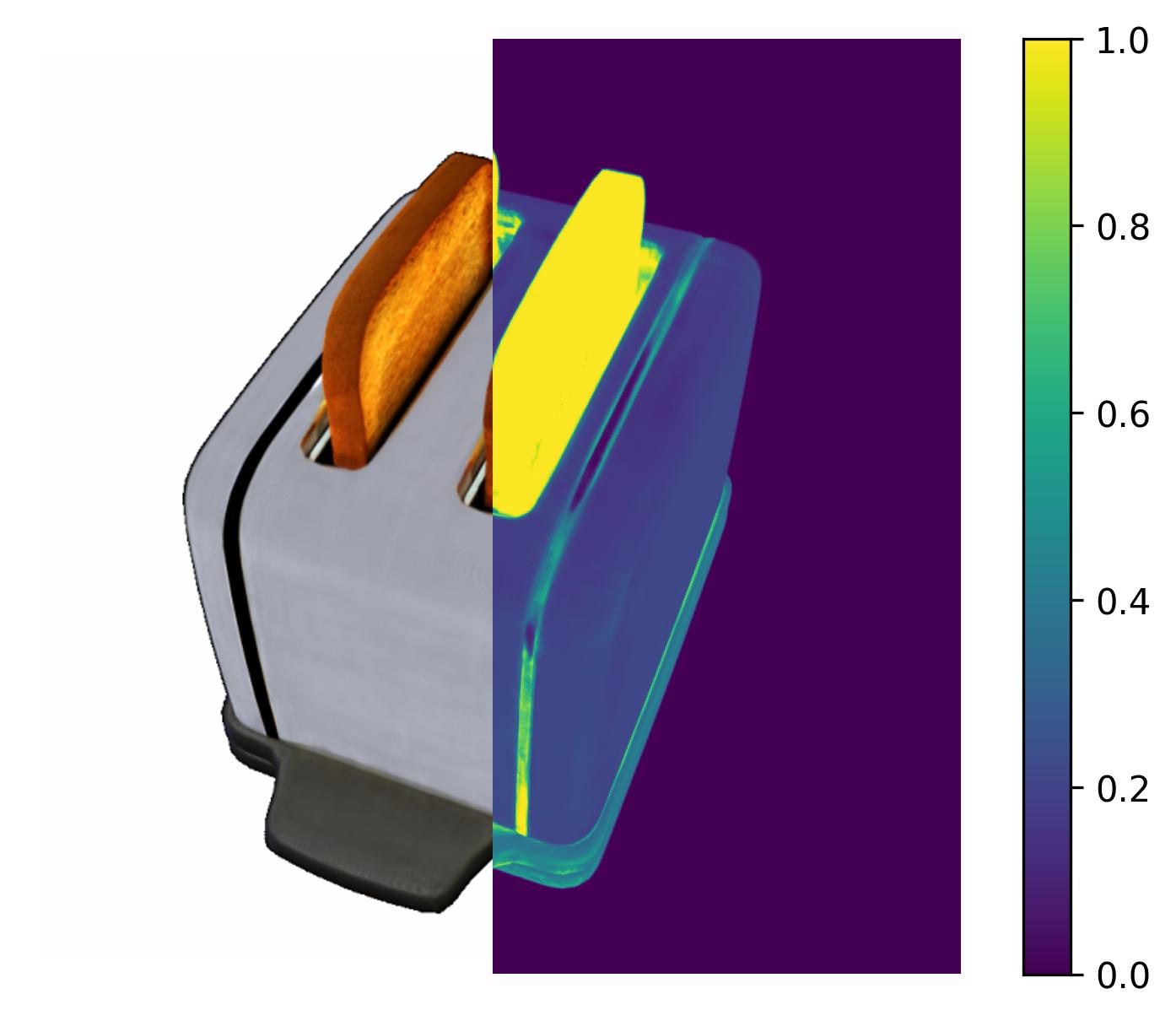} &
        \includegraphics[height=0.14\textwidth]{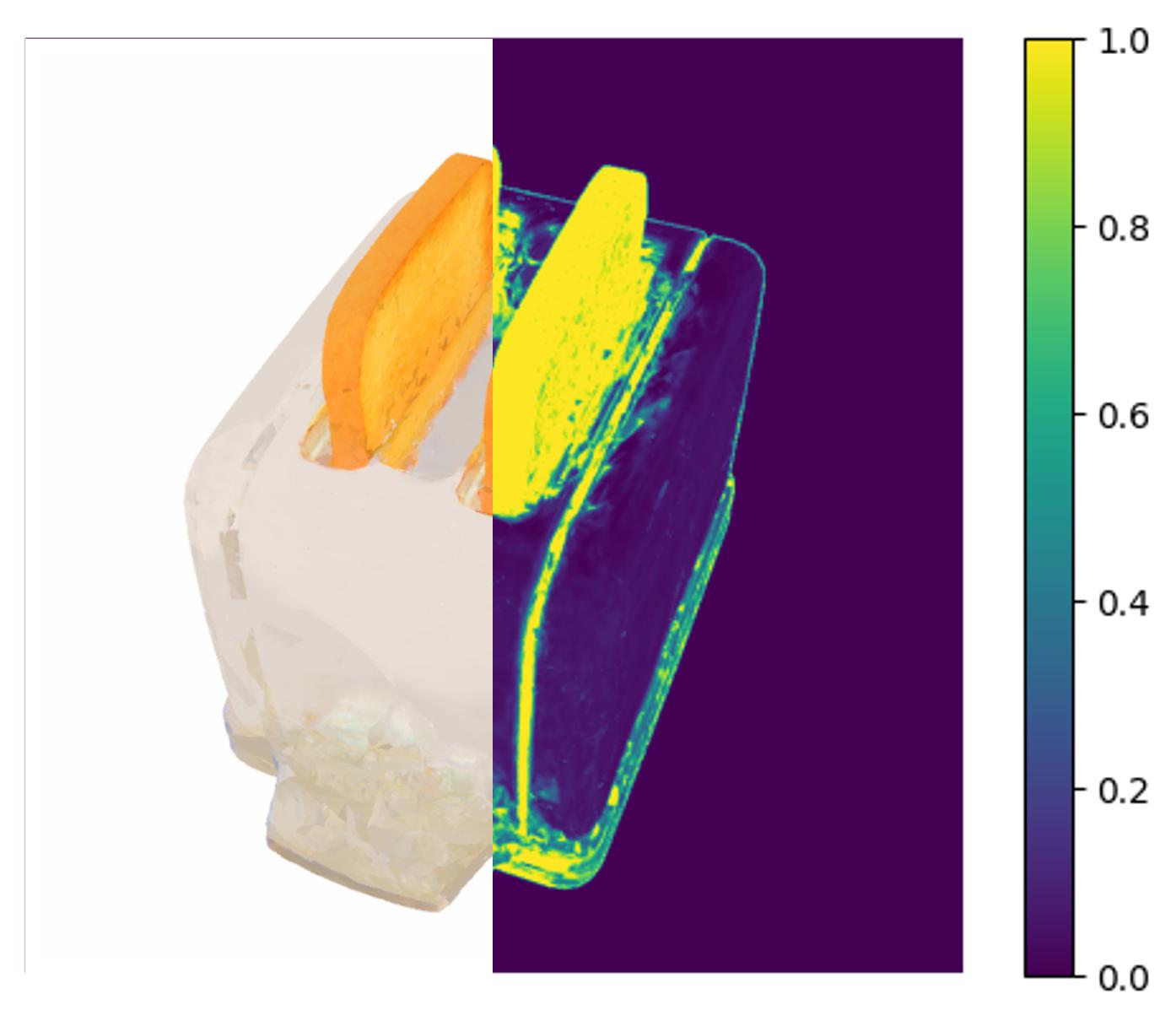} &
        \includegraphics[height=0.14\textwidth]{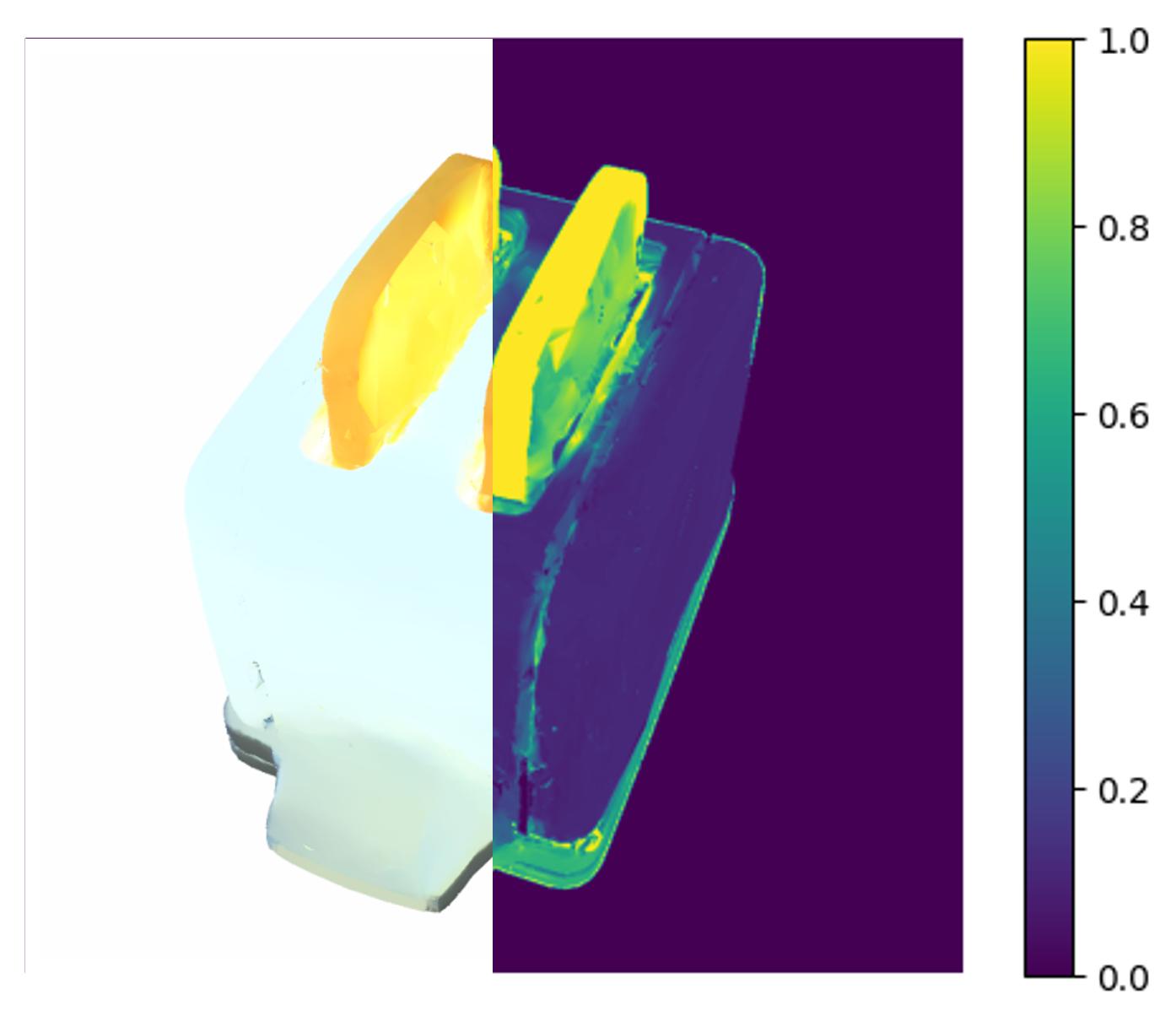} &
        \includegraphics[height=0.14\textwidth]{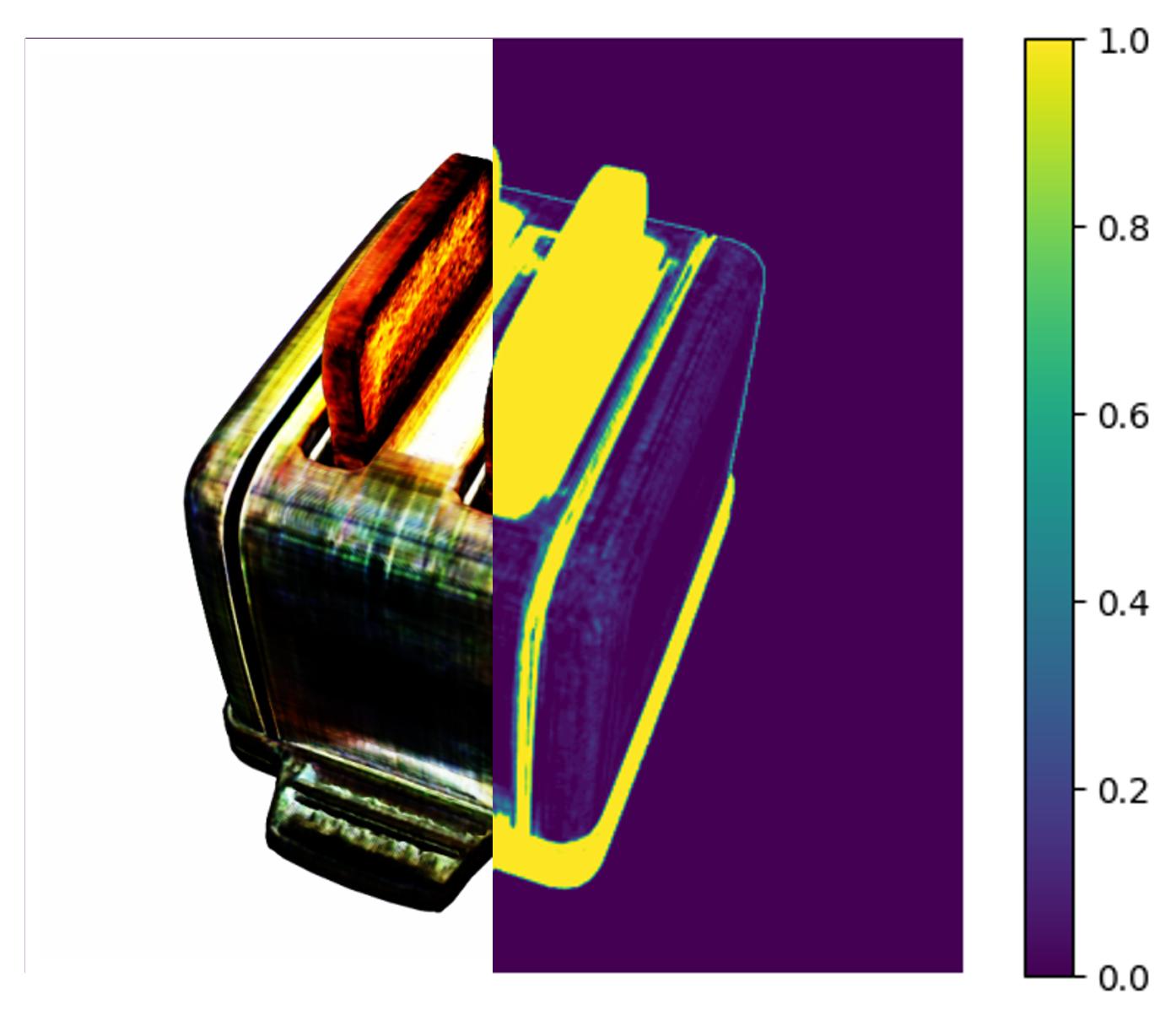} &
        \includegraphics[height=0.14\textwidth]{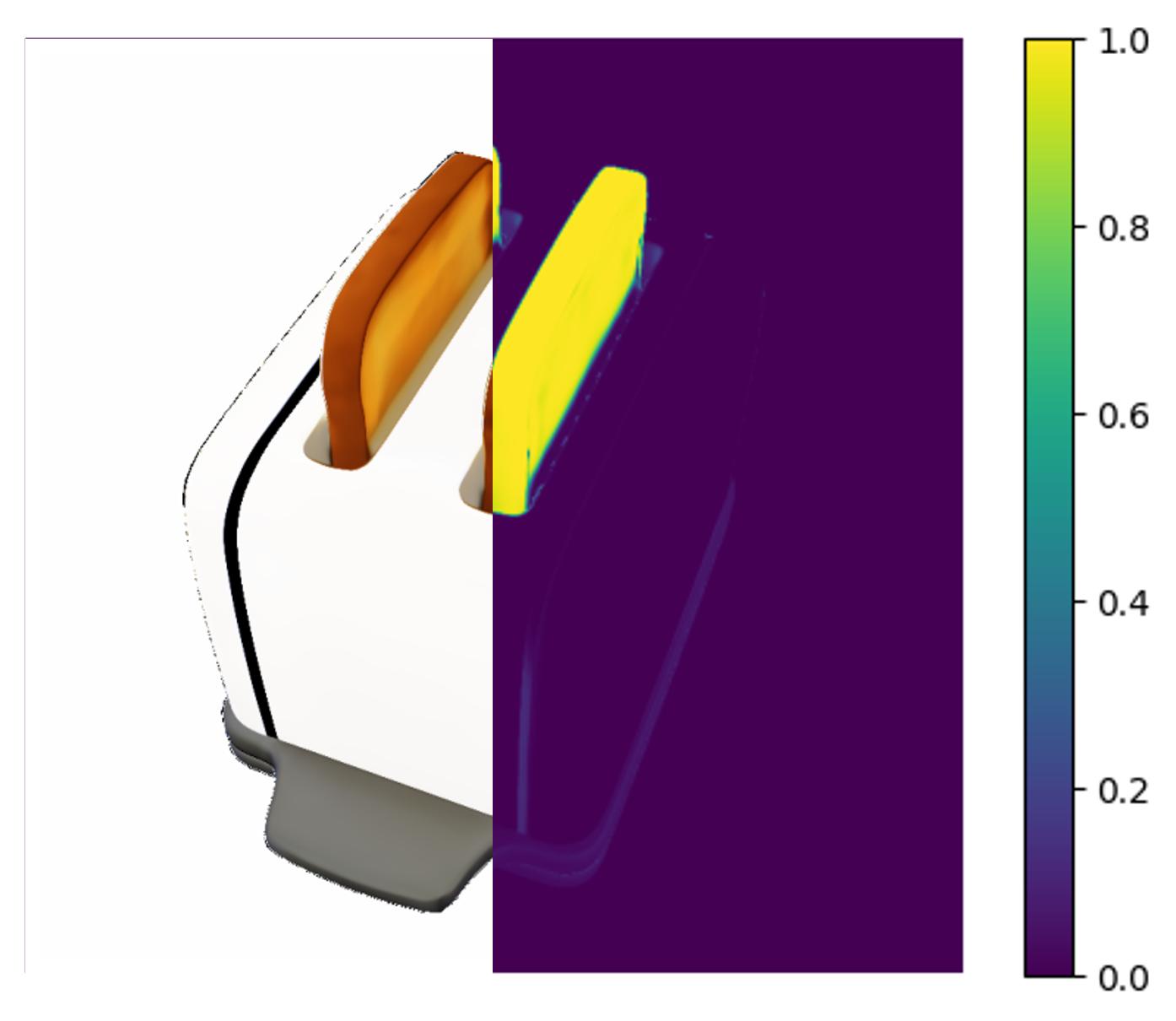} \\
         \includegraphics[height=0.14\textwidth]{figures/figure_only/material/helmet_gt_material.jpeg} &
        \includegraphics[height=0.14\textwidth]{figures/figure_only/material/helmet_ours_material.jpeg} &
        \includegraphics[height=0.14\textwidth]{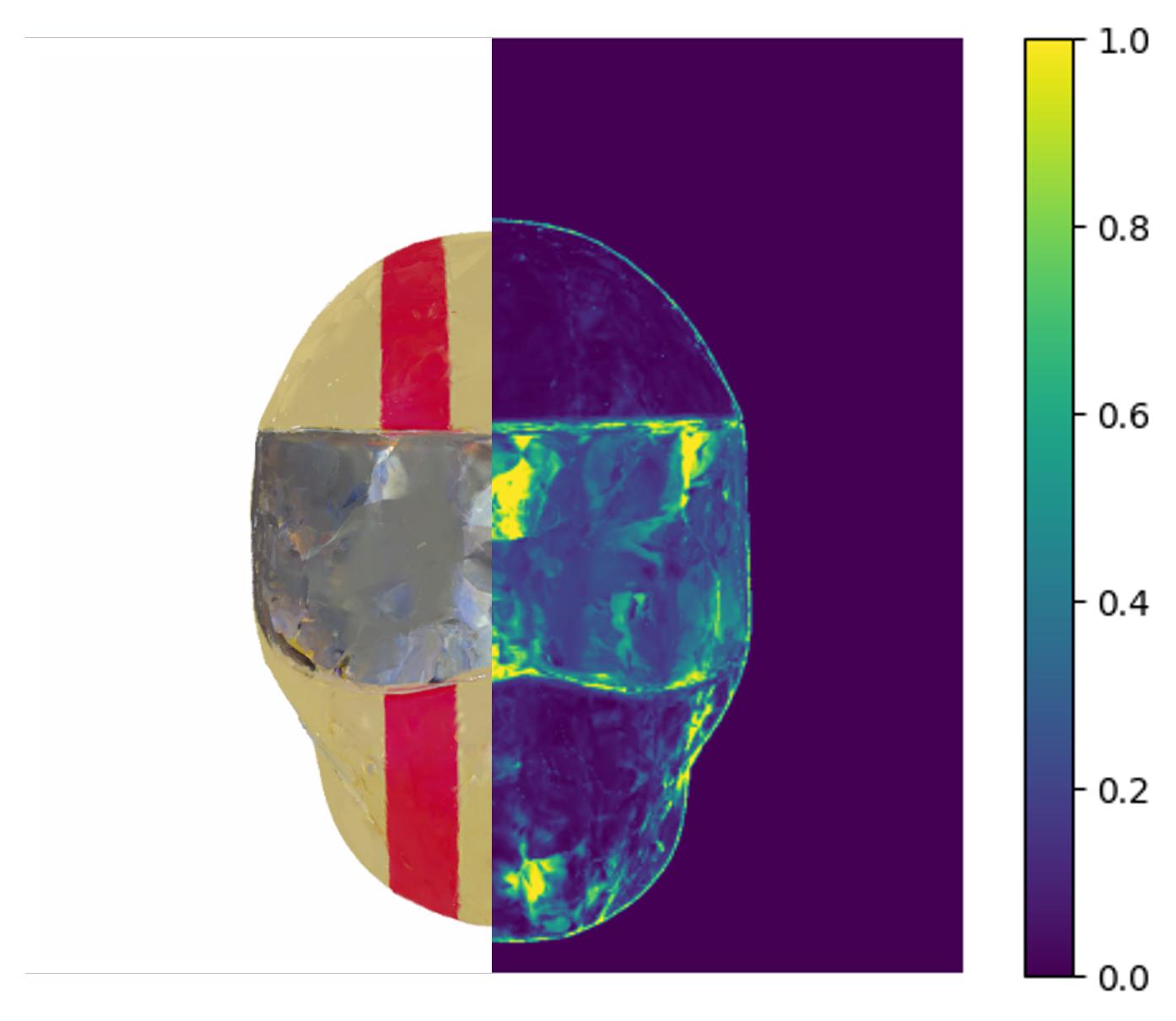} &
        \includegraphics[height=0.14\textwidth]{figures/figure_only/material/helmet_ndrmc_material.jpeg} &
        \includegraphics[height=0.14\textwidth]{figures/figure_only/material/helmet_nmf_material.jpeg} &
        \includegraphics[height=0.14\textwidth]{figures/figure_only/material/helmet_nero_material.jpeg} \\
        \includegraphics[height=0.14\textwidth]{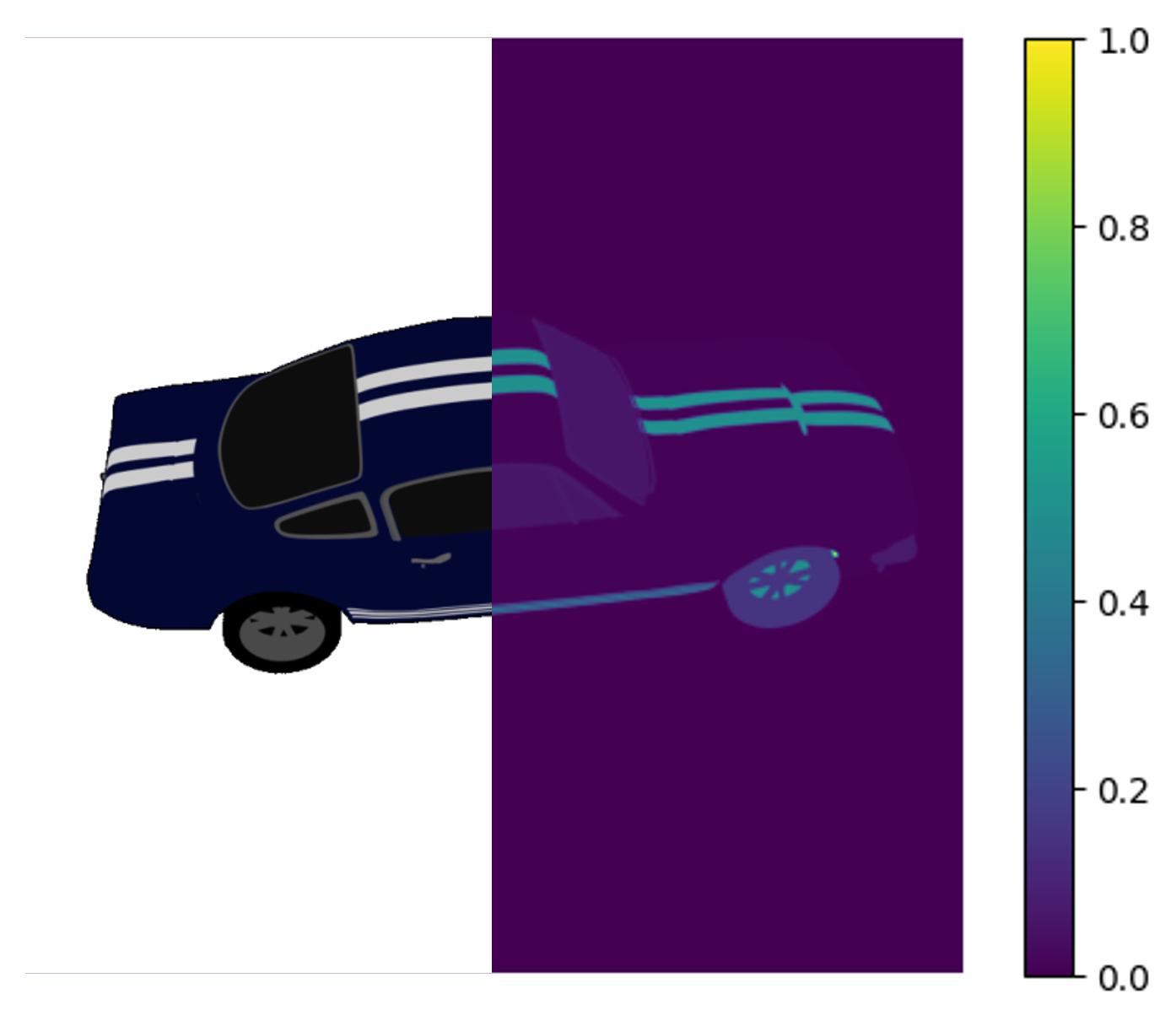} &
        \includegraphics[height=0.14\textwidth]{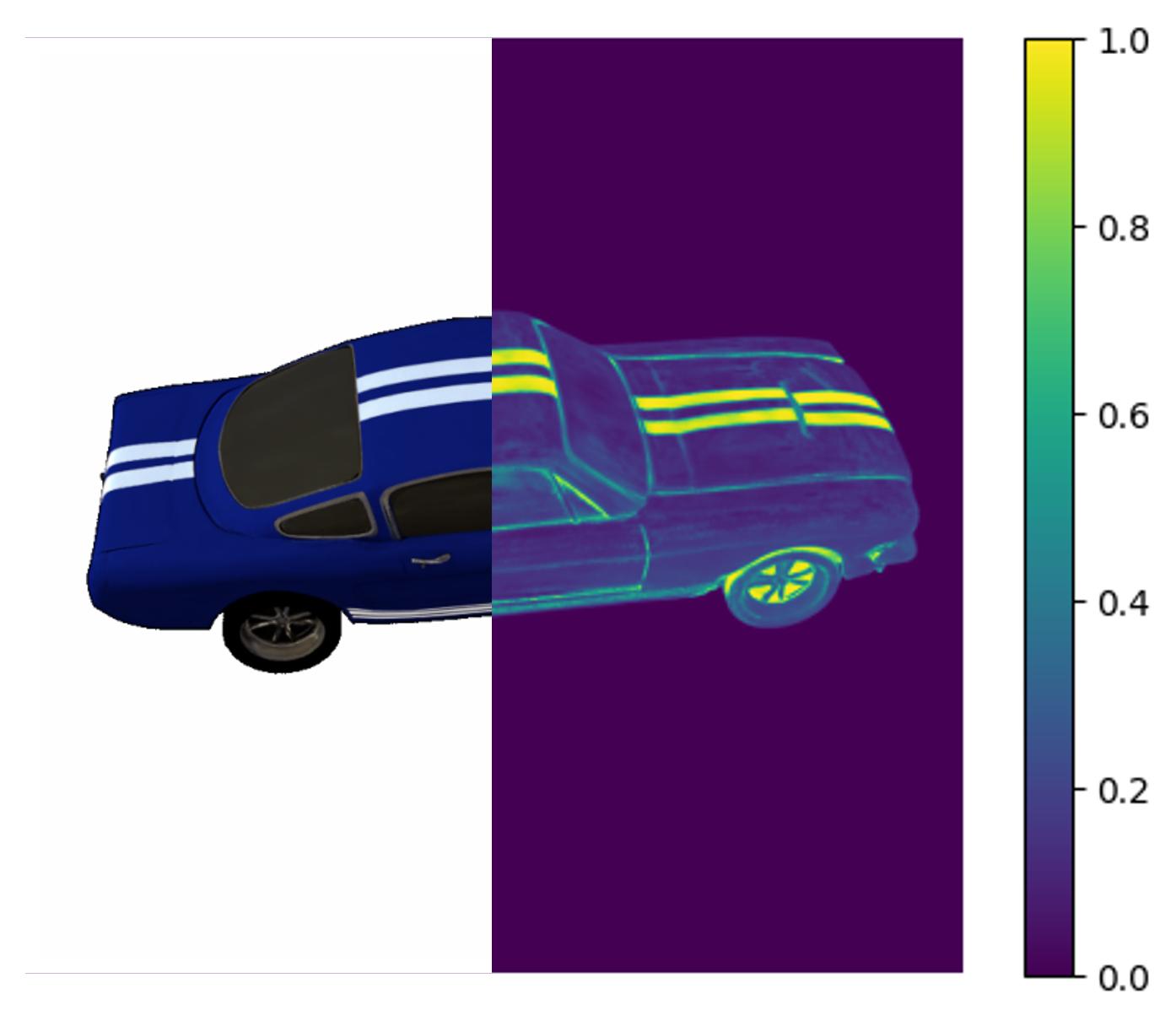} &
        \includegraphics[height=0.14\textwidth]{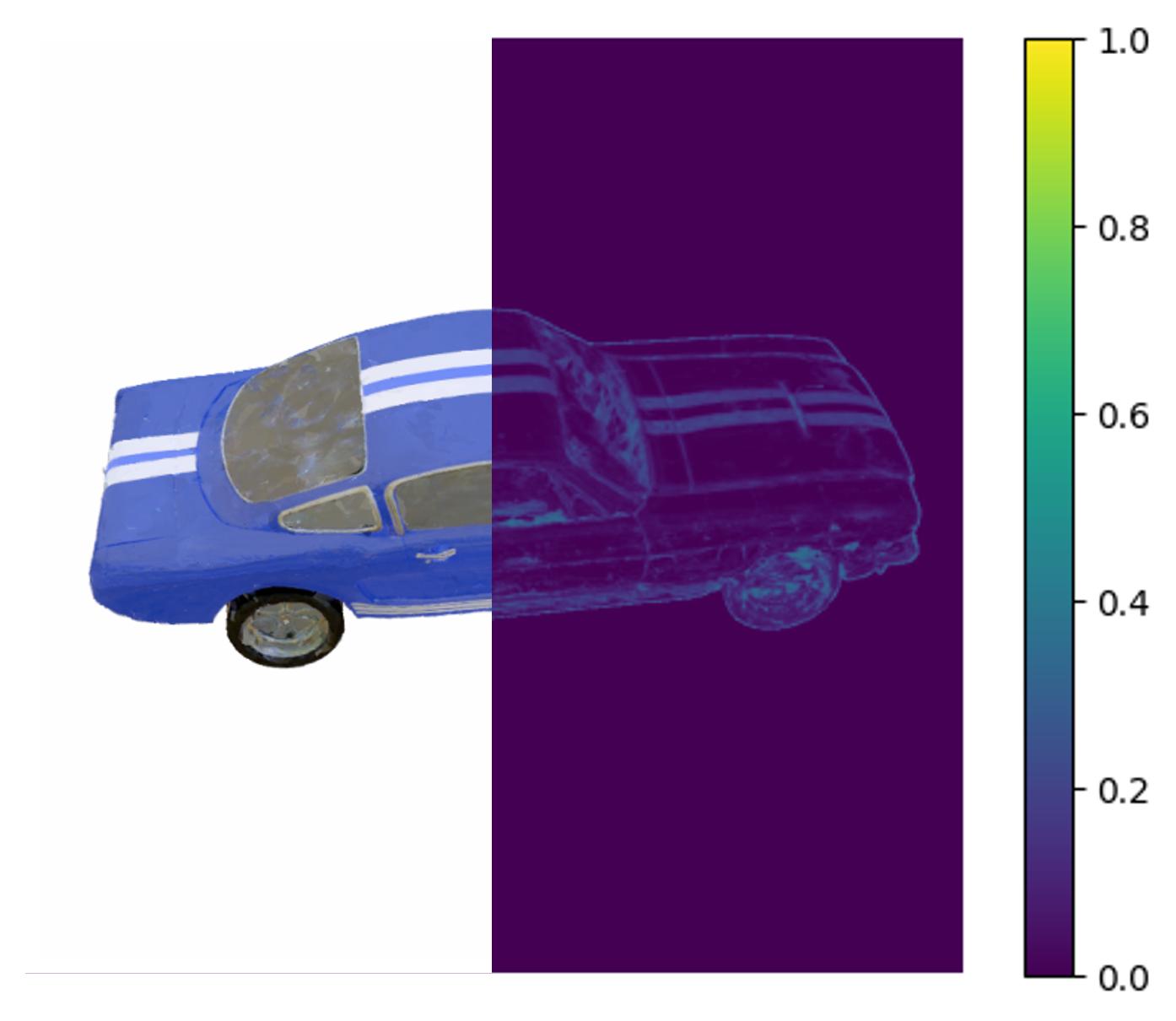} &
        \includegraphics[height=0.14\textwidth]{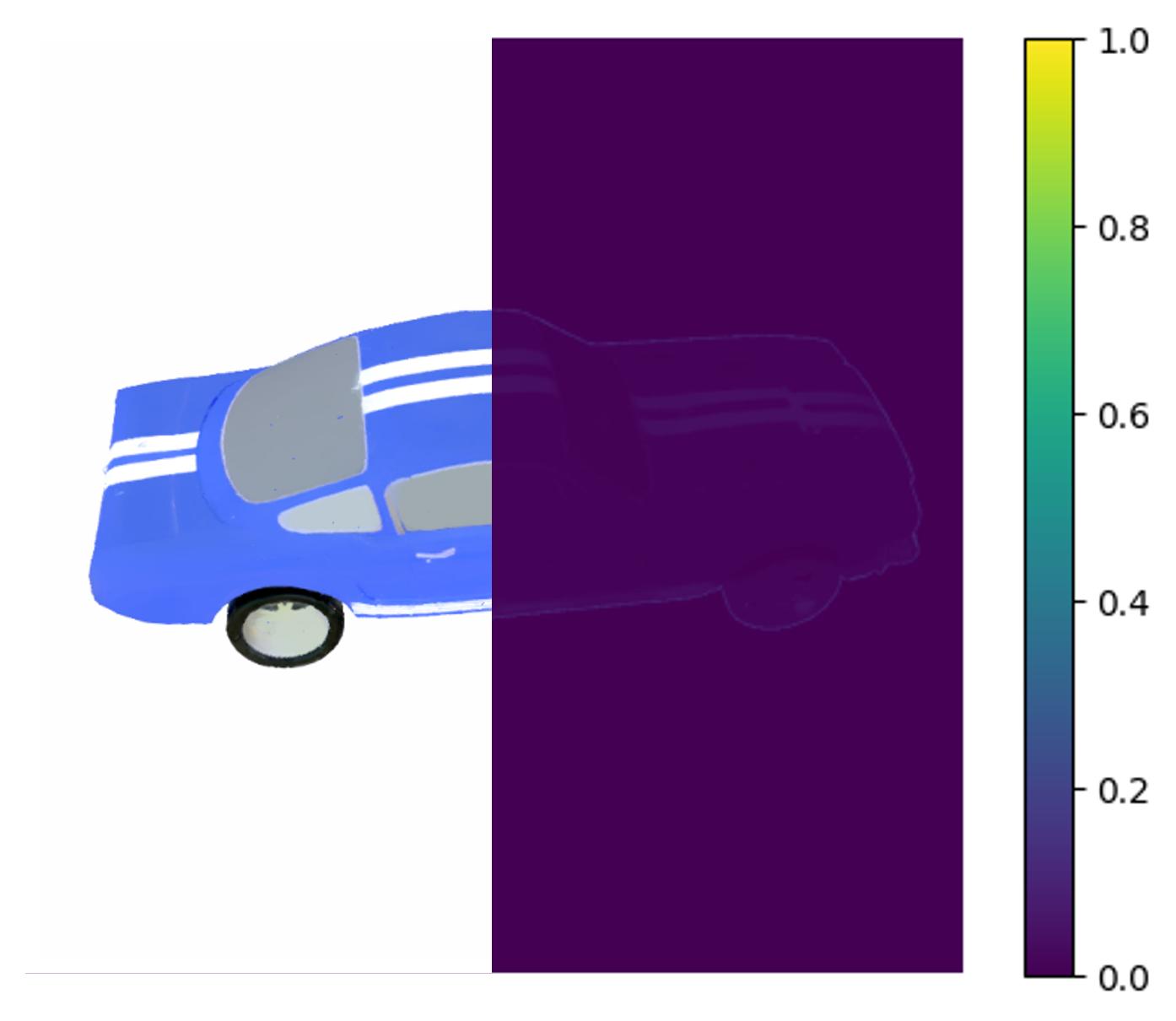} &
        \includegraphics[height=0.14\textwidth]{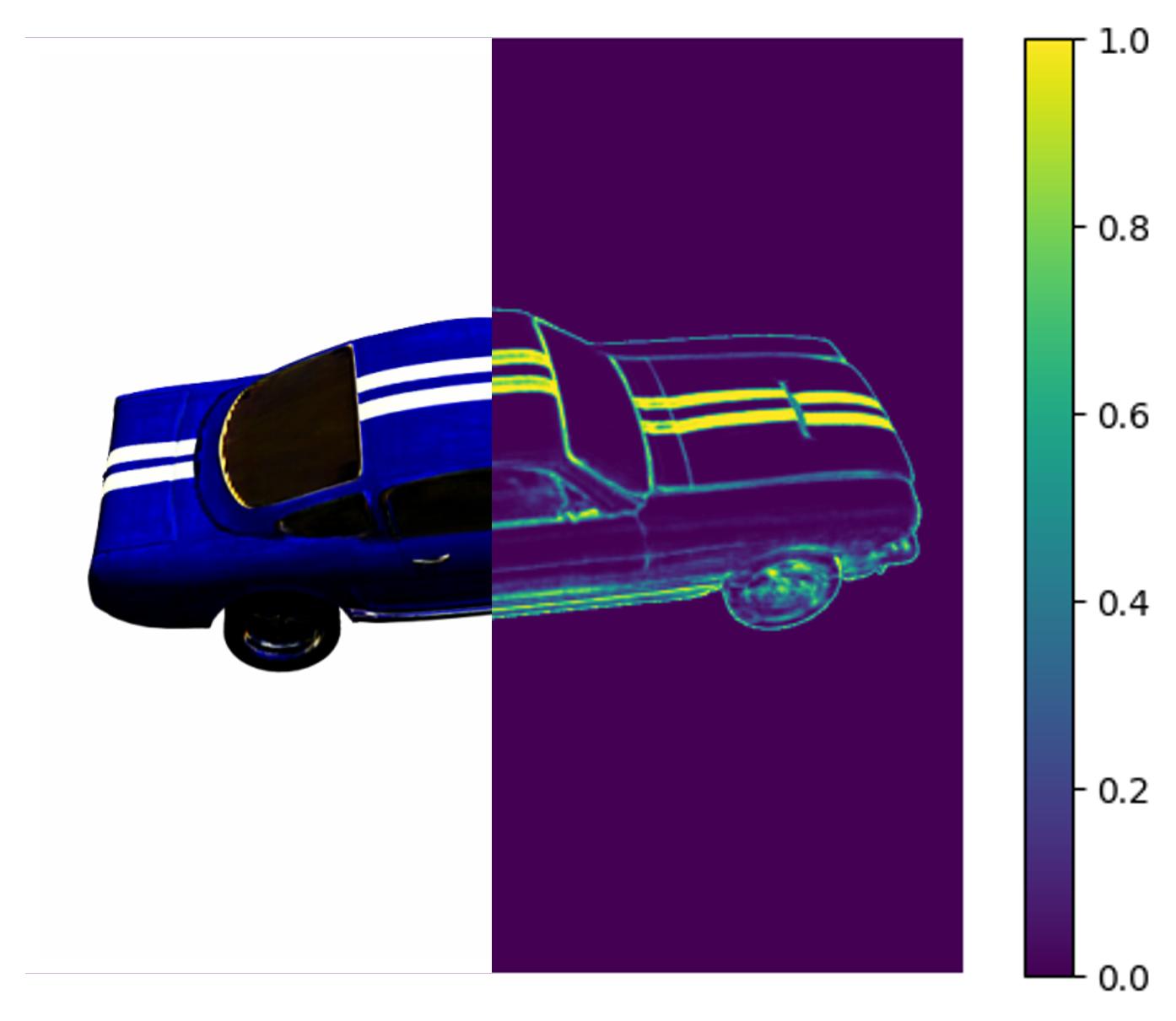} &
        \includegraphics[height=0.14\textwidth]{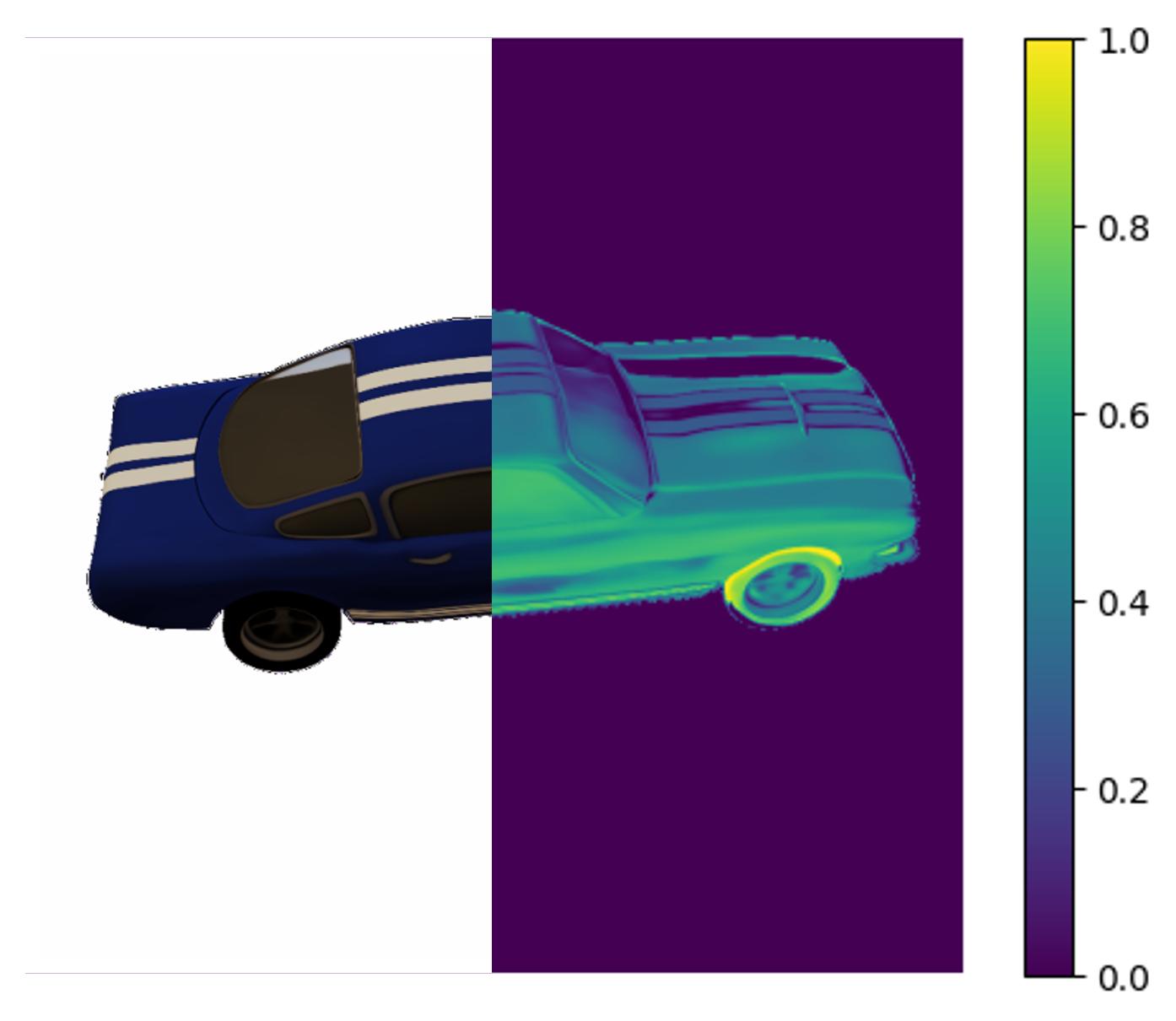} \\
        GT & Ours & NDR & NDRMC & NMF & NeRO\\
    \end{tabular}
    \caption{\textbf{Qualitative comparisons on material estimation. From top to bottom: toaster, helmet, car. In each figure, we show albedo on the left and roughness on the right. } }
    \label{fig:material}
\end{figure*}

\begin{figure*}[h]
    \centering
    \setlength\tabcolsep{1pt}
    \begin{tabular}{ccccccc}
         \includegraphics[height=0.14\textwidth]{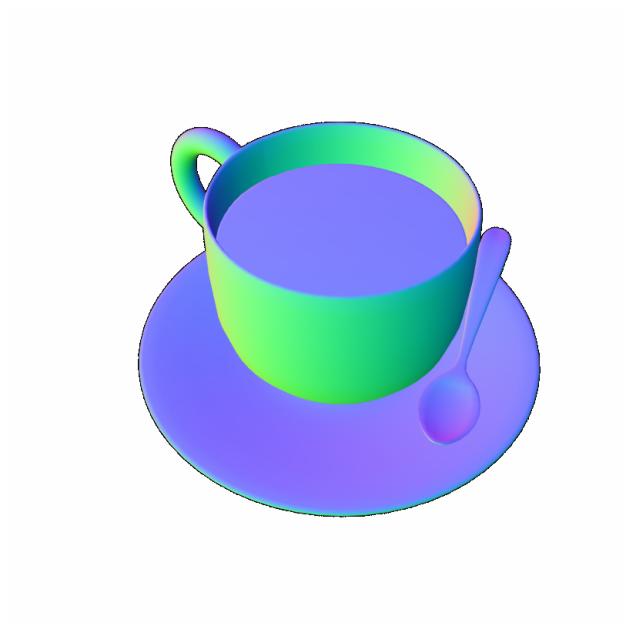}&
        \includegraphics[height=0.14\textwidth]{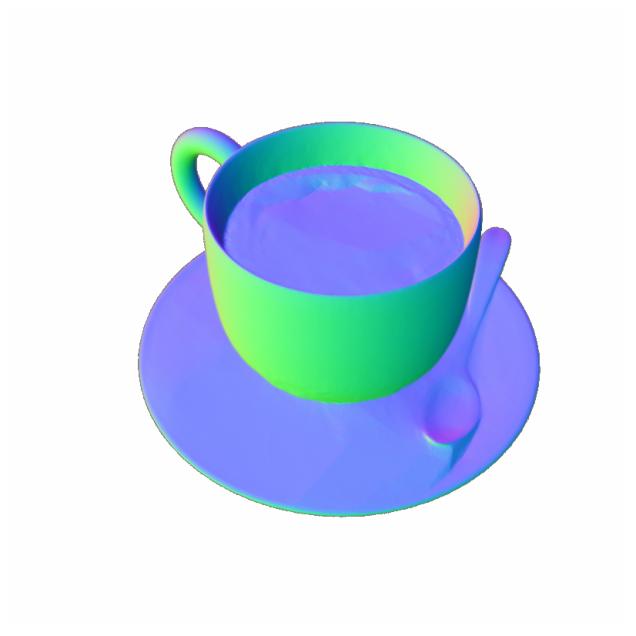} &
        \includegraphics[height=0.14\textwidth]{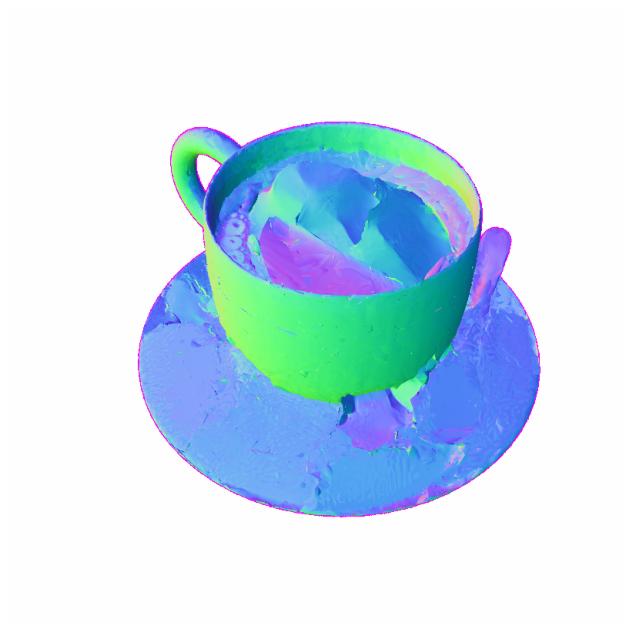} &
        \includegraphics[height=0.14\textwidth]{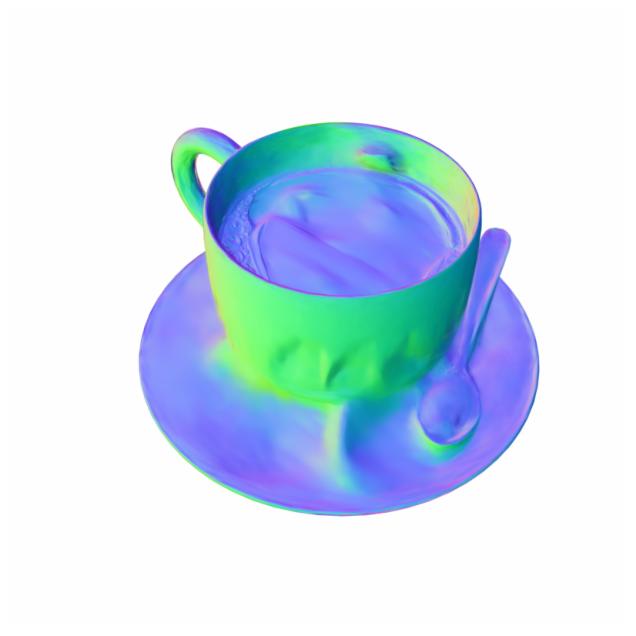} &
        \includegraphics[height=0.14\textwidth]{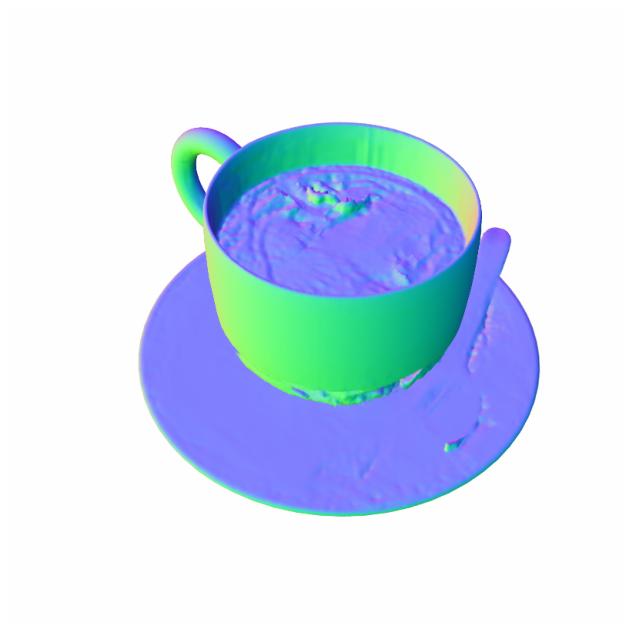} &
        \includegraphics[height=0.14\textwidth]{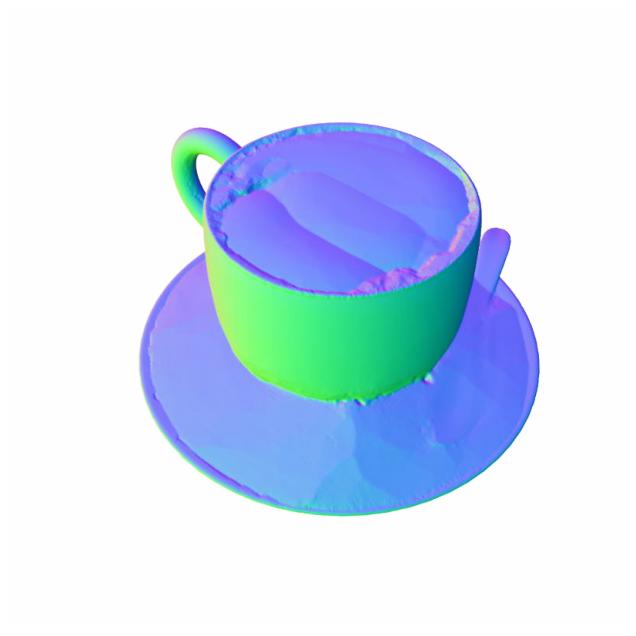} &
        \includegraphics[height=0.14\textwidth]{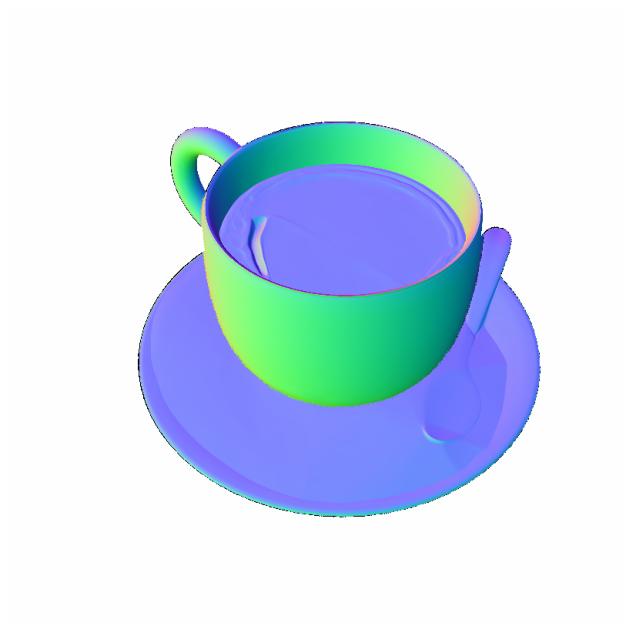} \\
        \includegraphics[height=0.14\textwidth]{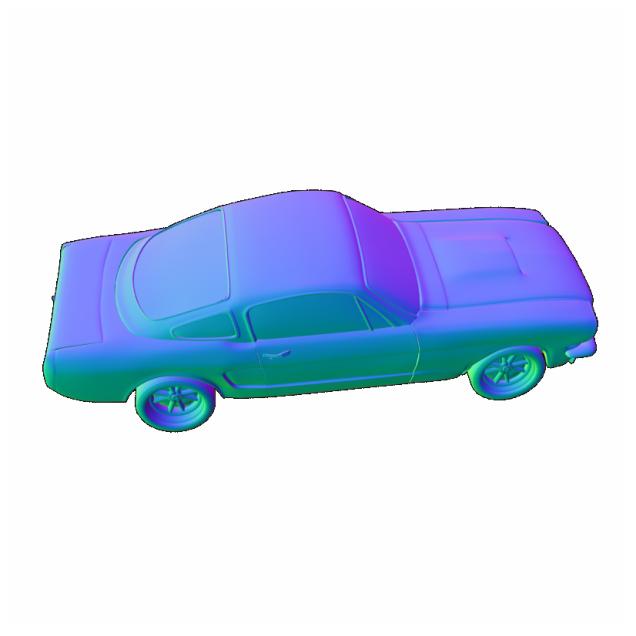}&
        \includegraphics[height=0.14\textwidth]{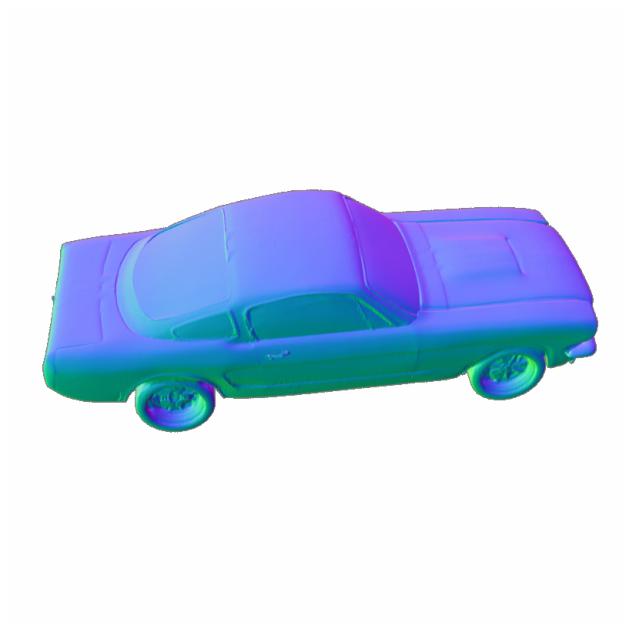} &
        \includegraphics[height=0.14\textwidth]{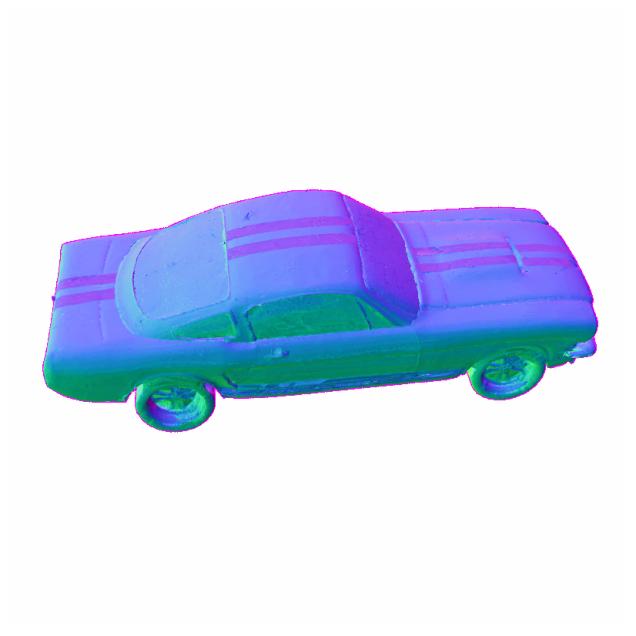} &
        \includegraphics[height=0.14\textwidth]{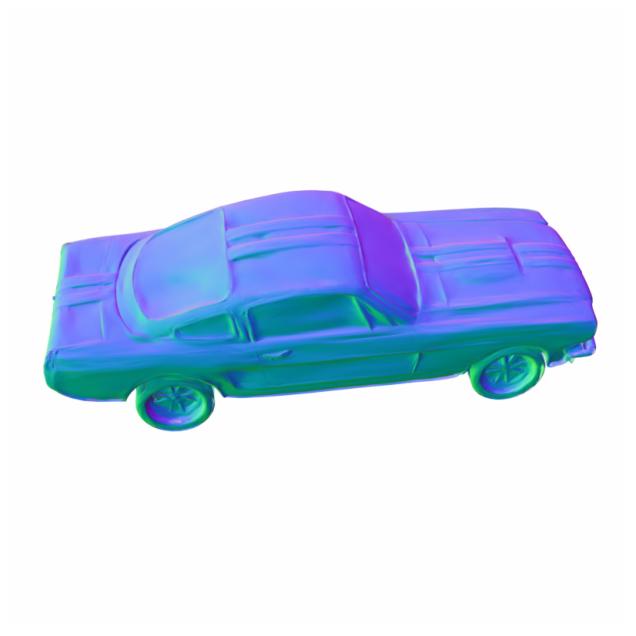} &
        \includegraphics[height=0.14\textwidth]{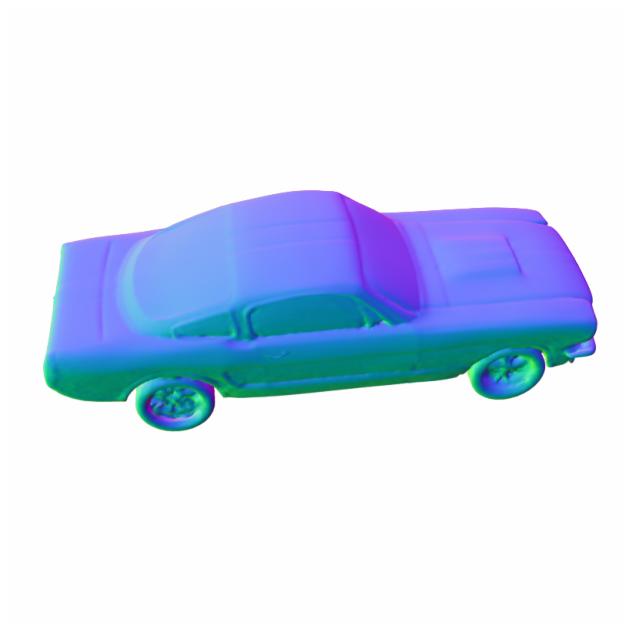} &
        \includegraphics[height=0.14\textwidth]{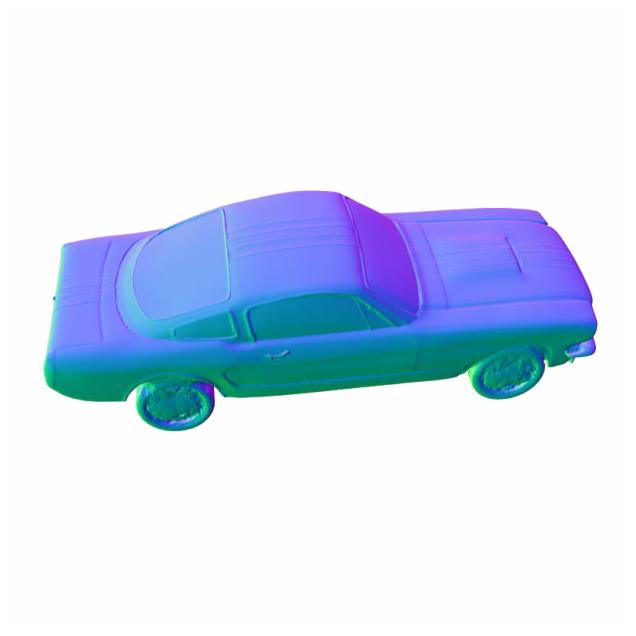} &
        \includegraphics[height=0.14\textwidth]{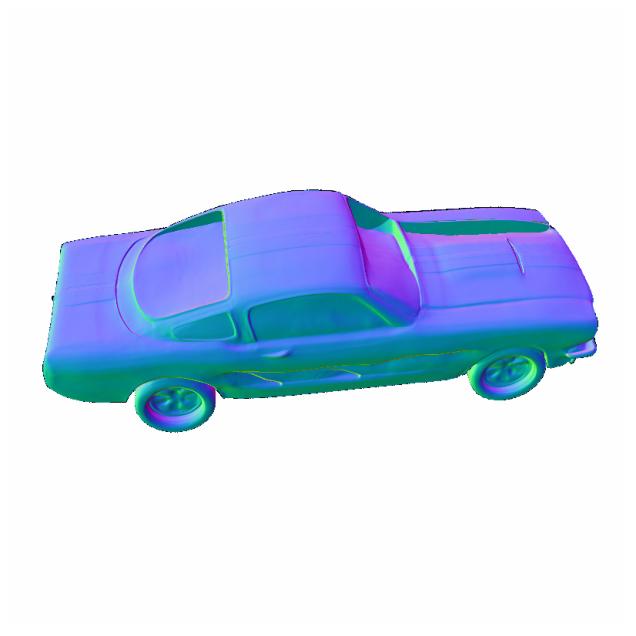} \\
        GT & Ours & NDRMC & GShader & NMF& ENVIDR & NeRO\\
    \end{tabular}
    \caption{\textbf{Qualitative comparisons on normal reconstruction. From top to bottom: toaster, coffee, car.} }
    \label{fig:normal}
\end{figure*}

\begin{figure*}[h]
    \centering
    \setlength\tabcolsep{1pt}
    \settoheight{\tempdima}{\includegraphics[width=0.1\textwidth]{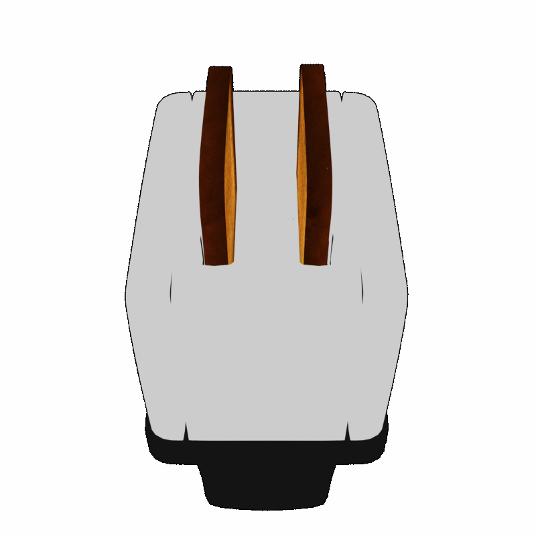}}
    \begin{tabular}{ccccc}
         & gt & ours (15k) & w.o. info (5k)&w.o. info (15k)\\
        \rowname{albedo}&
        \includegraphics[height=0.1\textwidth]{figures/info_share/toaster_gt_albedo.jpeg} &
        \includegraphics[height=0.1\textwidth]{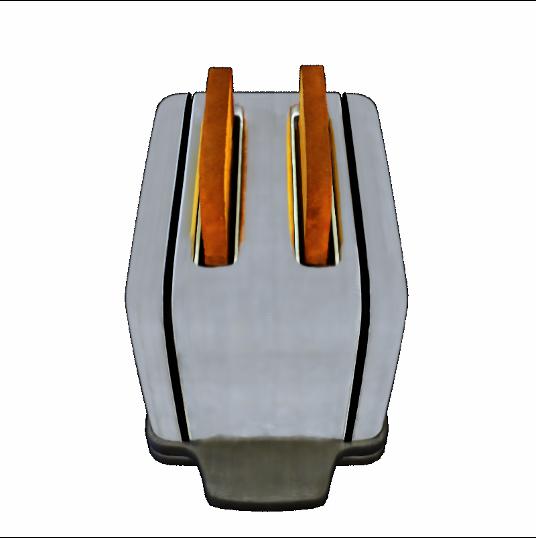} &
        \includegraphics[height=0.1\textwidth]{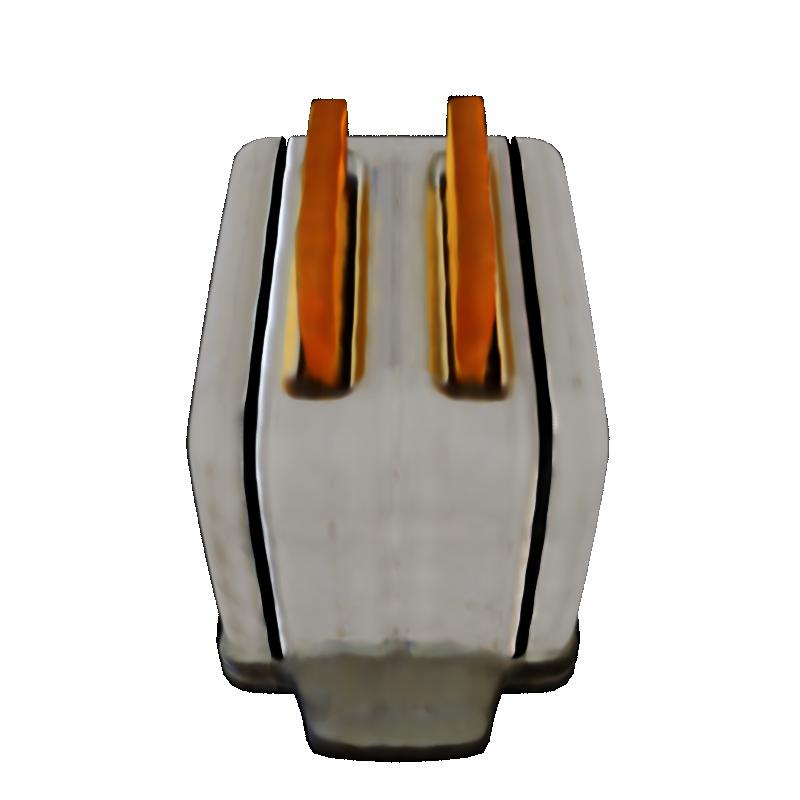} &
        \includegraphics[height=0.1\textwidth]{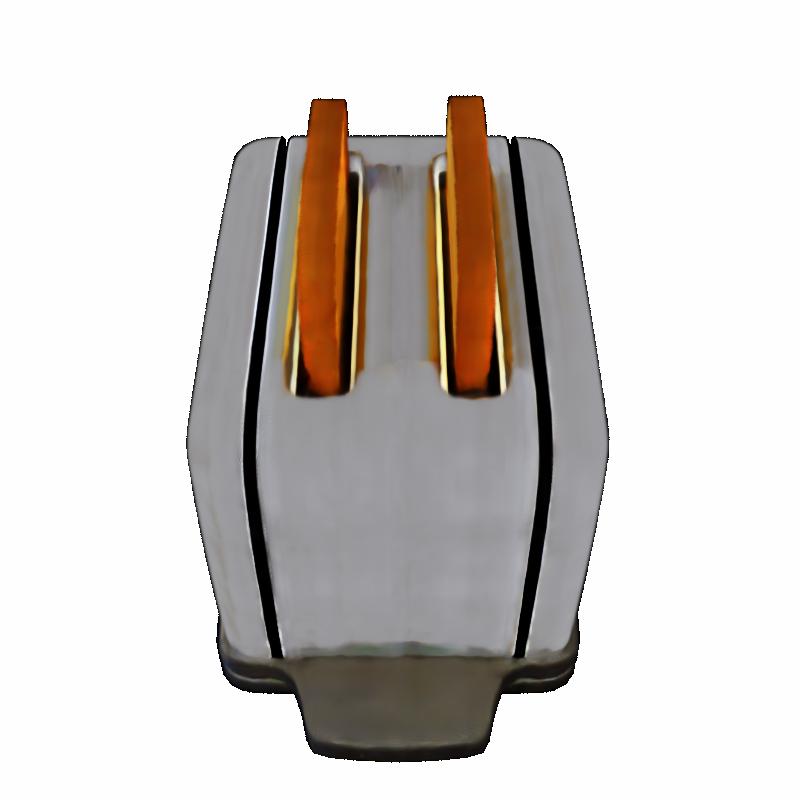} \\
    \rowname{roughness}&\includegraphics[height=0.1\textwidth]{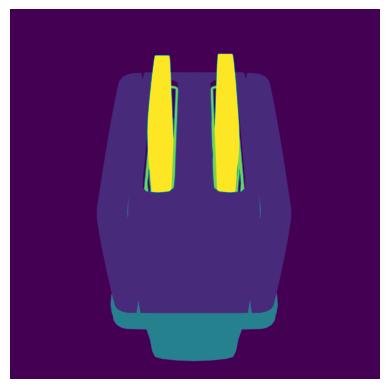} &
        \includegraphics[height=0.1\textwidth]{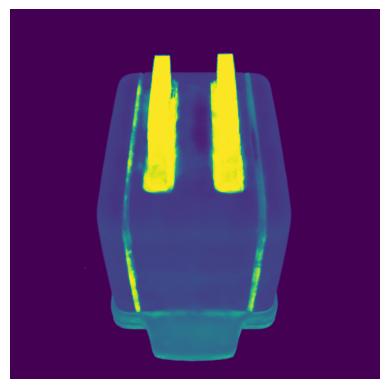} &
        \includegraphics[height=0.1\textwidth]{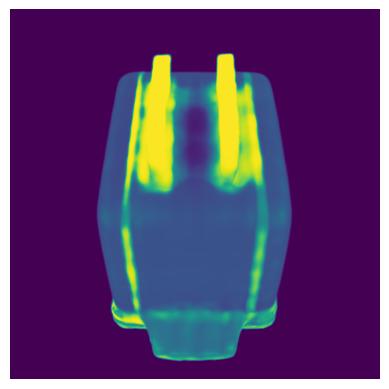} &
        \includegraphics[height=0.1\textwidth]{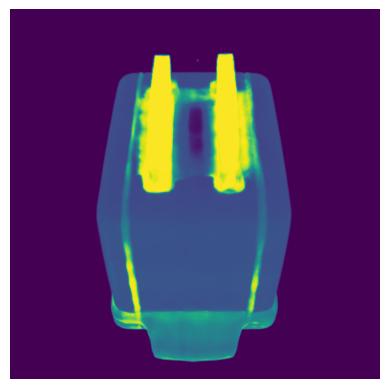} \\
        \rowname{normal}&\includegraphics[height=0.1\textwidth]{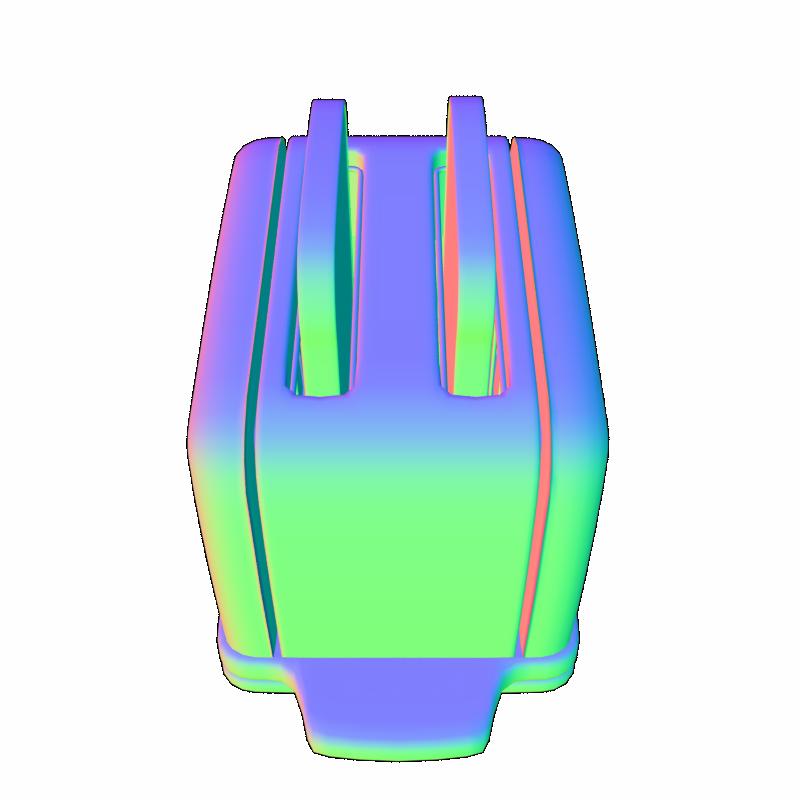} &
        \includegraphics[height=0.1\textwidth]{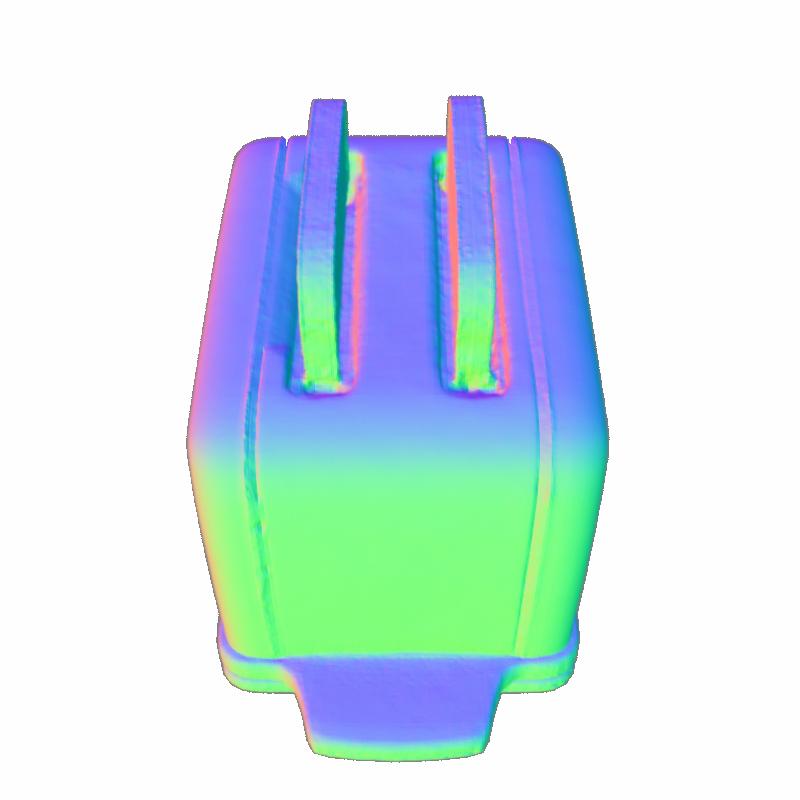} &
        \includegraphics[height=0.1\textwidth]{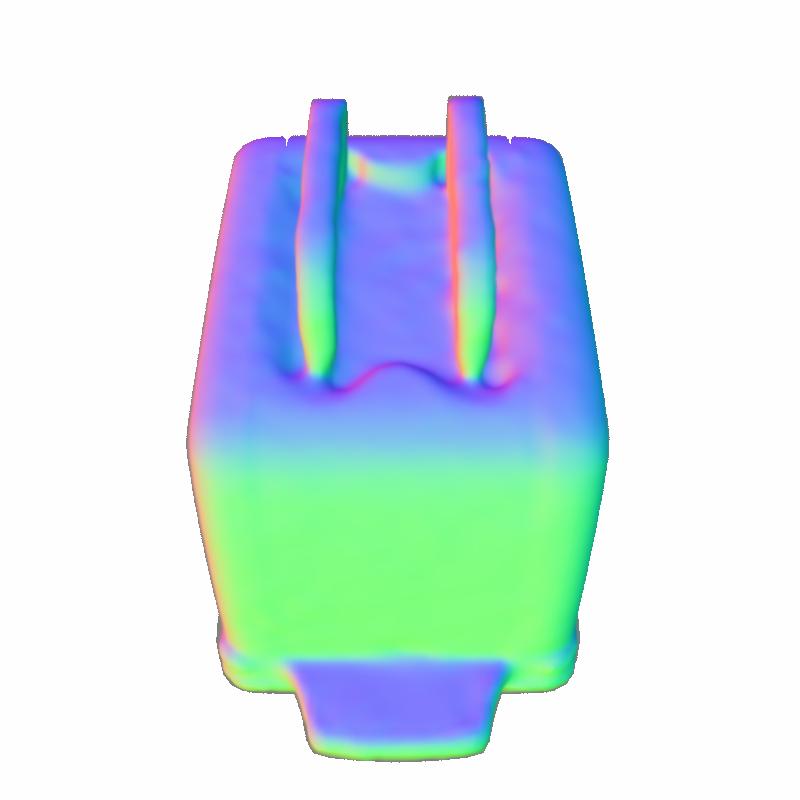} &
        \includegraphics[height=0.1\textwidth]{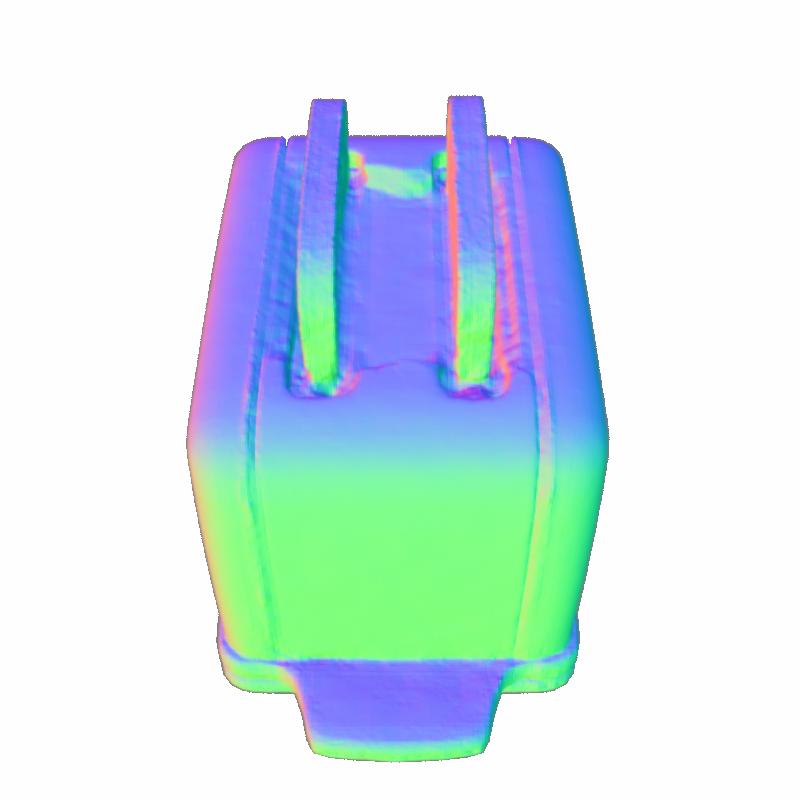} \\
        % GT & Ours & NDRMC & GShader & NMF& ENVIDR & NeRO\\
    \end{tabular} 
    \caption{\textbf{Ablation studies.}We compare our full model with our model without information sharing (physically based rendering only).}
    \label{fig:info_share}
\end{figure*}

\begin{figure*}[p]
    \centering
    \setlength\tabcolsep{1pt}
    \settoheight{\tempdima}{\includegraphics[width=0.09\textwidth]{figures/info_share/toaster_gt_albedo.jpeg}}
    \begin{tabular}{cccccc}
       &  bear & bunny & coral & maneki & vase\\
    \rowname{Mesh}&\includegraphics[height=0.1\textwidth]{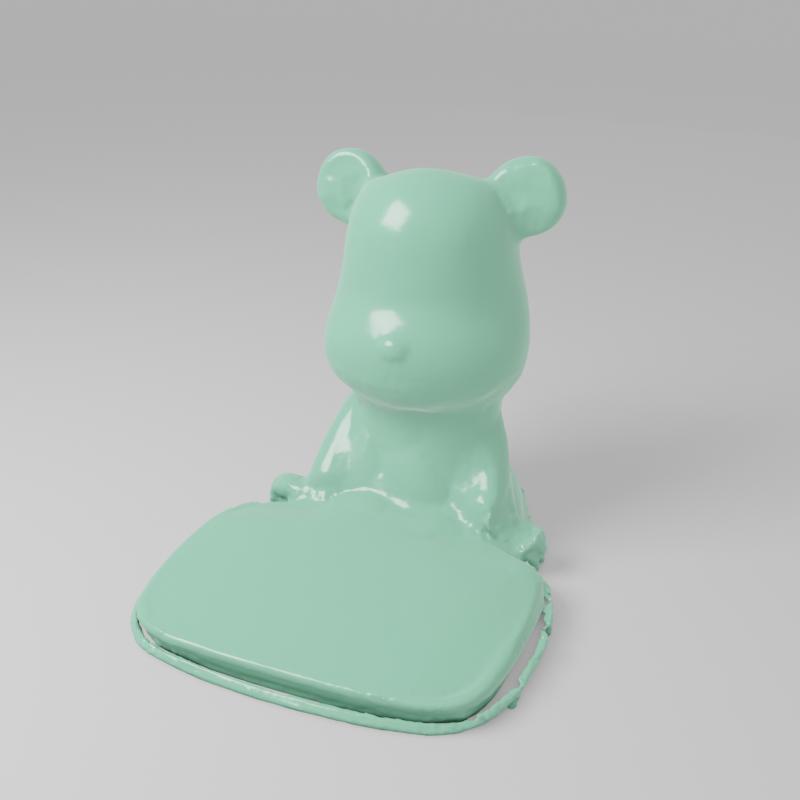} &
        \includegraphics[height=0.1\textwidth]{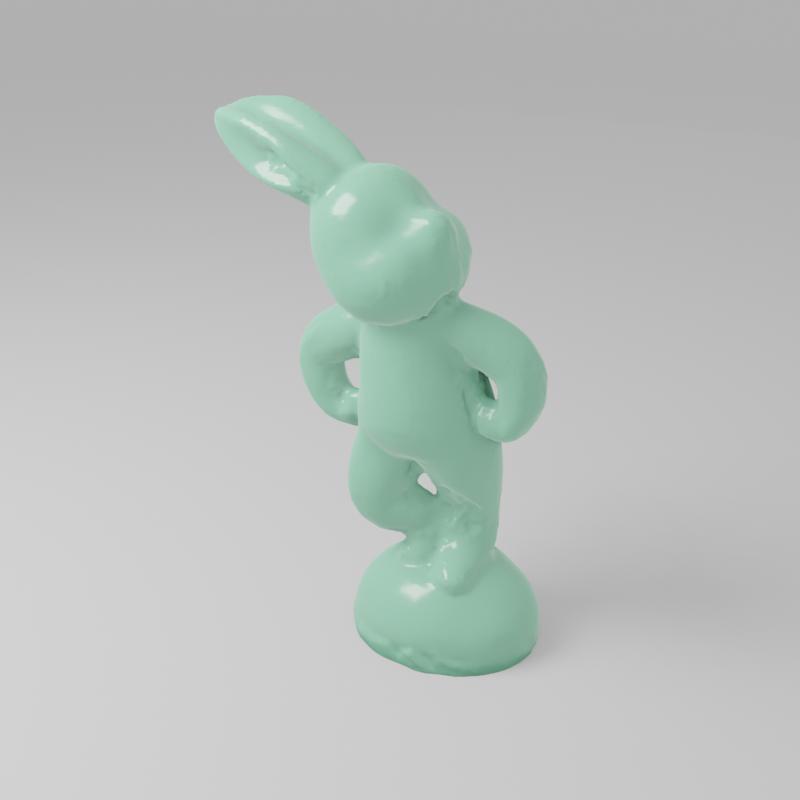} &
        \includegraphics[height=0.1\textwidth]{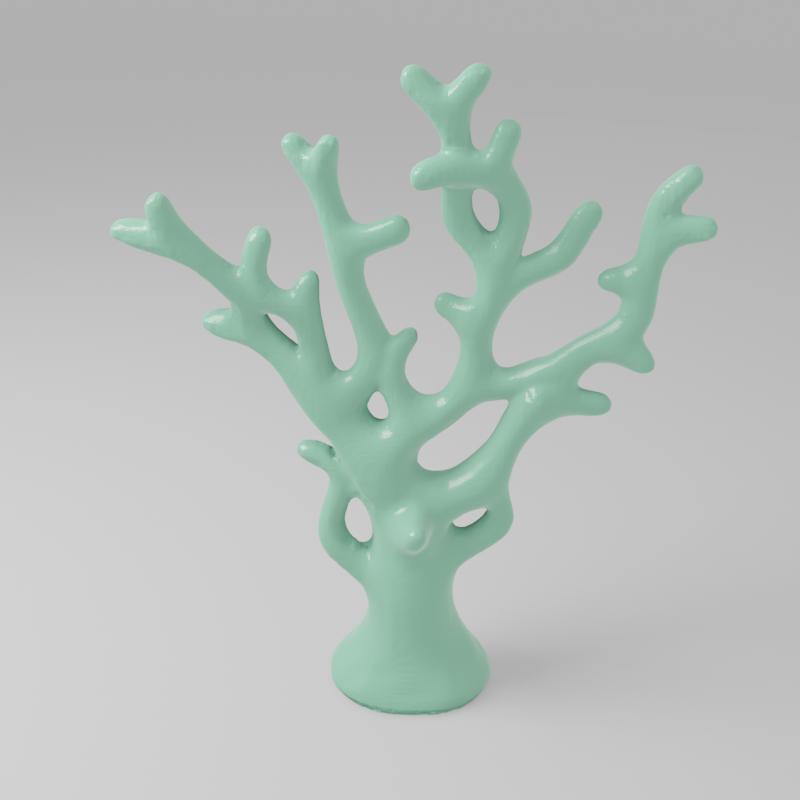} &
        \includegraphics[height=0.1\textwidth]{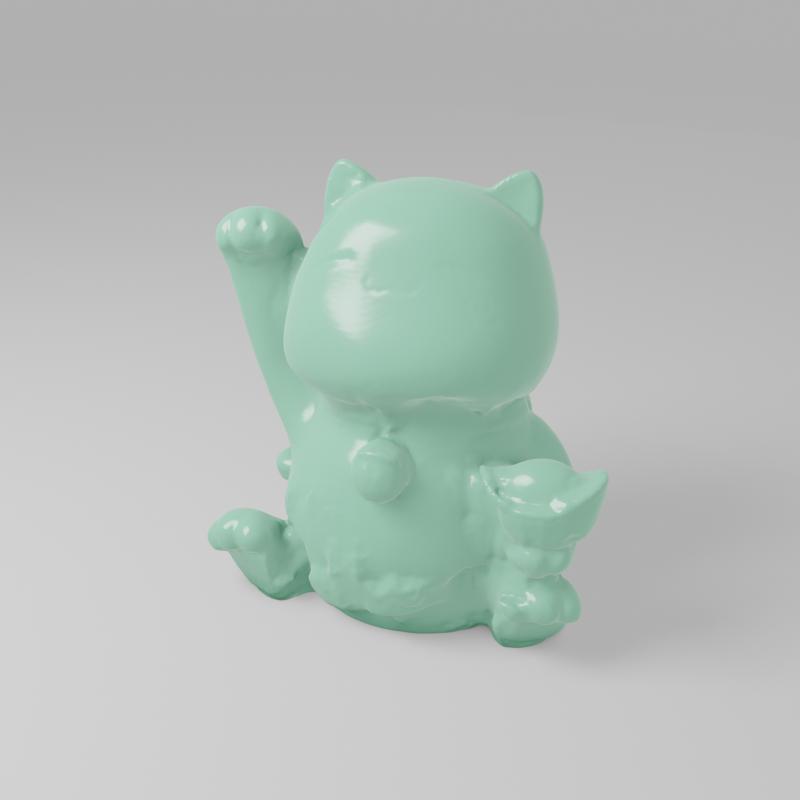} &
        \includegraphics[height=0.1\textwidth]{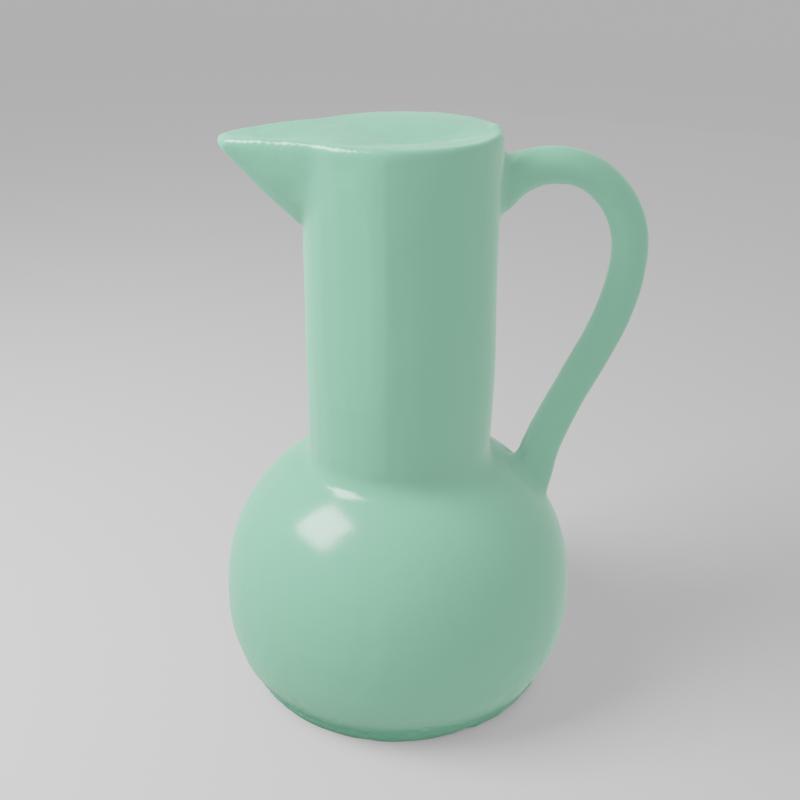} \\
         \rowname{NVS}&
        \includegraphics[height=0.1\textwidth]{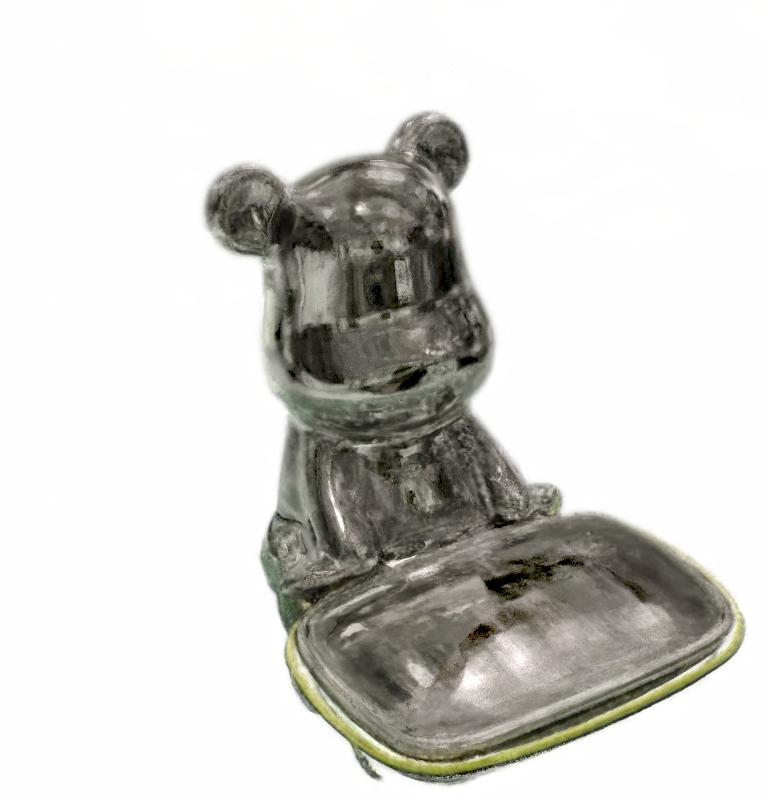} &
        \includegraphics[height=0.1\textwidth]{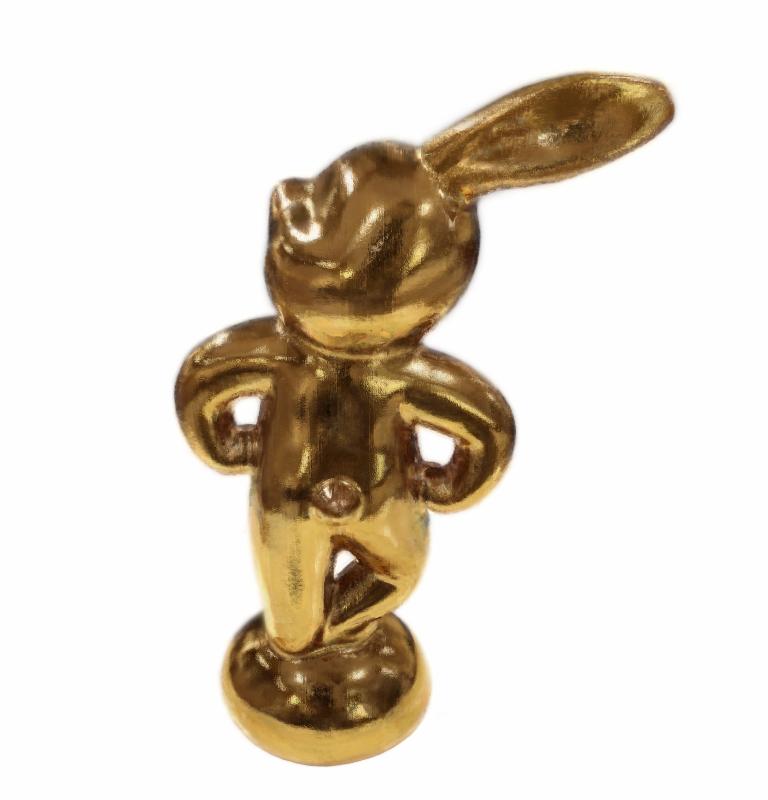} &
        \includegraphics[height=0.1\textwidth]{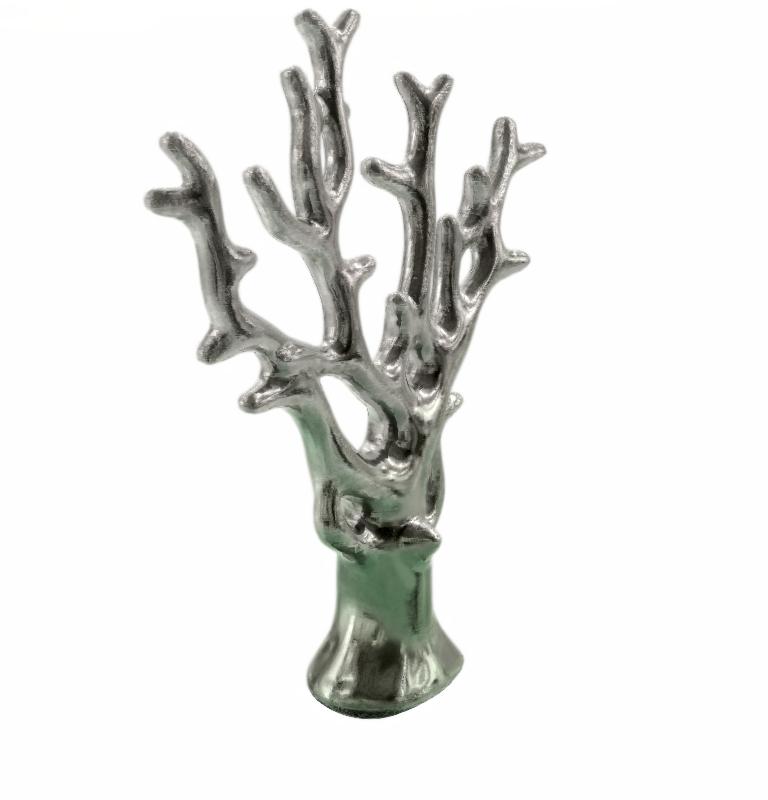} &
        \includegraphics[height=0.1\textwidth]{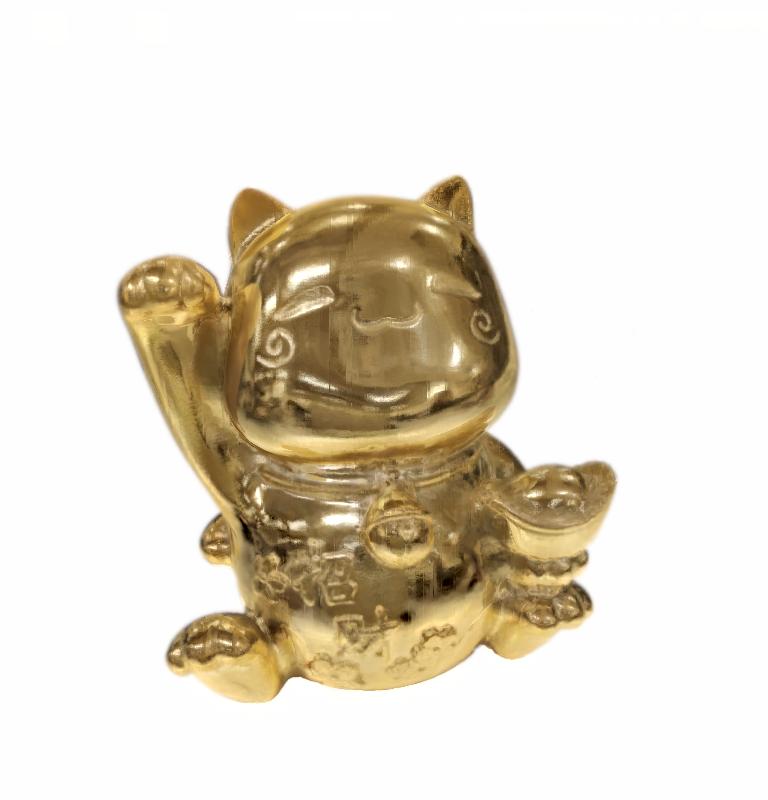} &
        \includegraphics[height=0.1\textwidth]{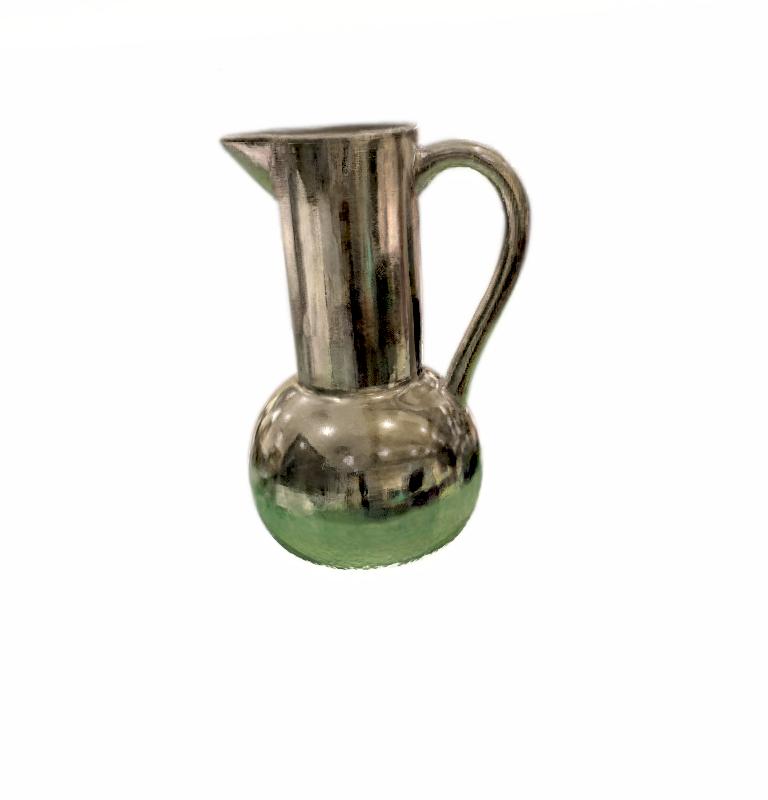} \\
        \rowname{Relighting}&\includegraphics[height=0.1\textwidth]{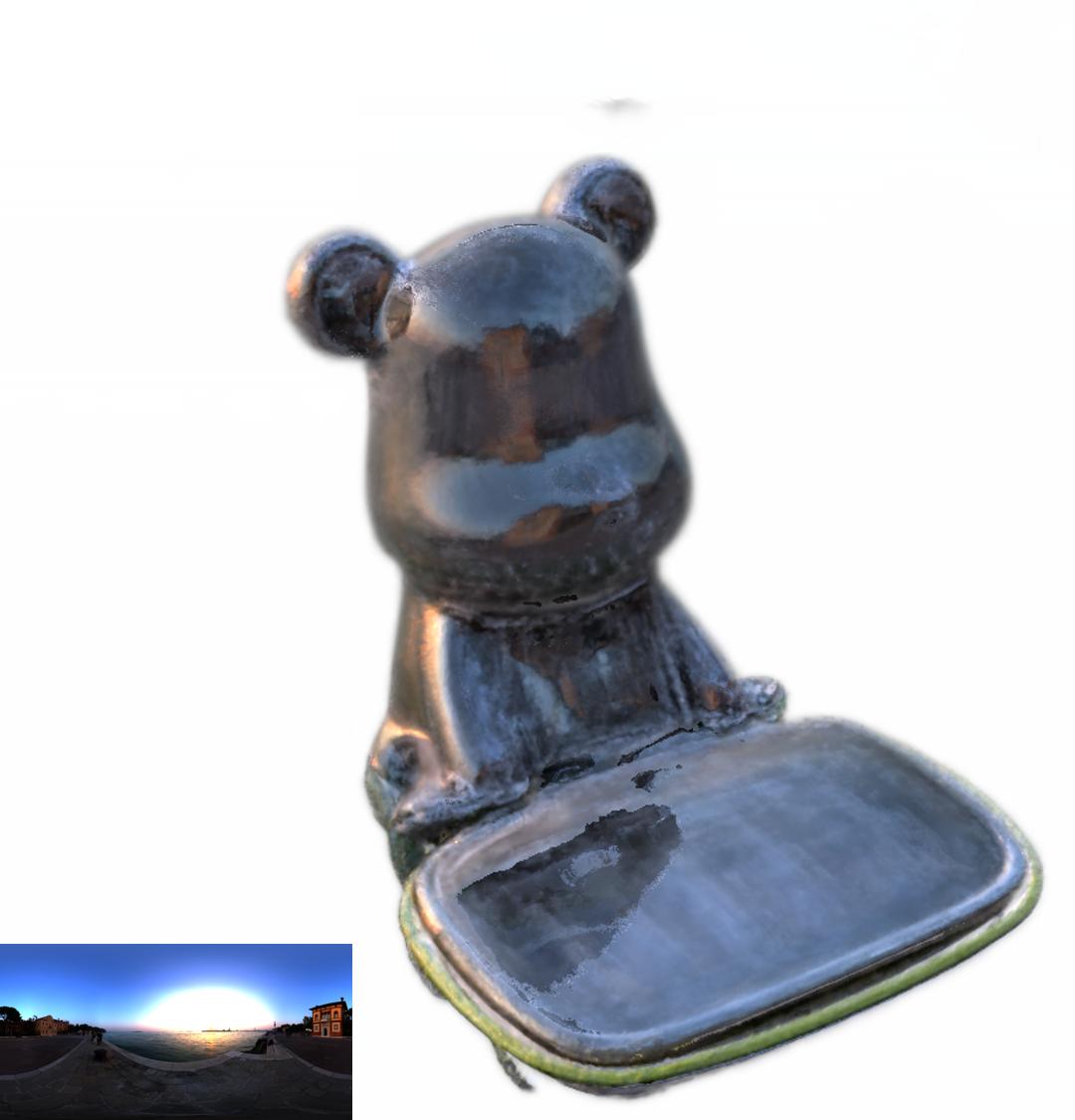} &
        \includegraphics[height=0.1\textwidth]{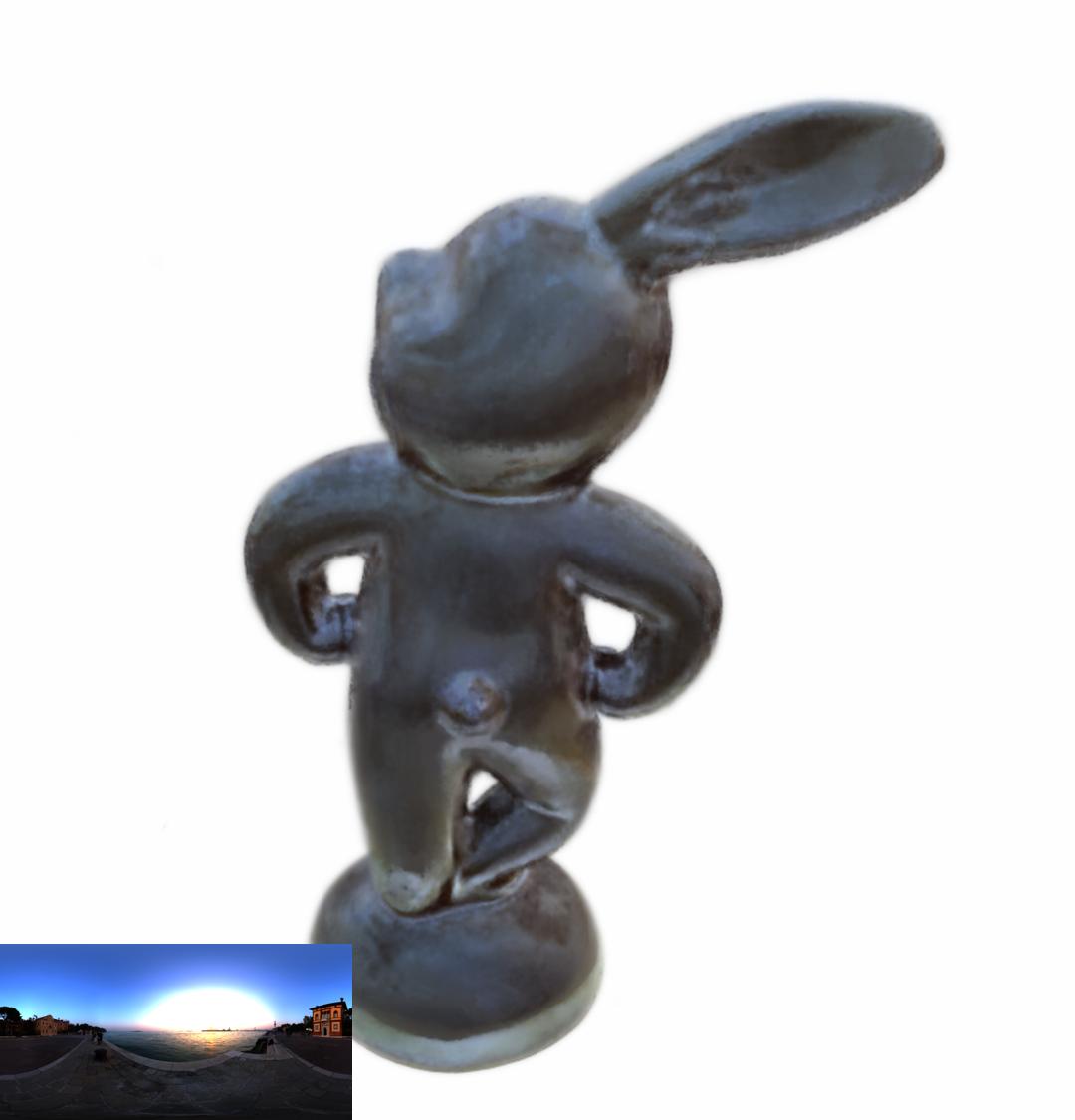} &
        \includegraphics[height=0.1\textwidth]{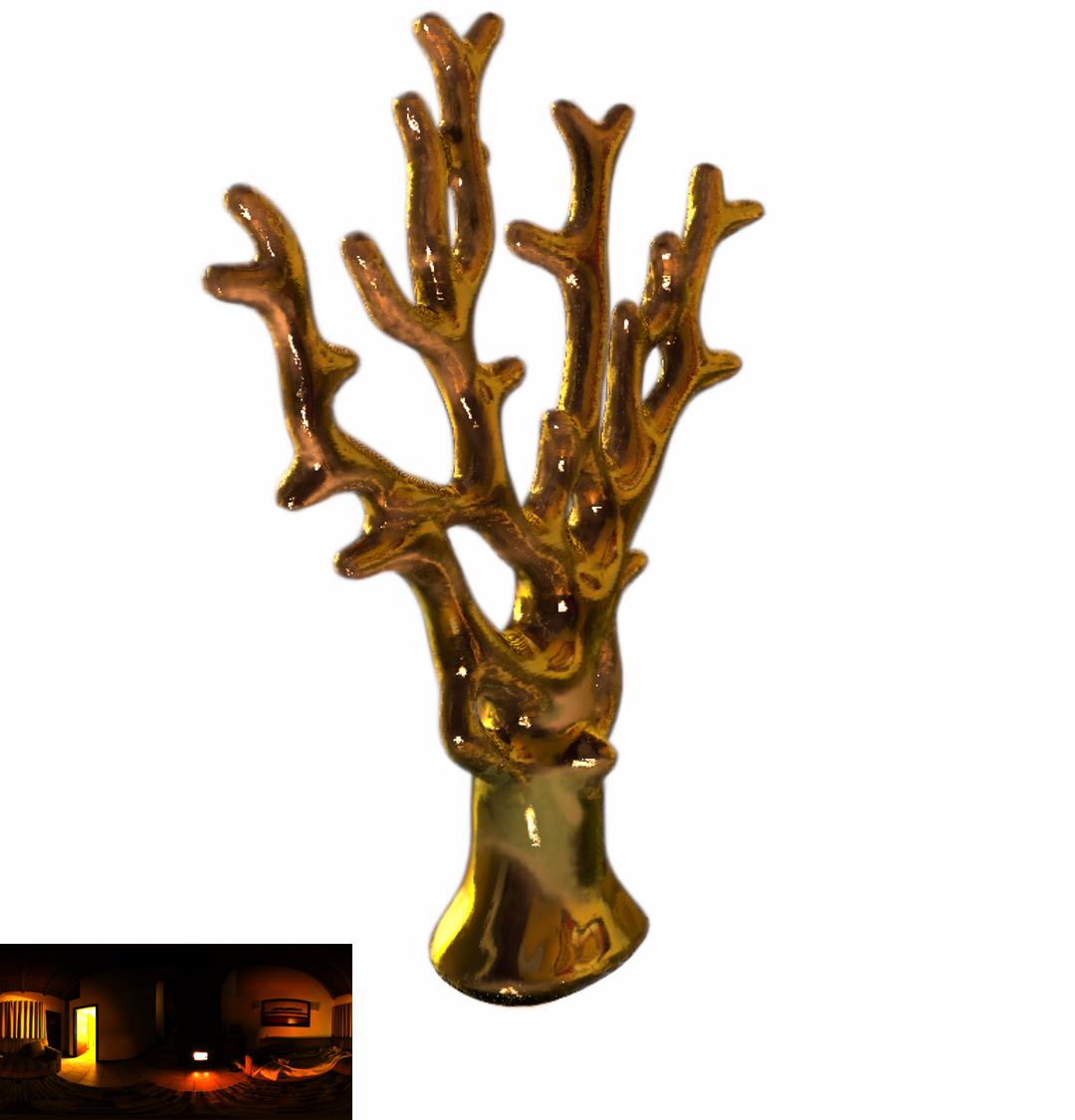} &
        \includegraphics[height=0.1\textwidth]{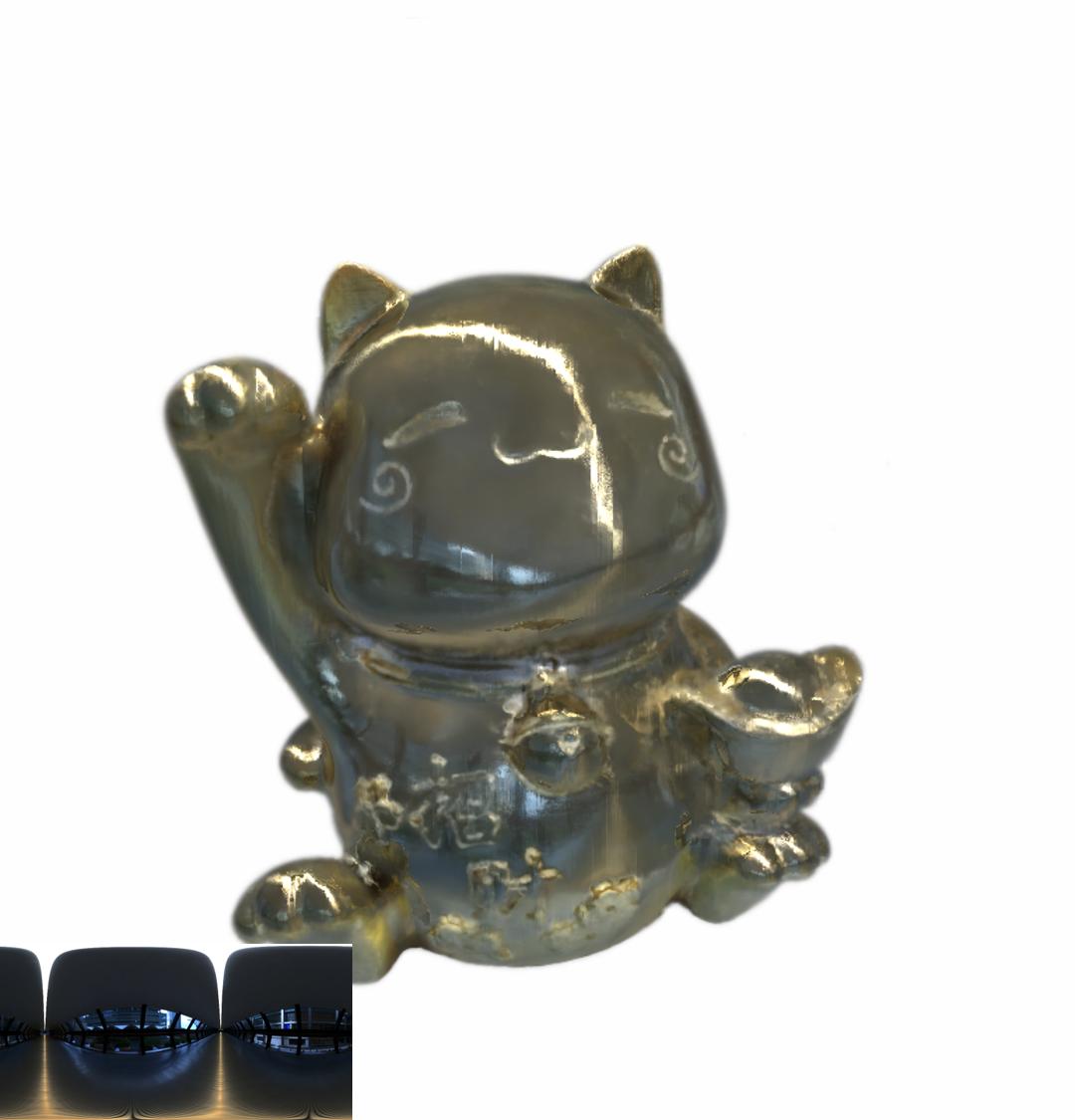} &
        \includegraphics[height=0.1\textwidth]{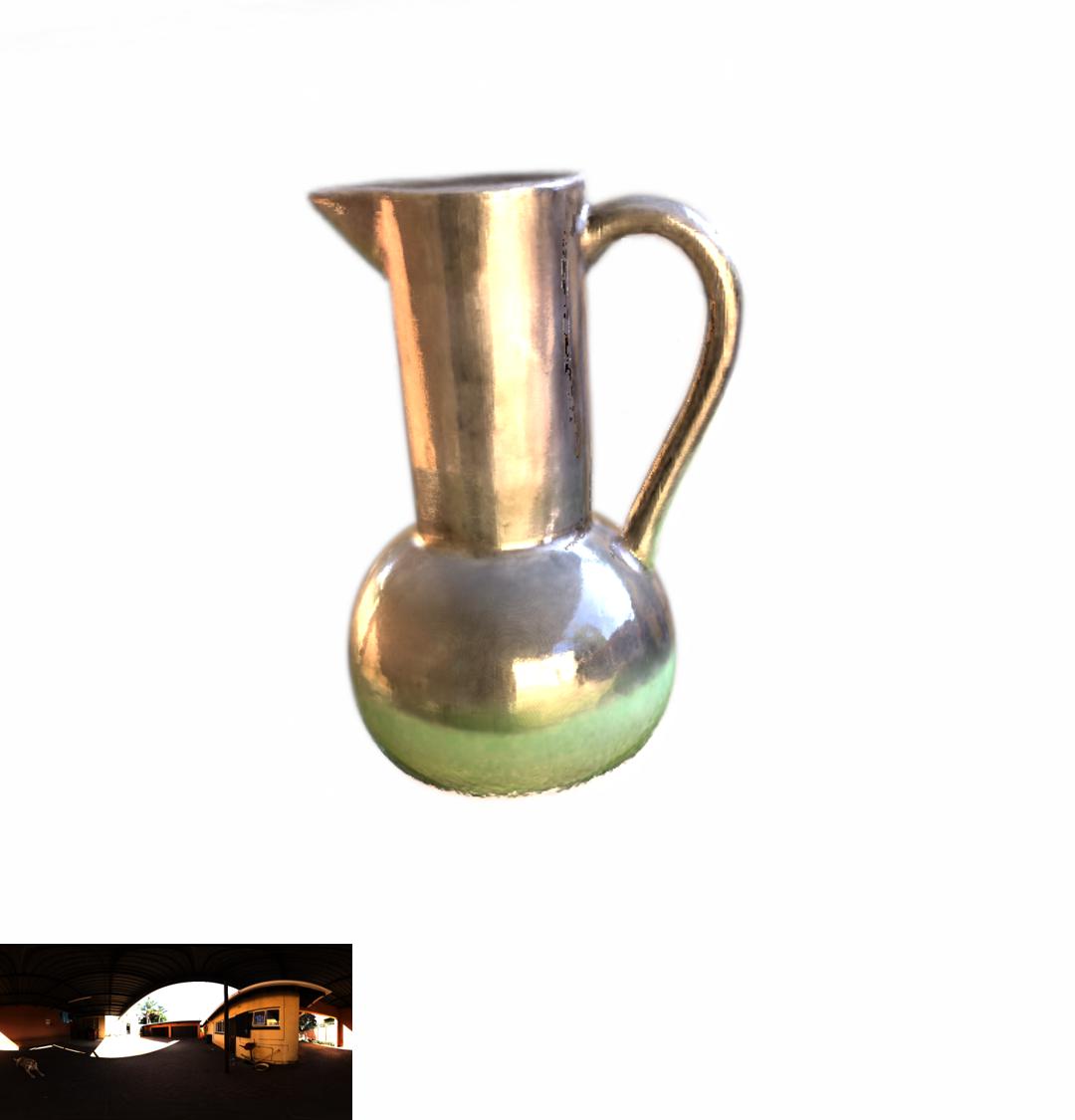} \\

    \end{tabular}
    \caption{\textbf{Extracted mesh, Novel view synthesis and relighting of the Glossy-Real dataset.} Our model is trained for 4 hours with ground truth object mask. We extract meshes from SDF using marching cubes with a resolution of 512.}
    \label{fig:glossy_real}
    \vspace{-2em}
\end{figure*}

\end{document}